\documentclass{egpubl_arxiv}

\ConferenceSubmission   
 \electronicVersion 

\ifpdf \usepackage[pdftex]{graphicx} \pdfcompresslevel=9
\else \usepackage[dvips]{graphicx} \fi

\PrintedOrElectronic

\usepackage{t1enc,dfadobe}

\usepackage{egweblnk}
\usepackage{cite}
\usepackage{amsmath}
\usepackage{amssymb}
\usepackage{subfig}

\title[Structure-preserving color transformations using Laplacian commutativity]%
      {Structure-preserving color transformations using Laplacian commutativity}

\author[Eynard et al.]
       {D. Eynard, A. Kovnatsky and M. M. Bronstein 
        \\
         Institute of Computational Science, Faculty of Informatics, University of Lugano, Switzerland\\
       }

\begin{document}

\maketitle

\begin{abstract}
Mappings between color spaces are ubiquitous in image processing problems such as gamut mapping, decolorization, and image optimization for color-blind people. Simple color transformations often result in information loss and ambiguities (for example, when mapping from RGB to grayscale), and one wishes to find an image-specific transformation that would preserve as much as possible the structure of the original image in the target color space. 
In this paper, we propose Laplacian colormaps, a generic framework for structure-preserving color transformations between images. We use the image Laplacian to capture the structural information, and show that if the color transformation between two images preserves the structure, the respective Laplacians have similar eigenvectors, or in other words, are approximately jointly diagonalizable. Employing the relation between joint diagonalizability and commutativity of matrices, we use Laplacians commutativity as a criterion of color mapping quality and minimize it w.r.t. the parameters of a color transformation to achieve optimal structure preservation. 
We show numerous applications of our approach, including color-to-gray conversion, gamut mapping, multispectral image fusion, and image optimization for color deficient viewers.


\end{abstract}


\newtheorem{theo}{Theorem}[section]
\newtheorem{prop}{Proposition}[section]
\newtheorem{lem}[theo]{Lemma}
\newcommand{\argmax}{\operatornamewithlimits{argmax~}}
\newcommand{\argmin}{\operatornamewithlimits{argmin~}}
\newcommand{\diag}{\operatorname{diag}}
\newcommand{\sinc}{\operatorname{sinc}}
\newcommand{\comb}{\operatorname{comb}}
\newcommand{\rect}{\operatorname{rect}}
\newcommand{\erf}{\operatorname{erf}}
\newcommand{\diver}{\operatorname{div}}
\newcommand{\sgn}{\operatorname{sign}}
\newcommand{\dom}{\operatorname{dom}}
\newcommand{\trace}{\operatorname{trace}}
\newcommand{\interior}{\operatorname{int}}

\newcommand{\bm}[1]{\boldsymbol{\mathrm{#1}}}

\newcommand{\bb}[1]{\pmb{\mathrm{#1}}}

\def\Tr{\mathrm{T}}

\newcommand{\EE}{\mathbb{E}}
\newcommand{\RR}{\mathbb{R}}
\newcommand{\HH}{\mathbb{H}}

\newcommand{\aaa}{\bb{a}}
\newcommand{\eee}{\bb{e}}
\newcommand{\pp}{\bb{p}}
\newcommand{\qq}{\bb{q}}
\newcommand{\uu}{\bb{u}}
\newcommand{\vv}{\bb{v}}
\newcommand{\ww}{\bb{w}}

\newcommand{\WW}{\bb{W}}

\newcommand{\xx}{\mathbf{x}}
\newcommand{\yy}{\mathbf{y}}
\newcommand{\zz}{\mathbf{z}}

\newcommand{\ones}{\mathbf{1}}
\newcommand{\zeros}{\mathbf{0}}

\newcommand{\Ee}{\mathbf{E}}

\newcommand{\Pp}{\mathbf{P}}
\newcommand{\ttt}{\mathbf{t}}

\newcommand{\Rr}{\mathbf{R}}

\newcommand{\Aa}{\mathbf{A}}
\newcommand{\Bb}{\mathbf{B}}
\newcommand{\Cc}{\mathbf{C}}
\newcommand{\Xx}{\mathbf{X}}
\newcommand{\Yy}{\mathbf{Y}}
\newcommand{\Vv}{\mathbf{V}}
\newcommand{\Uu}{\mathbf{U}}
\newcommand{\Dd}{\mathbf{D}}
\newcommand{\Ii}{\mathbf{I}}
\newcommand{\dd}{\partial}

\newcommand{\bt}{\bb{\theta}}

\newcommand{\Ww}{\bb{W}}
\newcommand{\Ll}{\mathbf{L}}
\newcommand{\LlTheta}{\tilde{\mathbf{L}}_{\Phi_\theta}}

\newcommand{\Llambda}{\mathbf{\Lambda}}
\newcommand{\Pphi}{\mathbf{\Phi}}
\newcommand{\Ppsi}{\mathbf{\Psi}}

\newcommand{\pphi}{\mathbf{\phi}}
\newcommand{\ttheta}{\mathbf{\theta}}

\newcommand{\gt}{\nabla _{\theta}}

\newcommand{\rv}{\textcolor{red}}

\newcommand{\deff}{{\: {\buildrel \Delta \over = } \:}}

\newcommand{\rgbtogray}[8] {
\begin{minipage}[b]{0.12\linewidth}
\includegraphics[width=\linewidth,trim=0 0 0 0,clip=true]{./figures/rgb2gray_#1_orig-eps-converted-to.pdf}
\end{minipage}
\begin{minipage}[b]{0.12\linewidth}
\includegraphics[width=\linewidth,trim=0 0 0 46,clip=true]{./figures/rgb2gray_#1_CIE-Y-eps-converted-to.pdf}
\end{minipage}
\begin{minipage}[b]{0.12\linewidth}
\includegraphics[width=\linewidth,trim=0 0 0 46,clip=true]{./figures/rgb2gray_#1_Gooch05-eps-converted-to.pdf}
\end{minipage}
\begin{minipage}[b]{0.12\linewidth}
\includegraphics[width=\linewidth,trim=0 0 0 46,clip=true]{./figures/rgb2gray_#1_Rasche05-eps-converted-to.pdf}
\end{minipage}
\begin{minipage}[b]{0.12\linewidth}
\includegraphics[width=\linewidth,trim=0 0 0 46,clip=true]{./figures/rgb2gray_#1_Grundland07-eps-converted-to.pdf}
\end{minipage}
\begin{minipage}[b]{0.12\linewidth}
\includegraphics[width=\linewidth,trim=0 0 0 46,clip=true]{./figures/rgb2gray_#1_Neumann07-eps-converted-to.pdf}
\end{minipage}
\begin{minipage}[b]{0.12\linewidth}
\includegraphics[width=\linewidth,trim=0 0 0 46,clip=true]{./figures/rgb2gray_#1_Smith08-eps-converted-to.pdf}
\end{minipage}
\begin{minipage}[b]{0.12\linewidth}
\includegraphics[width=\linewidth,trim=0 0 0 46,clip=true]{./figures/rgb2gray_#1_Ours-eps-converted-to.pdf}
\end{minipage}
\\
\begin{minipage}[b]{0.12\linewidth}
\hspace{13mm}
\vspace{-0.5mm}
\includegraphics[width=4.8mm,height=4.5mm,trim=0 0 0 0,clip=true]{./figures/rgb2gray_#1_bar-eps-converted-to.pdf}\\
\end{minipage}
\begin{minipage}[b]{0.12\linewidth}
\includegraphics[width=\linewidth,trim=0 0 0 46,clip=true]{./figures/rgb2gray_#1_CIE-Y_rwms-eps-converted-to.pdf}\\
\centering \small{#2}
\end{minipage}
\begin{minipage}[b]{0.12\linewidth}
\includegraphics[width=\linewidth,trim=0 0 0 46,clip=true]{./figures/rgb2gray_#1_Gooch05_rwms-eps-converted-to.pdf}\\
\centering \small{#3}
\end{minipage}
\begin{minipage}[b]{0.12\linewidth}
\includegraphics[width=\linewidth,trim=0 0 0 46,clip=true]{./figures/rgb2gray_#1_Rasche05_rwms-eps-converted-to.pdf}\\
\centering \small{#4}
\end{minipage}
\begin{minipage}[b]{0.12\linewidth}
\includegraphics[width=\linewidth,trim=0 0 0 46,clip=true]{./figures/rgb2gray_#1_Grundland07_rwms-eps-converted-to.pdf}\\
\centering \small{#5}
\end{minipage}
\begin{minipage}[b]{0.12\linewidth}
\includegraphics[width=\linewidth,trim=0 0 0 46,clip=true]{./figures/rgb2gray_#1_Neumann07_rwms-eps-converted-to.pdf}\\
\centering \small{#6}
\end{minipage}
\begin{minipage}[b]{0.12\linewidth}
\includegraphics[width=\linewidth,trim=0 0 0 46,clip=true]{./figures/rgb2gray_#1_Smith08_rwms-eps-converted-to.pdf}\\
\centering \small{#7}
\end{minipage}
\begin{minipage}[b]{0.12\linewidth}
\includegraphics[width=\linewidth,trim=0 0 0 46,clip=true]{./figures/rgb2gray_#1_Ours_rwms-eps-converted-to.pdf}\\
\centering \small{#8}
\end{minipage}
}

\newcommand{\rgbtograyCadik}[8] {
\begin{minipage}[b]{0.12\linewidth} 
\includegraphics[width=\linewidth,trim=0 0 0 0,clip=true]{./figures/cadik/rgb2gray_#1_orig-eps-converted-to.pdf}
\end{minipage}
\begin{minipage}[b]{0.12\linewidth}
\includegraphics[width=\linewidth,trim=0 0 0 46,clip=true]{./figures/cadik/rgb2gray_#1_CIE-Y-eps-converted-to.pdf}
\end{minipage}
\begin{minipage}[b]{0.12\linewidth}
\includegraphics[width=\linewidth,trim=0 0 0 46,clip=true]{./figures/cadik/rgb2gray_#1_Gooch05-eps-converted-to.pdf}
\end{minipage}
\begin{minipage}[b]{0.12\linewidth}
\includegraphics[width=\linewidth,trim=0 0 0 46,clip=true]{./figures/cadik/rgb2gray_#1_Rasche05-eps-converted-to.pdf}
\end{minipage}
\begin{minipage}[b]{0.12\linewidth}
\includegraphics[width=\linewidth,trim=0 0 0 46,clip=true]{./figures/cadik/rgb2gray_#1_Grundland07-eps-converted-to.pdf}
\end{minipage}
\begin{minipage}[b]{0.12\linewidth}
\includegraphics[width=\linewidth,trim=0 0 0 46,clip=true]{./figures/cadik/rgb2gray_#1_Neumann07-eps-converted-to.pdf}
\end{minipage}
\begin{minipage}[b]{0.12\linewidth}
\includegraphics[width=\linewidth,trim=0 0 0 46,clip=true]{./figures/cadik/rgb2gray_#1_Smith08-eps-converted-to.pdf}
\end{minipage}
\begin{minipage}[b]{0.12\linewidth}
\includegraphics[width=\linewidth,trim=0 0 0 46,clip=true]{./figures/cadik/rgb2gray_#1_Ours-eps-converted-to.pdf}
\end{minipage}
\\
\begin{minipage}[b]{0.12\linewidth}
\hspace{13mm}
\vspace{-0.5mm}
\includegraphics[width=4.8mm,height=4.5mm,trim=0 0 0 0,clip=true]{./figures/cadik/rgb2gray_#1_bar-eps-converted-to.pdf}\\
\end{minipage}
\begin{minipage}[b]{0.12\linewidth}
\includegraphics[width=\linewidth,trim=0 0 0 46,clip=true]{./figures/cadik/rgb2gray_#1_CIE-Y_rwms-eps-converted-to.pdf}\\
\centering \small{#2}
\end{minipage}
\begin{minipage}[b]{0.12\linewidth}
\includegraphics[width=\linewidth,trim=0 0 0 46,clip=true]{./figures/cadik/rgb2gray_#1_Gooch05_rwms-eps-converted-to.pdf}\\
\centering \small{#3}
\end{minipage}
\begin{minipage}[b]{0.12\linewidth}
\includegraphics[width=\linewidth,trim=0 0 0 46,clip=true]{./figures/cadik/rgb2gray_#1_Rasche05_rwms-eps-converted-to.pdf}\\
\centering \small{#4}
\end{minipage}
\begin{minipage}[b]{0.12\linewidth}
\includegraphics[width=\linewidth,trim=0 0 0 46,clip=true]{./figures/cadik/rgb2gray_#1_Grundland07_rwms-eps-converted-to.pdf}\\
\centering \small{#5}
\end{minipage}
\begin{minipage}[b]{0.12\linewidth}
\includegraphics[width=\linewidth,trim=0 0 0 46,clip=true]{./figures/cadik/rgb2gray_#1_Neumann07_rwms-eps-converted-to.pdf}\\
\centering \small{#6}
\end{minipage}
\begin{minipage}[b]{0.12\linewidth}
\includegraphics[width=\linewidth,trim=0 0 0 46,clip=true]{./figures/cadik/rgb2gray_#1_Smith08_rwms-eps-converted-to.pdf}\\
\centering \small{#7}
\end{minipage}
\begin{minipage}[b]{0.12\linewidth}
\includegraphics[width=\linewidth,trim=0 0 0 46,clip=true]{./figures/cadik/rgb2gray_#1_Ours_rwms-eps-converted-to.pdf}\\
\centering \small{#8}
\end{minipage}
}

\newcommand{\cieLab}[1] {
\begin{minipage}[b]{0.25\linewidth} 
\includegraphics[width=\linewidth,trim=0 0 0 0,clip=true]{./figures/cadik/rgb2gray_#1_orig}
\end{minipage}
\begin{minipage}[b]{0.25\linewidth}
\includegraphics[width=\linewidth,trim=0 0 0 46,clip=true]{./figures/cadik/rgb2gray_#1_Ours}
\end{minipage}
\begin{minipage}[b]{0.25\linewidth}
\includegraphics[width=\linewidth,trim=0 0 0 0,clip=true]{./figures/cieLab/cielab_#1}
\end{minipage}
}

\section{Introduction}

A wide class of image processing problems relies on transformations between color spaces. 
Some notable examples include 
gamut mapping, 
image optimization for color-deficient viewers, 
and multispectral image fusion. 
Often these transformations imply a reduction in the dimensionality of the original color space, resulting in information loss and ambiguities. 

\begin{figure*}[ht]
\begin{minipage}[b]{0.13\linewidth}
\centering \small Original image
\end{minipage}
\begin{minipage}[b]{0.26\linewidth}
\centering \small Luma 
\end{minipage}
\begin{minipage}[b]{0.26\linewidth}
\centering \small \cite{Lau11}
\end{minipage}
\begin{minipage}[b]{0.26\linewidth}
\centering \small \textbf{Laplacian}
\end{minipage}
\begin{minipage}[b]{0.05\linewidth}
\end{minipage}
\\
\begin{minipage}[b]{0.13\linewidth}
\includegraphics[width=\linewidth,trim=0 0 0 0,clip=true]{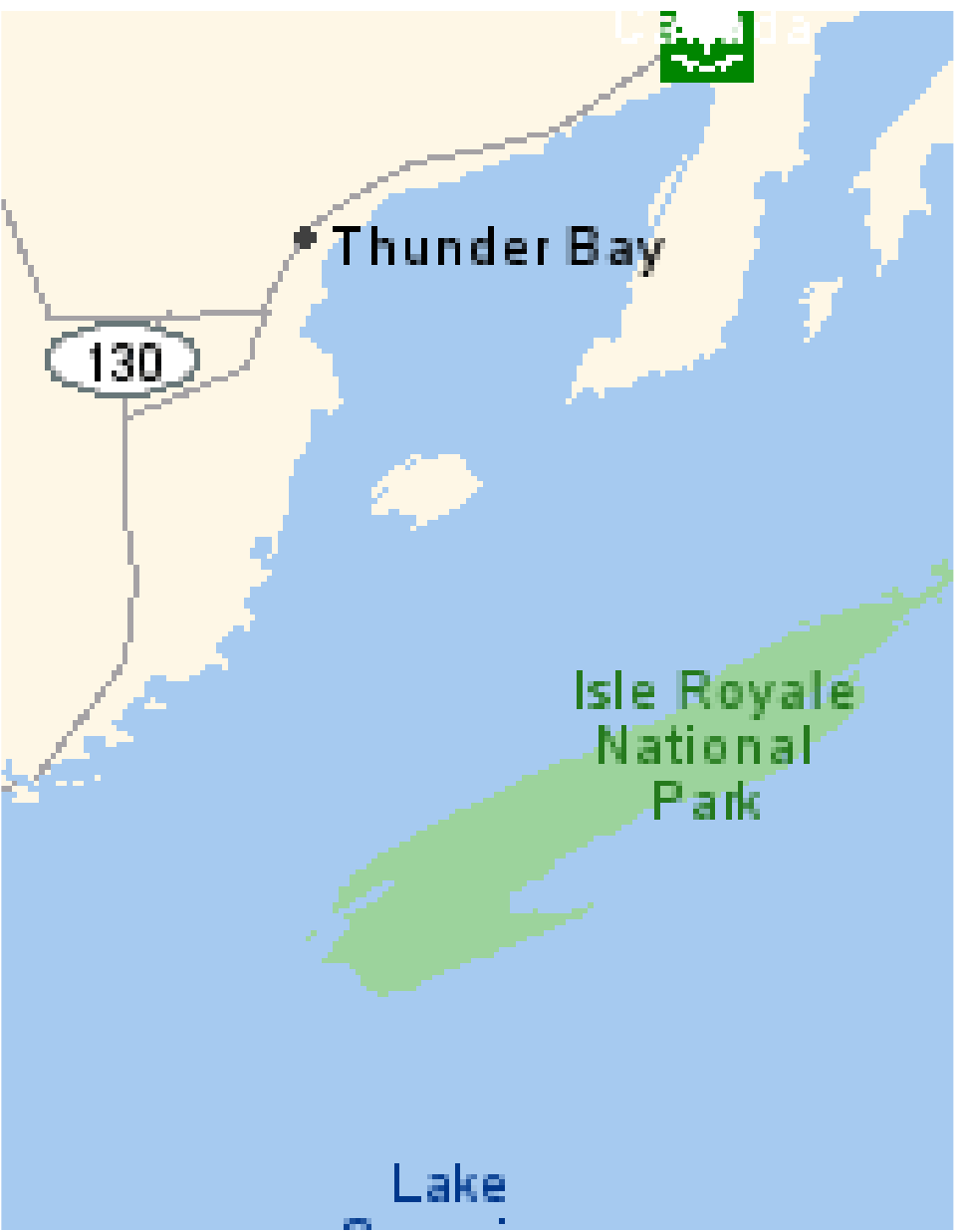}\\
\centering \small \ 
\end{minipage}
\begin{minipage}[b]{0.13\linewidth}
\includegraphics[width=\linewidth,trim=0 0 0 42,clip=true]{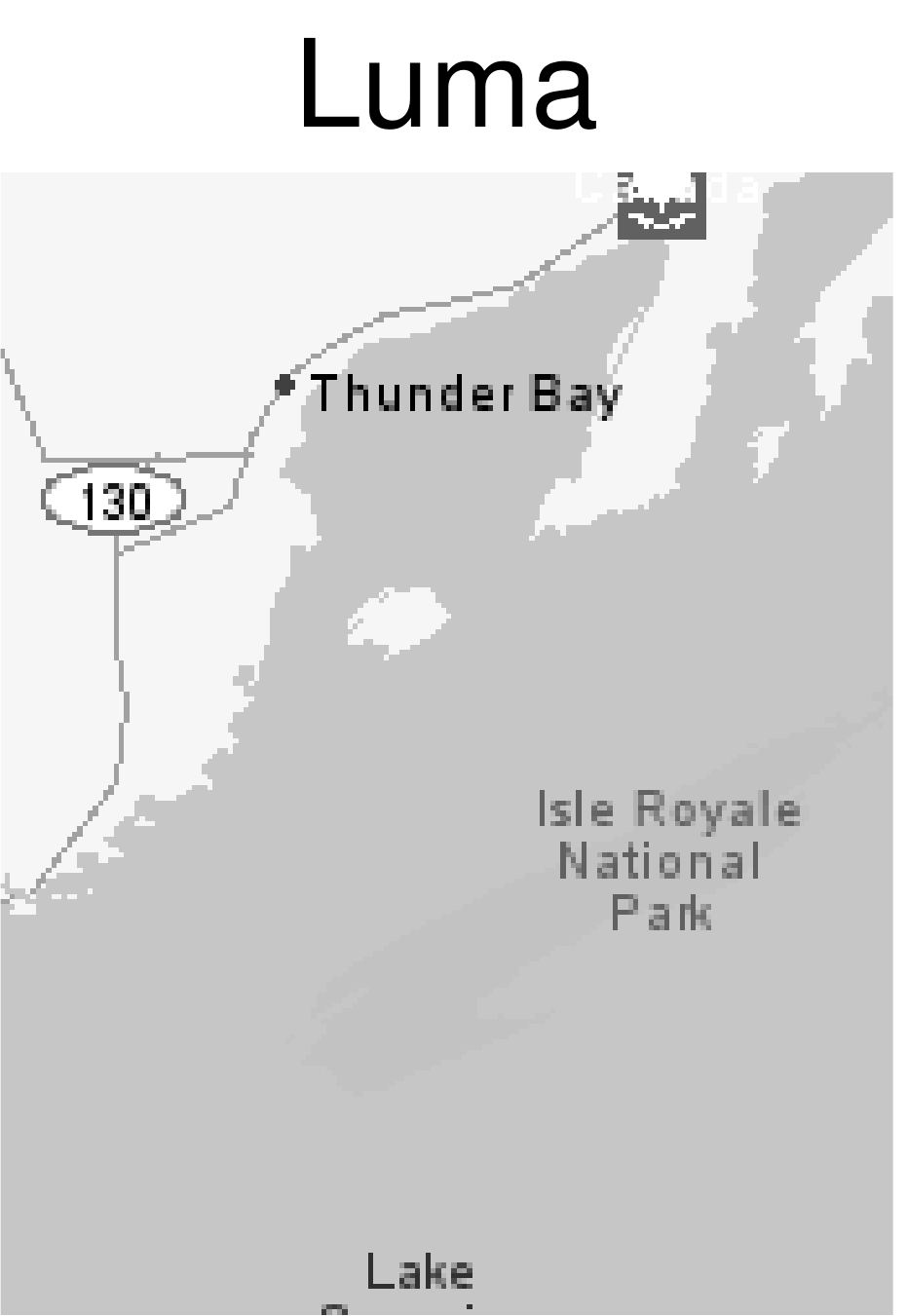}\\
\centering \small \ 
\end{minipage}
\begin{minipage}[b]{0.13\linewidth}
\includegraphics[width=\linewidth,trim=0 0 0 42,clip=true]{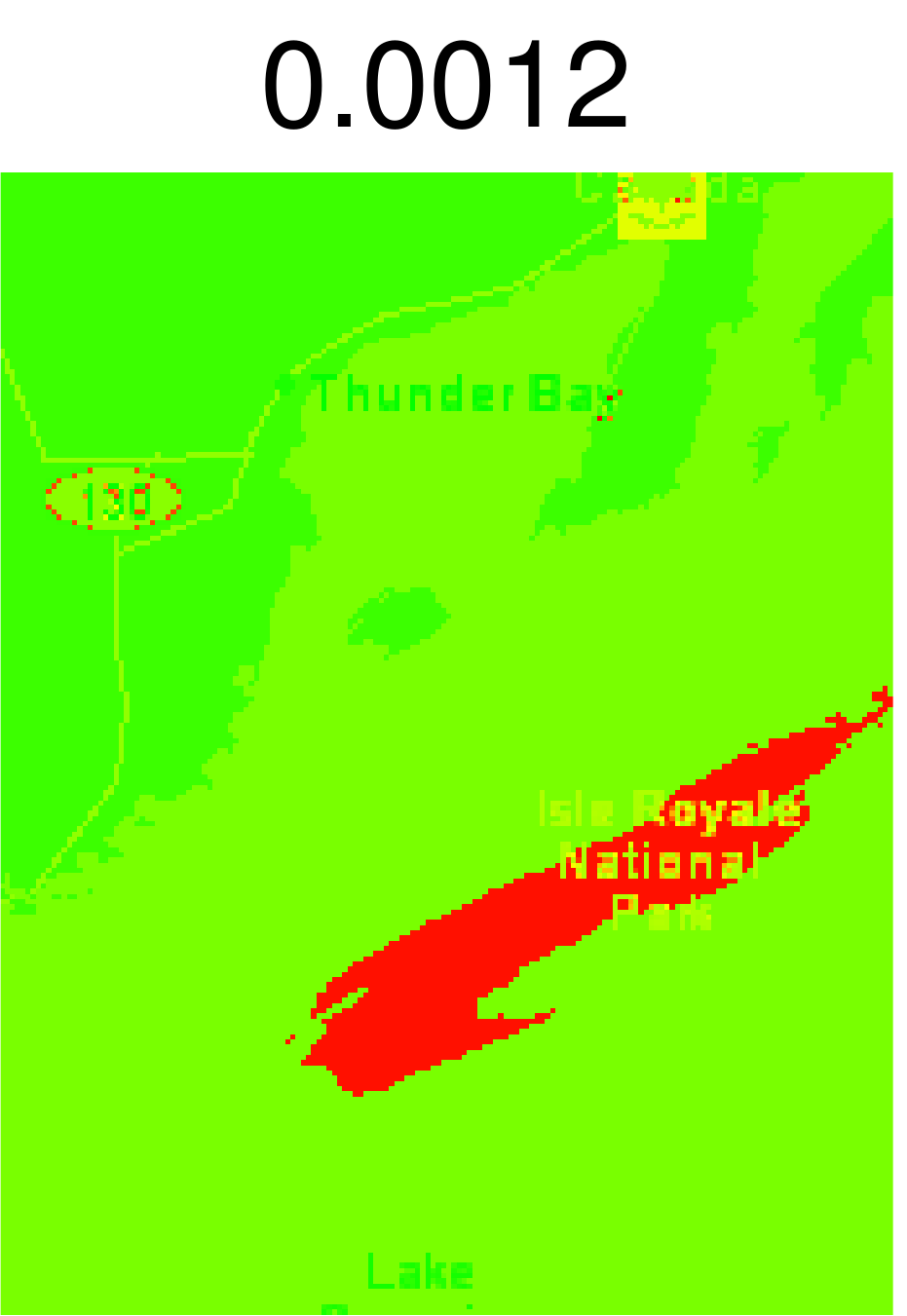}\\
\centering \small 0.98
\end{minipage}
\begin{minipage}[b]{0.13\linewidth}
\includegraphics[width=\linewidth,trim=0 0 0 42,clip=true]{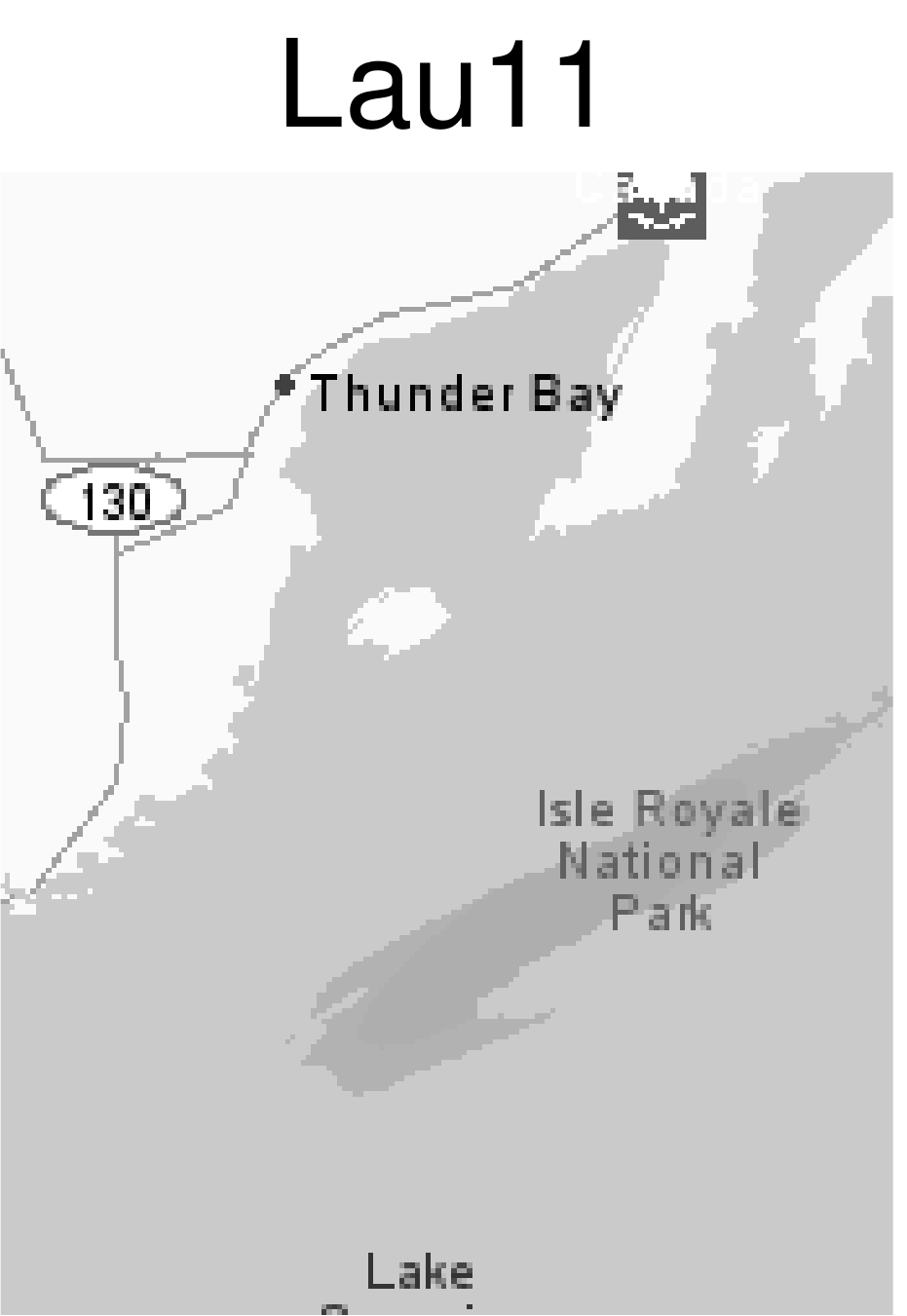}\\
\centering \small \ 
\end{minipage}
\begin{minipage}[b]{0.13\linewidth}
\includegraphics[width=\linewidth,trim=0 0 0 42,clip=true]{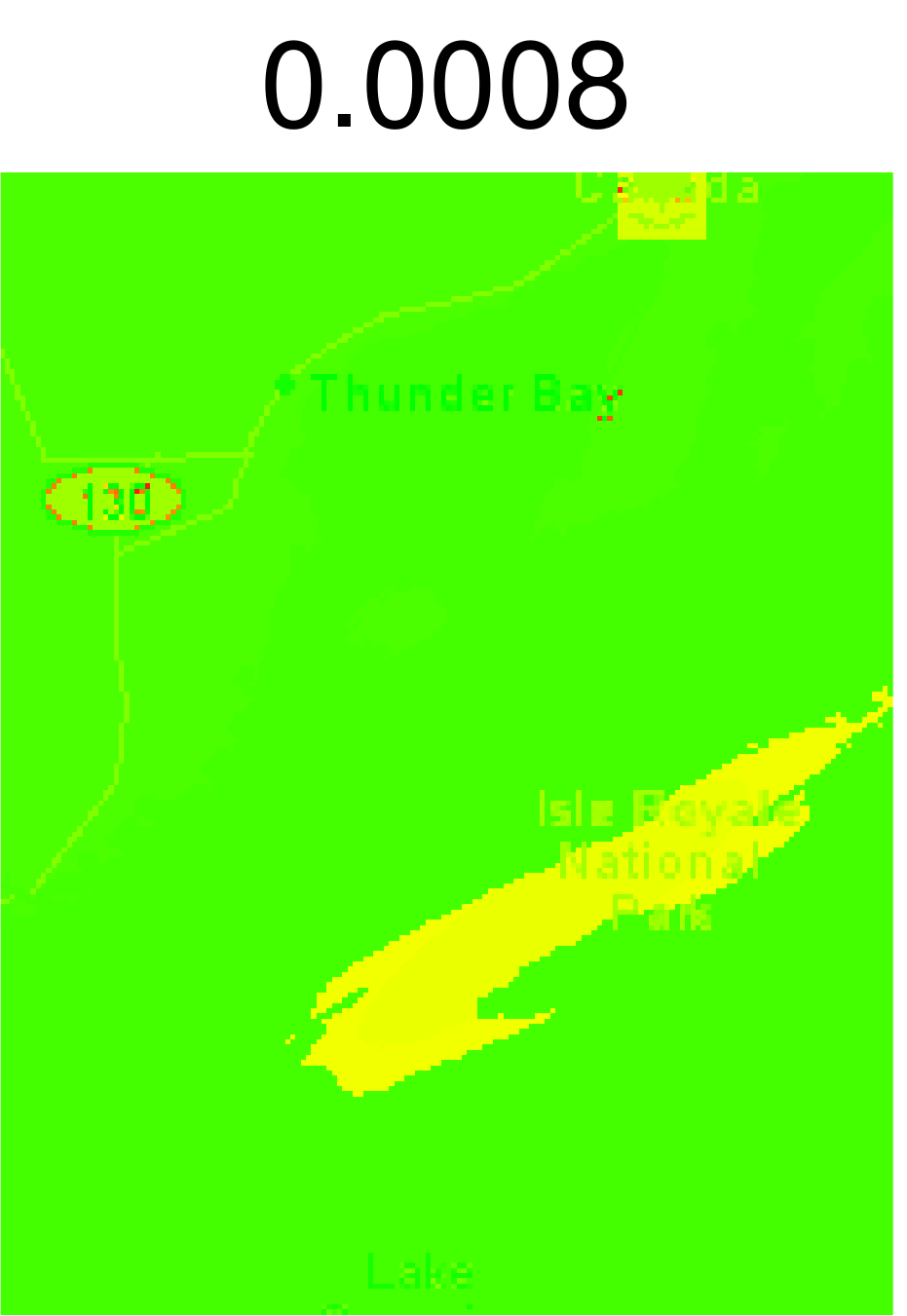}\\
\centering \small 1.23
\end{minipage}
\begin{minipage}[b]{0.13\linewidth}
\includegraphics[width=\linewidth,trim=0 0 0 42,clip=true]{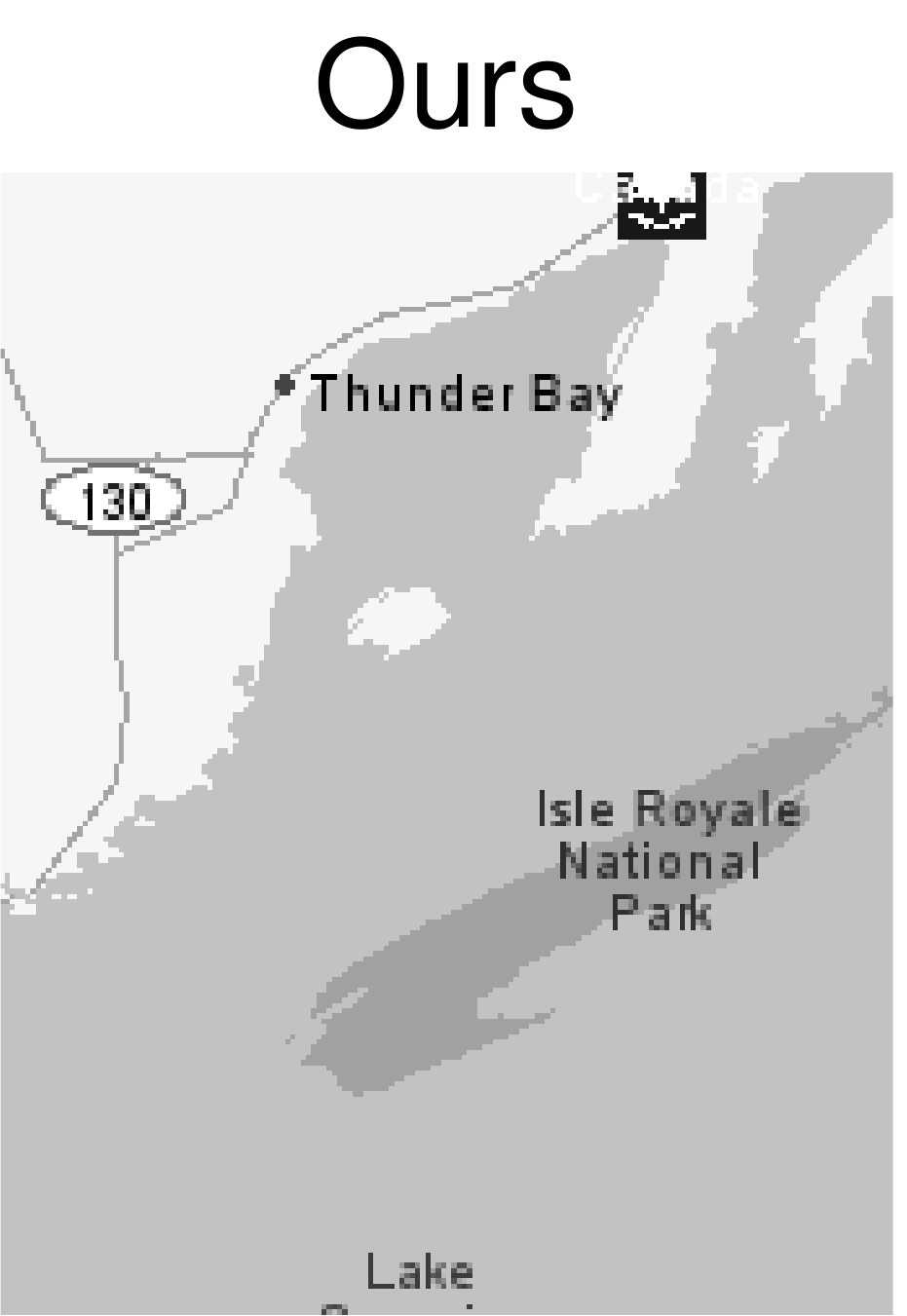}\\
\centering \small \ 
\end{minipage}
\begin{minipage}[b]{0.13\linewidth}
\includegraphics[width=\linewidth,trim=0 0 0 42,clip=true]{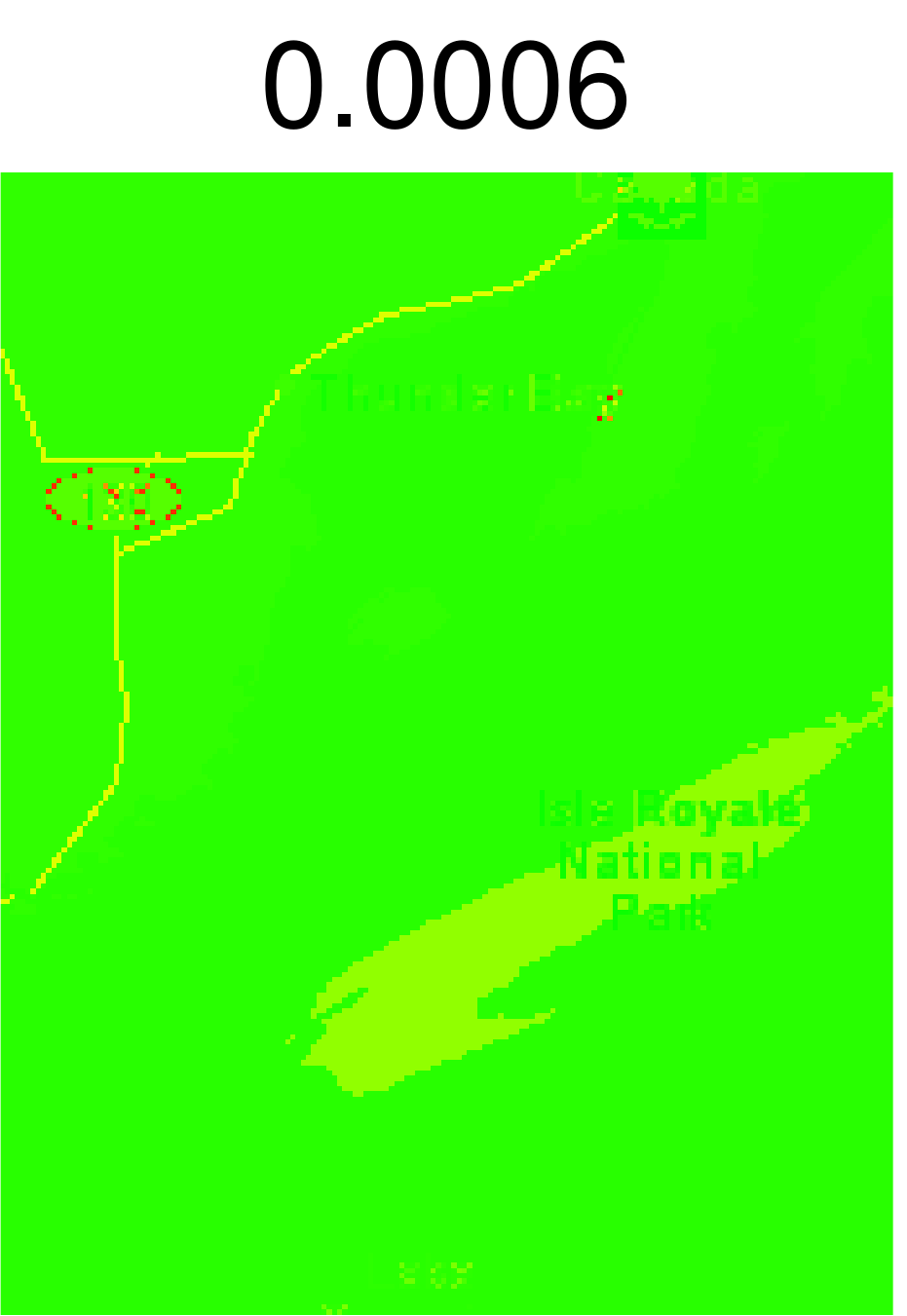}\\
\centering \small 0.50
\end{minipage}
\begin{minipage}[b]{0.05\linewidth}
\includegraphics[width=4.8mm,height=4.5mm,trim=0 0 0 0,clip=true]{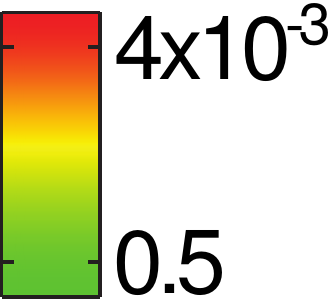}\\
\end{minipage}
\caption{Decolorization experiment results. From left: original RGB image, grayscale conversion results using the Luma channel, the method of \cite{Lau11}, and our Laplacian colormap, with their respective RWMS error images and mean RWMS score values. Luma conversion results in loss of image structure due to metamerism (the green island disappears). The proposed Laplacian colormap better preserves the original image structure. }
\label{fig:rgb2grayLau}
\end{figure*}

\textbf{Decolorization} or {\em color-to-gray conversion} is a classical example one frequently encounters when printing a color image on a black-and-white printer. The ambiguity of such a conversion (called {\em metamerism}, when many different RGB colors are mapped to the same gray level) may result in a loss of important structure in the image (see Figure~\ref{fig:rgb2grayLau}). 
Preserving salient characteristics of the original image is thus crucial for a quality color transformation process. 
These characteristics can be represented in different ways, e.g. as contrasts between color pixels in terms of their luminance and chrominance~\cite{Gooch05}, color distances~\cite{Grundland07}, image gradients~\cite{Zhou12} and Laplacians~\cite{bansal2013joint}.  

Color-to-gray maps can be classified into \emph{global} (using the same map for each pixel) and \emph{local} (or spatial, allowing different pixels with the same color to be mapped to different gray values, at the advantage of a better perception of color contrasts).
Members of the first group include the pixel-based approaches by Gooch et al. \cite{Gooch05} and Grundland et al. \cite{Grundland07}, and the color-based ones by Rasche et al. \cite{Rasche05}, Kuhn et al. \cite{Kuhn08}, Kim et al. \cite{Kim09}, Lu et al. \cite{Lu12}.
Between local methods \cite{Neumann07,Kuk10,Zhou12}, several try to preserve information in the gradient domain. 
Smith et al. \cite{Smith08} present a hybrid (local+global) approach that relies on both an image-independent global mapping and a multiscale local contrast enhancement. 
Lau et al. \cite{Lau11} propose an approach defined as `semi-local', as it clusters pixels based on both their spatial and chromatic similarities. The color mapping problem is solved with an optimization aimed at finding optimal cluster colors such that the contrast between clusters is preserved.

\textbf{Gamut mapping} is the process of adjusting the colors of an input image into the constrained color gamut of a given device. 
Gamut mapping algorithms can be mainly divided into \emph{clipping} and \emph{compression} approaches~\cite{Morovic08}. The former ones change the source colors that fall outside of the destination gamut
(e.g. HPMINDE~\cite{CIE04,Bonnier06}); the latter also modify the in-gamut colors. 
Similarly to color-to-gray conversion, gamut mapping methods can also be categorized as global and local. 
To address metamerism in gamut mapping, local approaches~\cite{Bala01,Nakauchi99,Kimmel05} allow two spatially-distant pixels of equal color  to be mapped to different in-gamut colors.   
Global approaches, conversely, will always apply the same map to two pixels of the same color, regardless of their location. 
%
Many gamut mapping algorithms optimize some image difference criterion 
\cite{Nakauchi99,Kimmel05,Alsam09,Lau11}. 

\textbf{Color-blind viewers} cannot perceive differences between some given colors, due to the lack of one or more types of cone cells in their eyes~\cite{daltonize,Meyer88}. 
Image perception by a color-deficient observer is typically simulated by first applying a linear transformation from a standard color space such as RGB~\cite{Kim12,Brettel07,Vienot99}, XYZ~\cite{Meyer88,Rasche05b}, or CIE Lab*~\cite{Kuhn08b,Huang07} to a special LMS space, which specifies colors in terms of the relative excitations of the cones. 
Then, the color domain is reduced in accordance with the color deficiency (typically, by means of a linear transformation in the LMS space~\cite{Vienot99,Kim12,Huang07}). Finally, the reduced LMS space is mapped back to RGB. 

When trying to adapt an image for a color-blind viewer, one has to ensure that the structure of the original image is not lost due to color ambiguities. 
Kuhn et al. \cite{Kuhn08b} focus on obtaining natural images by preserving, as much as possible, the original image colors. 
Rasche et al. \cite{Rasche05b}, instead, try to maintain distance ratios during the reduction process. 
Lau et al. \cite{Lau11} is aimed at preserving both the contrast between color clusters and the reduced image colors.

\textbf{Multispectral image fusion} aims to combine a collection of images captured at different wavelengths into a single one, containing details from several spectra. 
Zhang et al. \cite{Zhang08} and Lau et al. \cite{Lau11} present a method that adaptively adjusts the contrast of photographs by using the contrast and texture information from near-infrared (NIR) image.
Kim et al. \cite{Kim11} show how to use different bands of the invisible spectrum to improve the visual quality of old documents. 
S{\"u}sstrunk and Fredembach \cite{Susstrunk10} provide a good introduction to the topic and present, as examples of image enhancements, haze removal and realistic skin smoothing.

\textbf{General approaches.} 
We should stress that despite a significant corpus of research on color transformations, most of the methods are targeted to specific applications and lack the generality of a framework that could be applied to different classes of problems.  
At the same time, there is an obvious common denominator between the aforementioned problems: for example, both color-blind transformations~\cite{Rasche05} and color-to-gray conversions~\cite{Cui10,Zhao10,Zhou12} can be regarded as  mappings to a gamuts of lower dimension~\cite{Gooch05}. 
To the best of our knowledge, only the recent work of \cite{Lau11} introduces a comprehensive approach that works with generic color transformation and easily adapts to different applications.

\textbf{Main contribution}. 
In this paper, we present {\em Laplacian colormaps}, a new generic framework for computing structure-preserving color transformations that can be applied to different problems.  
Our main motivation comes from recent works on Laplacians as structure descriptors \cite{bansal2013joint} and 
 joint diagonalization of Laplacians \cite{eynard2012multimodal,kovnatsky2013coupled,glashoff2013matrix,bronstein2013making}   
(to the best of our knowledge, our paper is the first application of these methods in the domain of image analysis).

Using Laplacians as image structure descriptors, we observe that an ideal color transformation should preserve the Laplacian eigenstructure, implying that the Laplacians of the original and color-converted image should be jointly diagonalizable. 
Employing the relation between joint diagonalizability and commutativity of matrices~\cite{glashoff2013matrix,bronstein2013making}, we use Laplacians commutativity as a criterion of image structure preservation. 
We try to find such a colormap that would produce a converted image whose Laplacian commutes as much as possible with the Laplacian of the original image. 
Since Laplacians can be defined in any colorspace, our approach is generic and applicable to any kind of color conversions (in particular, color-to-gray, gamut mapping, color-blind optimization, etc.). 
Furthermore, we can work with both global and local colormaps.

The rest of the paper is organized as follows: in Section~2, we review the main results related to joint diagonalization and commutativity of matrices. 
In Section~3 we formulate our optimization problem and discuss its numerical solution. 
Section~4 shows examples of applications of our framework to different problems involving color transformations. 
Finally, Section~5 concludes the paper. 
Technical derivations are given in the Appendix.

\begin{figure*}[ht]
\vspace{-1mm}
\centering
\includegraphics[trim=0 0 0 0]{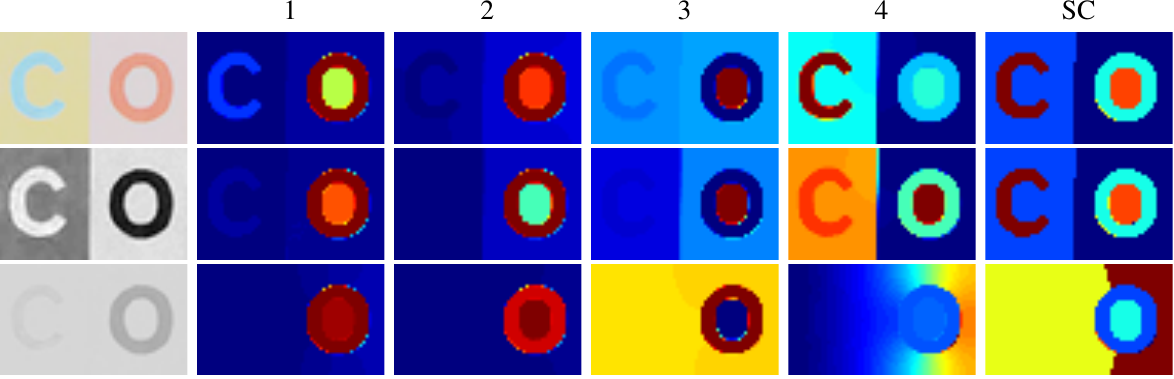}
\caption{Image structure similarity is conveyed by the eigenstructure of their Laplacians. Top: original RGB image; middle: grayscale conversion by our method; bottom: luma only conversion. Left-to-right: original image, first four eigenvectors of the corresponding Laplacian, result of spectral clustering. }
\label{fig:eigs}\vspace{-5mm}
\end{figure*}

\section{Background}

\textbf{Notation and definitions. }
We denote by $\Aa$ a matrix, by $\bb{a}$ a (column) vector, and by $a$ a scalar. 
We denote by 
\begin{equation}
\| \Aa\|_\mathrm{F}=\textstyle \left( \sum_{ij} a^2_{ij} \right)^{1/2}; \,\,
\| \bb{a} \|_2 = \textstyle \left( \sum_{i} a^2_{i} \right)^{1/2} 
\end{equation}
the Frobenius norm of the matrix $\Aa$ and the Euclidean norm of a vector $\bb{a}$, respectively. 
$\mathrm{diag}(a_1,\hdots,a_n)$ is a diagonal matrix with diagonal elements $a_1, \hdots, a_n$, $\mathrm{diag}(\Aa)$ are the diagonal elements of $\Aa$ arranged as a column vector, $\mathrm{Diag}(\Aa)$ is a diagonal matrix obtained by setting to zero the off-diagonal elements of $\Aa$, and $\mathrm{vec}(\Aa)$ is a column vector obtained by column-stacking of $\Aa$. 

Let us be given an $N\times M$ image with $d$ color channels, column-stacked into an $NM\times d$ matrix $\Xx = (\xx_1, \hdots, \xx_{NM})^\Tr$. 
The problem of color conversion is creating a new image $\Yy = \Phi(\Xx)$ with $d'$ color channels, by means of a {\em colormap} $\Phi: \RR^{NM\times d} \rightarrow \RR^{NM \times d'}$. 
In particular, we are interested in parametric colormaps $\Phi_{\bb{\theta}}$, parametrized by an  $n$-dimensional vector of parameters $\bb{\theta}$. 
In the simplest case, $\Phi_{\bb{\theta}}$ is a global color transformation applied pixel-wise, i.e., each pixel $\xx_i \in \RR^d$ of the original image is mapped by means of the same $\phi_{\bb{\theta}} : \RR^d \rightarrow \RR^{d'}$ such that $\Phi_{\bb{\theta}}(\Xx) = (\phi_{\bb{\theta}}(\xx_1), \hdots, \phi_{\bb{\theta}}(\xx_{NM}))^\Tr$ 
(a simple example is linear RGB to gray mapping, where $d=3$, $d'=1$, $n=3$ and $\phi_{\bb{\theta}}(\xx_i) = \sum_{j=1}^d \theta_i \xx_{ij}$, where in addition we require $\bb{\theta}\geq 0$ and $\sum_{i=1}^d \theta_i = 1$).

Let $\{ k_1, \hdots, k_L \} \subseteq \{1, \hdots, NM\}$ denote a subset of the image pixel indices (this subset can be the whole set of $NM$ pixels, a regularly subsampled $M/s\times N/s$ image, `representative' pixels obtained by clustering the image, etc.). 
Considering these pixels as vertices of a graph, we define edge weights ({\em adjacencies}) as a combination of spatial and `radiometric' distances, 
\begin{eqnarray}
\label{eq:laplacian1}
w_{ij} = e^{- \frac{ \delta_{ij}^2 }{2\sigma_s^2}} e^{-\frac{\| \xx_{k_i} - \xx_{k_j} \|^2_2}{2\sigma_r^2}},
\end{eqnarray}
where $\delta_{ij}$ is the spatial distance between pixels $k_i$ and $k_j$, and $\sigma_s, \sigma_r \geq 0$ are parameters (more generally, the `radiometric' part of the adjacency $w_{ij}$ does not have to work on pixel-wise colors, and one can consider some local features, the simples of which are patches \cite{wetzler2012efficient}). 
For practical computations, it is usually assumed that $w_{ij} \approx 0$ between spatially-distant pixels, so they are disconnected. 
We define the (unnormalized) {\em Laplacian}\footnote{There exist numerous ways of defining Laplacian matrices; we consider the unnormalized one merely for the sake of simplicity. The ability to work with practically any operator capturing the image structure is one of the strengths of our method. } of this graph as a symmetric positive semi-definite $L \times L$ matrix $\Ll_{\Xx} = \Dd_{\Xx} - \Ww_{\Xx}$, where $\Ww_{\Xx}$ is the adjacency matrix with elements as in~(\ref{eq:laplacian1}), and $\Dd_{\Xx} = \mathrm{diag}(\sum_{j\neq i}w_{ij})$. In the following, we refer to $\Ll_{\Xx}$ as the Laplacian of image $\Xx$.

Since $\Ll_{\Xx}$ is symmetric, it admits an orthonormal eigendecomposition by means of a matrix $\Uu$, such that $\Uu^\Tr \Ll_{\Xx} \Uu = \bb{\Lambda}_{\Xx}$, where the columns of $\Uu$ are orthonormal eigenvectors, and $\bb{\Lambda}_{\Xx} = \mathrm{diag}(\lambda^{\Xx}_1,\hdots, \lambda^{\Xx}_{L})$ are the corresponding eigenvalues, sorted in ascending order $0 = \lambda_1 \leq \lambda_2 \leq \hdots \leq \lambda_{L}$. 
For simplicity, we assume that there are no repeating eigenvalues, and thus the eigenvectors are defined up to sign. 
We say that two Laplacians $\Ll_{\Xx}$ and $\Ll_{\Yy}$ are {\em jointly diagonalizable} if they have the same eigenvectors $\Uu$, i.e., 
$\Uu^\Tr \Ll_{\Xx} \Uu = \bb{\Lambda}_{\Xx}$ and $\Uu^\Tr \Ll_{\Yy} \Uu = \bb{\Lambda}_{\Yy}$.
$\Ll_{\Xx}$ and $\Ll_{\Yy}$ are said to {\em commute} if their {\em commutator} is $[\Ll_{\Xx},\Ll_{\Yy}] = \Ll_{\Xx}\Ll_{\Yy} - \Ll_{\Yy}\Ll_{\Xx} = \bb{0}$.

\textbf{Image Laplacians as structure descriptors. }
%
Laplacians have been successfully used in image processing to guide anisotropic diffusion 
\cite{sochen1998general}. 
Shi and Malik \cite{shi2000normalized} showed that a spectral relaxation of the normalized cut criterion for image segmentation boils down to finding the first eigenvectors of an image Laplacian and performing segmentation in the low-dimensional eigensubspace. 
This approach inspired the popular spectral clustering algorithm \cite{ng2002spectral}. 
More recently, Bansal and Daniilidis \cite{bansal2013joint} used the eigenvectors of image Laplacians to perform matching of images taken in different  illumination conditions, arguing that the Laplacian acts as a self-similarity descriptor \cite{shechtman2007matching} of the image.

Applying this idea to color transformations, we can use the similarity of Laplacian eigenspaces as a criterion of structural similarity of two images. 
Figure~\ref{fig:eigs} shows three images (original RGB image and two decolorized versions thereof, a `bad' and a `good' one) and the first eigenvectors of the corresponding Laplacians. One can see that a good colormap preserves the image structure, which is manifested in the two Laplacians having similar eigenvectors (first and third rows). In particular, if one applies spectral clustering to such images, the resulting segmentation will be similar.  
Thus, an ideal color transformation $\Phi(\Xx)$ should make the corresponding Laplacians $\Ll_{\Xx}$ and $\Ll_{\Phi(\Xx)}$ {\em jointly diagonalizable}.

\textbf{Joint approximate diagonalization (JAD) } 
is a way to enforce two matrices to have the same eigenstructure. 
Given two matrices $\Aa$ and $\Bb$,  
one seeks a joint approximate eigenbasis $\hat{\Uu}$ such that 
$\hat{\Uu}^\Tr \Aa \hat{\Uu}$ and $\hat{\Uu}^\Tr \Bb \hat{\Uu}$ are approximately diagonal, 
%
\begin{eqnarray}
\label{eq:jad}
J(\Aa,\Bb ) = \min_{ \hat{\Uu} }  \,\, \mathrm{off}( \hat{\Uu}^\Tr \Aa \hat{\Uu} ) + \mathrm{off}( \hat{\Uu}^\Tr \Bb \hat{\Uu} ) 
\,\,\, \, \mathrm{s.t.} \,\,\, \, 
 \hat{\Uu}^\Tr  \hat{\Uu}  = \Ii,  
\end{eqnarray}
where $\mathrm{off}(\bb{A}) = \sum_{i\neq j} a_{ij}^2$. 
%
%
%
JAD has been recently applied to jointly diagonalize Laplacian matrices in order to find compatible Fourier bases on graphs \cite{eynard2012multimodal} and surfaces \cite{kovnatsky2013coupled}. 
The drawback of this formulation is that {\em both} matrices are assumed to be given, while in our problem only one matrix (the original image Laplacian, $\Ll_{\Xx}$) is given, while the other (the transformed image Laplacian, $\Ll_{\Yy}$) has to be found. 

\textbf{Closest commuting operators (CCO). } 
Joint diagonalizability is intimately related to matrix commutativity. It is well-known that $\Aa$ and $\Bb$ are jointly diagonalizable iff they commute, i.e., $[\Aa,\Bb] = \bb{0}$ \cite{horn1990matrix}. 
It appears that this relation also holds for almost-commuting matrices, 
in the following sense: 
\begin{theo}[Glashoff-Bronstein \cite{glashoff2013matrix}]
Let $\Aa, \Bb$ be two $N \times N$ symmetric matrices normalized such that $\| \Aa \|_\mathrm{F} = \| \Bb \|_\mathrm{F} = 1$. Then, 
$$
\delta_1(\| [\Aa, \Bb] \|_\mathrm{F}) \leq J( \Aa,\Bb ) \leq \delta_2(\| [\Aa, \Bb] \|_\mathrm{F})
$$
where $\delta_1(x), \delta_2(x)$ are functions satisfying $\lim_{x\rightarrow 0}\delta_i(x) = 0$;  
\end{theo}
or in other words, almost commuting matrices are almost jointly diagonalizable.

Bronstein et al. \cite{bronstein2013making} studied an alternative problem of finding the closest commuting matrices $\tilde{\Aa}, \tilde{\Bb}$ to the given $\Aa$ and $\Bb$,  
%
\begin{equation}
\label{eq:cco}
C(\Aa,\Bb) = \min_{ \tilde{\Aa}, \tilde{\Bb} } \hspace{0mm}\| \tilde{\Aa} - \Aa \|_\mathrm{F}^2 + \| \tilde{\Bb} - \Bb \|_\mathrm{F}^2 \,\,\,\,\,\,\,\,\, \mathrm{s.t.}  \,\,
\tilde{\Aa} \tilde{\Bb} = \tilde{\Bb} \tilde{\Aa}%
\end{equation}
Since $\tilde{\Aa}, \tilde{\Bb}$ commute, they are jointly diagonalizable. 
Furthermore, if $\Aa, \Bb$ approximately commute, $C(\Aa,\Bb)$ is guaranteed to be small, i.e., almost-commuting matrices are close to commuting ones \cite{Huang_Lin}.

Somewhat surprisingly, it turns out that the JAD and CCO problems are equivalent, in the following sense: 
\begin{theo}[Bronstein et al. \cite{bronstein2013making}]
Let $\Aa, \Bb$ be symmetric matrices, 
$\hat{\Uu}$ be the minimizer of the JAD problem~(\ref{eq:jad}), and 
$\tilde{\Aa}, \tilde{\Bb}$  
be the minimizers of the CCO problem~(\ref{eq:cco}), jointly diagonalized by $\tilde{\Uu}$. 
Then:
\begin{enumerate}
\item $C(\Aa,\Bb) = J(\Aa,\Bb)$; 
\item $\tilde{\Uu} = \hat{\Uu}$; 
\item $\tilde{\Aa} = \hat{\Uu} \mathrm{Diag}(\hat{\Uu}^\Tr \Aa \hat{\Uu}) \hat{\Uu}^\Tr$ and $\tilde{\Bb} = \hat{\Uu} \mathrm{Diag}(\hat{\Uu}^\Tr \Bb \hat{\Uu}) \hat{\Uu}^\Tr$. 
\end{enumerate}
\end{theo}

Despite being equivalent, JAD and CCO problem have a key difference: in the former, optimization is performed w.r.t. the joint eigenbasis, while in the latter, optimization is performed w.r.t. closest commuting matrices. 
In our problem, the CCO formulation allows to optimize w.r.t. Laplacians, which, in turn, can be parametrized through the colormap.

Let us summarize the main results of this section, which will motivate our approach described in the following. 
First, Laplacians can be used as structural descriptors of images. 
Second, two images having similar structures translates into having the corresponding Laplacians jointly diagonalizable. 
Third, joint diagonalizability is equivalent to commutativity.

The key idea of this paper is to find such a colormap $\Phi(\Xx)$
that the Laplacian $\Ll_{\Xx}$ of the input image and the Laplacian $\Ll_{\Phi(\Xx)}$ of the output image commute as much as possible.  
Due to the relation between approximate commutativity and joint diagonalizability, it will imply that $\Ll_{\Xx}$ and $\Ll_{\Phi(\Xx)}$ have similar eigenvectors, and thus the underlying images are structurally similar.

\section{Laplacian colormaps}

\textbf{Problem formulation. }
Let $\Xx$ be a given $NM\times d$ original image and $\Phi_{\bb{\theta}}(\Xx)$ be the desired color-converted $NM\times d'$ image. 
Our goal is to find a set of parameters $\bb{\theta}$ such that the structures of the images $\Xx$ and $\Phi_{\bb{\theta}}(\Xx)$ are as similar as possible, where the similarity is judged by the commutativity of the corresponding Laplacians. 
This leads us to a class of optimization problems of the form 
%
\begin{eqnarray}
\label{eq:costfunction1}
\min_{ \theta \in \RR^n} && \hspace{0mm}
\mu_0 \|[\Ll_{\Xx}, \Ll_{\Phi_{\bb{\theta}}(\Xx)} ] \|_\mathrm{F}^2 + \mu_1\|\Ll_{\Xx} - \Ll_{\Phi_{\bb{\theta}}(\Xx)} \|_\mathrm{F}^2 \nonumber \\
&& + \mu_2 \|\bb{\theta}-\bb{\theta}_0\|_2^2 + \mu_3 \|\Phi_{\bb \theta}(\Xx_\mathrm{c}) -\Yy_\mathrm{c}\|_\mathrm{F}^2 \\ \nonumber 
&& \mathrm{s.t.} \,\,\,\,\, \text{constraints on } \bb{\theta}.
\end{eqnarray}
One can easily recognize in problem~(\ref{eq:costfunction1}) a parametric version of the CCO problem~(\ref{eq:cco}) with one of the Laplacians fixed. 
Note that the Laplacian $\Ll_{\Phi_\theta(X)}$ is parametrized by a small number of degrees of freedom $n \ll L$, and thus it would be usually impossible to make it exactly commute with the given $\Ll_X$ - hence, unlike the CCO problem, the commutator norm appears as a penalty rather than a constraint.

Additional regularization (third and fourth terms in~(\ref{eq:costfunction1})) is used if we have some `nominal' parameters $\bb{\theta}_0$ representing a standard color transformation, or if some colors $\Xx_\mathrm{c} = (\xx_1, \hdots, \xx_p)^\Tr$ should be mapped into some $\Yy_\mathrm{c} = (\yy_1,\hdots, \yy_p)^\Tr$ known in advance (for example, in some cases it is important to preserve black and white colors). 
Finally, depending on the type of the colormap $\Phi_{\bb{\theta}}$, one may impose some constraints on the parameters $\bb{\theta}$ (e.g., in linear RGB-to-gray conversion, $\bb{\theta} \geq \bb{0}$ and $\bb{\theta}^\Tr\bb{1} = 1$). 



\textbf{Local maps. }
Our approach imposes no limitations on the complexity of the colormap $\Phi_{\bb{\theta}}$; in particular, this map does not have to be global. 
Let us assume that the source image is partitioned into $q$ (soft) regions, represented by weight vectors $\bb{w}_1, \hdots, \bb{w}_q$ of size $NM\times 1$, such that $\sum_{i=1}^q \bb{w}_i = \bb{1}$ and $\bb{w}_i \geq \bb{0}$. 
In each region $i$, we allow for a different colormap $\Phi_{\bb{\theta}_i}$. 
Then, the overall colormap is given as 
$
\Phi_{\bb{\theta}}(\Xx) = \sum_{i=1}^q \Phi_{\bb{\theta}_i}(\Xx), 
$
parametrized by $\bb{\theta} = (\bb{\theta}_1,\hdots, \bb{\theta}_q)$. 
 Optimization w.r.t. to the parameters of the local colormap is performed in exactly the same manner as described above.

\textbf{Multiple Laplacians. } 
 In some applications like multispectral image fusion, one may wish to impose structural similarity between the output image and {\em multiple} images, $\Xx_1, \hdots, \Xx_K$ with colorspaces of dimensionality $d_1, \hdots, d_K$. The input image $\Xx$ may be one of the $K$ images or a merged image with $\sum_{k=1}^K d_k$-dimensional colorspace. 
 In this case, our optimization problem~(\ref{eq:costfunction1}) assumes the form 
 \begin{eqnarray}
\label{eq:costfunction3}
\min_{ \theta \in \RR^n} && \hspace{0mm}
\sum_{k=1}^K \mu_{0k}
\|[\Ll_{\Xx_k}, \Ll_{\Phi_{\bb{\theta}}(\Xx)} ] \|_\mathrm{F}^2 
+  \mu_{1k} \|\Ll_{\Xx_k} - \Ll_{\Phi_{\bb{\theta}}(\Xx)} \|_\mathrm{F}^2 \nonumber \\
&& + \mu_2 \|\bb{\theta}-\bb{\theta}_0\|_2^2 + \mu_3 \|\Phi_{\bb \theta}(\Xx_\mathrm{c}) -\Yy_\mathrm{c}\|_\mathrm{F}^2 \\ \nonumber 
&& \mathrm{s.t.} \,\,\,\,\, \text{constraints on } \bb{\theta},
\end{eqnarray}
where $\mu_{01},\hdots,\mu_{0K}, \mu_{11},\hdots, \mu_{1K}, \mu_2, \mu_3 \geq 0$ are constants determining the tradeoff between different penalties.

\section{Results and Applications}
\label{sec:res}

In this section, we show several applications of our approach for decolorization, image optimization for color-blind people, gamut mapping, and multispectral image fusion, providing extensive comparison to previous works. 
%
As a quantitative criterion of the colormap quality, we use the {\em root weighted 
mean square} (RWMS) error proposed by \cite{Kuhn08}, measuring the distortion of relative color distances in two images, 
\begin{equation}
\epsilon_i = \left( \frac{1}{NM} \sum_{j =1}^{NM} \frac{( R_{\Yy} \|\xx_i - \xx_j \|  - R_{\Xx} \| \yy_i - \yy_j \|)^2}{ R^2_{\Yy} \|\xx_i - \xx_j \|^2 } \right)^{1/2}, 
\end{equation}
where $N\times M$ is the image size, $\xx_i \in \RR^d$ and $\yy_i \in \RR^{d'}$ denote the $i$th pixel of the input and output images, respectively, and $R_{\Xx} = \max_{ij} \|\xx_i - \xx_j \|$ is the color range of image $\Xx$. 
Plotting the pixel-wise RWMS error $\epsilon_i$ as an image allows to see which pixels are most affected by the color transformation. 
The average $\frac{1}{NM}\sum_{i=1}^{NM}\epsilon_i$ is used as a single number representing the quality of the colormap.

All experiments share a common setup: first of all, RGB values are scaled by 255. Then we calculate a weighted adjacency matrix according to~(\ref{eq:laplacian1}) using all pixels ($L=MN$) if the images are small enough, and resizing the image to have long side of 300 pixels otherwise. We used fixed 4-neighbors connectivity and parameters $\sigma_r=1$, $\sigma_s=0$. Default weights for the cost function are $\mu_1 = 1, \mu_2 = 1, \mu_3 = 0$, and regularization term $\bb{\theta}_0 = \bb{0}$. Parameters are initialized randomly and normalized to satisfy the condition $\bb{\theta}^\Tr\bb{1} = 1$.

 As a last step, since mapping might produce color values out of the $[0, 1]$ range, output channels are normalized.
Optimization was implemented in MATLAB, using interior-point method from the Optimization Toolbox.

\textbf{Decolorization. }
For RGB-to-gray mapping, we used a global colormap, applying in each pixel $\xx_i$ the following transformation:
$
y_i = \alpha + \sum_{j=1}^3 \beta_i x_{ij}^{\gamma_i},
$
where $x_{ij}$ is the $j$th RGB channel of the $i$th pixel, $y_i$ is the grayscale output, and $\bb{\theta} = (\alpha, \beta_1, \gamma_1, \hdots, \beta_3, \gamma_3)$ 
are the colormap parameters w.r.t. which the optimization is performed.

Images used for this experiment were taken from \cite{Cadik08}.
Figure \ref{fig:rgb2gray} shows the results of our transformations, compared to previous works~\cite{Gooch05,Rasche05,Grundland07,Neumann07,Smith08}. Results were evaluated using two different 
metrics: quantitative (RWMS) and qualitative perceptual evaluation following \cite{Cadik08}. 
In the perceptual evaluation conducted through a Web survey, 107 volunteers were shown the original RGB image together with a pair of its gray conversions, and were asked which of the two results better preserved the original image. 
Then, we used Thurstone's law of comparative judgments to convert the 1857 pairwise evaluations into interval {\em z-score} scales~\cite{Thurstone27,Tsukida11}. 
Table~\ref{table:color2gray} provides average RWMS values and z-scores calculated on an 8-images subset of \v{C}adik's.
Our approach performs the best w.r.t. both criteria.

\begin{table*}[htdp]
\begin{center}
\begin{tabular}{rccccccc}
& CIE Y & \cite{Gooch05} & \cite{Rasche05} & \cite{Grundland07} & \cite{Neumann07} & \cite{Smith08} & \textbf{Laplacian}\\ 
\hline
{\em RWMS}     & 2.86   & 2.31   & 2.49   & 2.21   & 4.91   & 2.94   & \textbf{1.42}  \\
{\em z-score}  & -0.14 & -0.24 & -0.55 & 0.63 & -0.45 & -0.06 & \textbf{0.81} \\
\hline
\end{tabular}
\end{center}
\caption{Comparison of color-to-gray conversions in terms of mean RWMS value and z-score, averaged on all images.}
\label{table:color2gray}
\end{table*}

\begin{figure}[ht]
\centering
\begin{minipage}[b]{0.32\linewidth}
\centering \small Original image
\includegraphics[width=\linewidth,trim=0 0 0 0,clip=true]{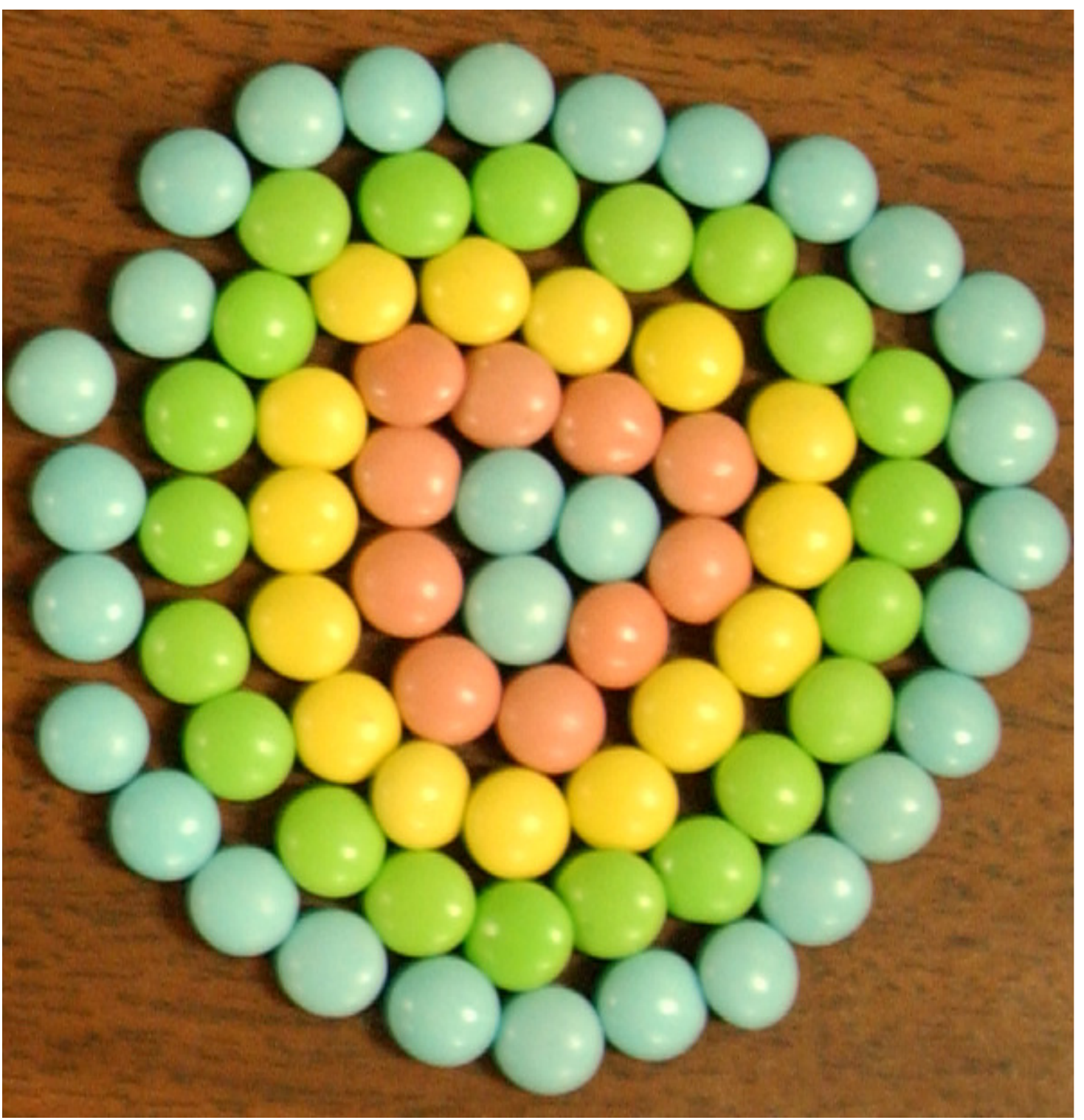}\\
\end{minipage}
\begin{minipage}[b]{0.32\linewidth}
\centering \small Luma
\includegraphics[width=\linewidth,trim=0 0 0 0,clip=true]{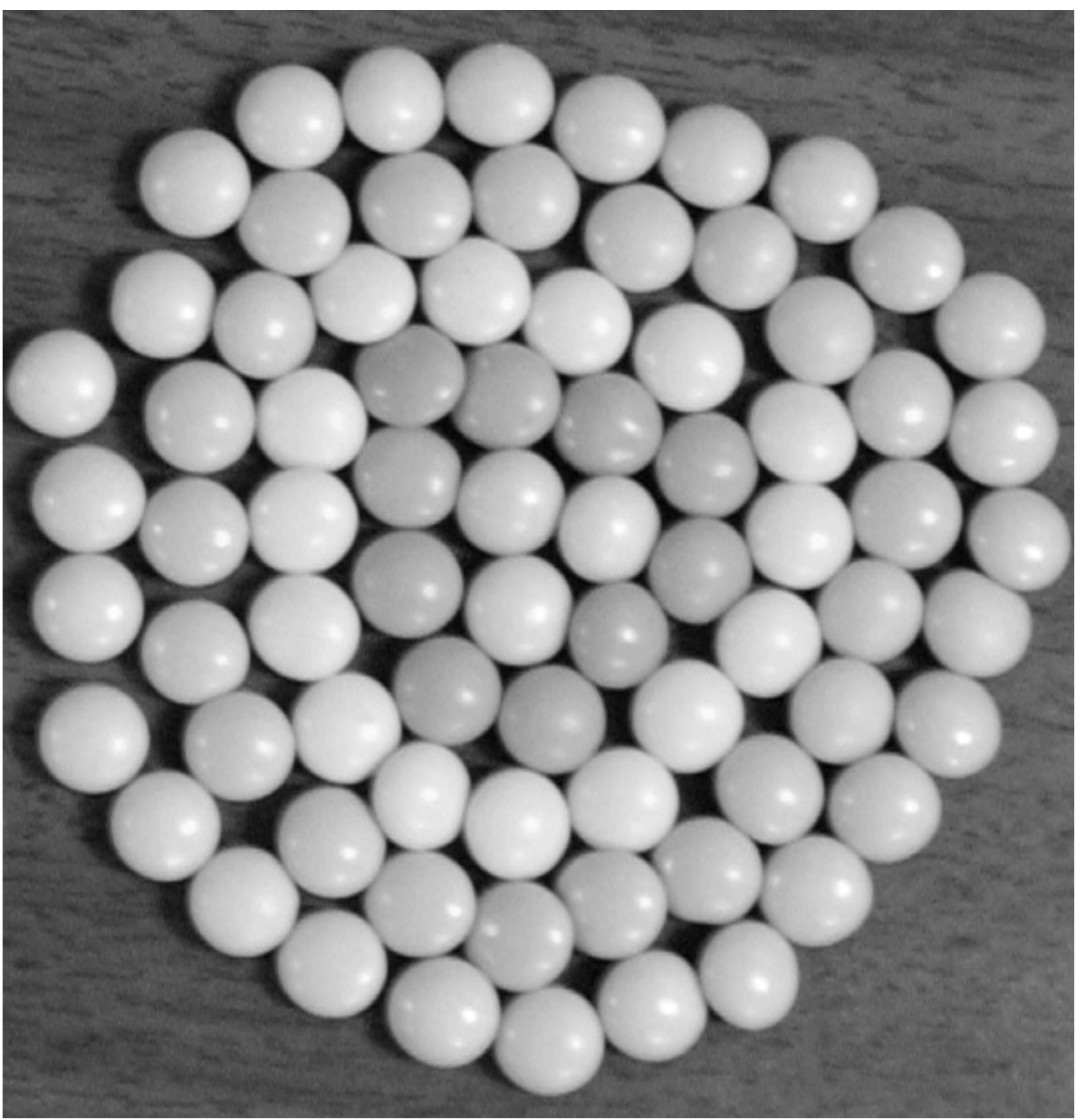}\\
\end{minipage}
\begin{minipage}[b]{0.32\linewidth}
\centering \small \cite{Lau11}
\includegraphics[width=\linewidth,trim=0 0 0 0,clip=true]{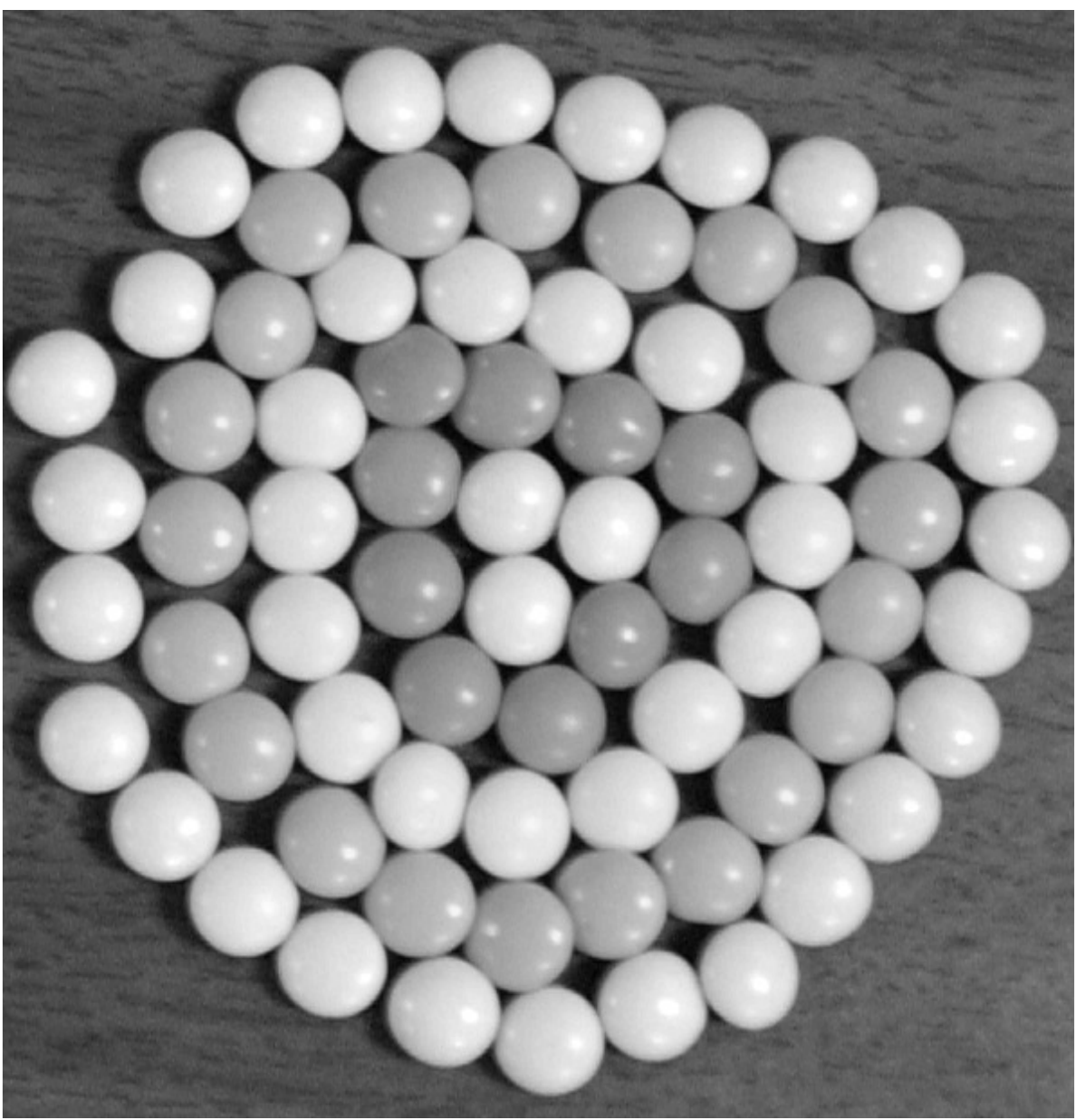}\\
\end{minipage}
\\
\begin{minipage}[b]{0.32\linewidth}
\centering \small \textbf{Laplacian (global) }
\includegraphics[width=\linewidth,trim=0 0 0 0,clip=true]{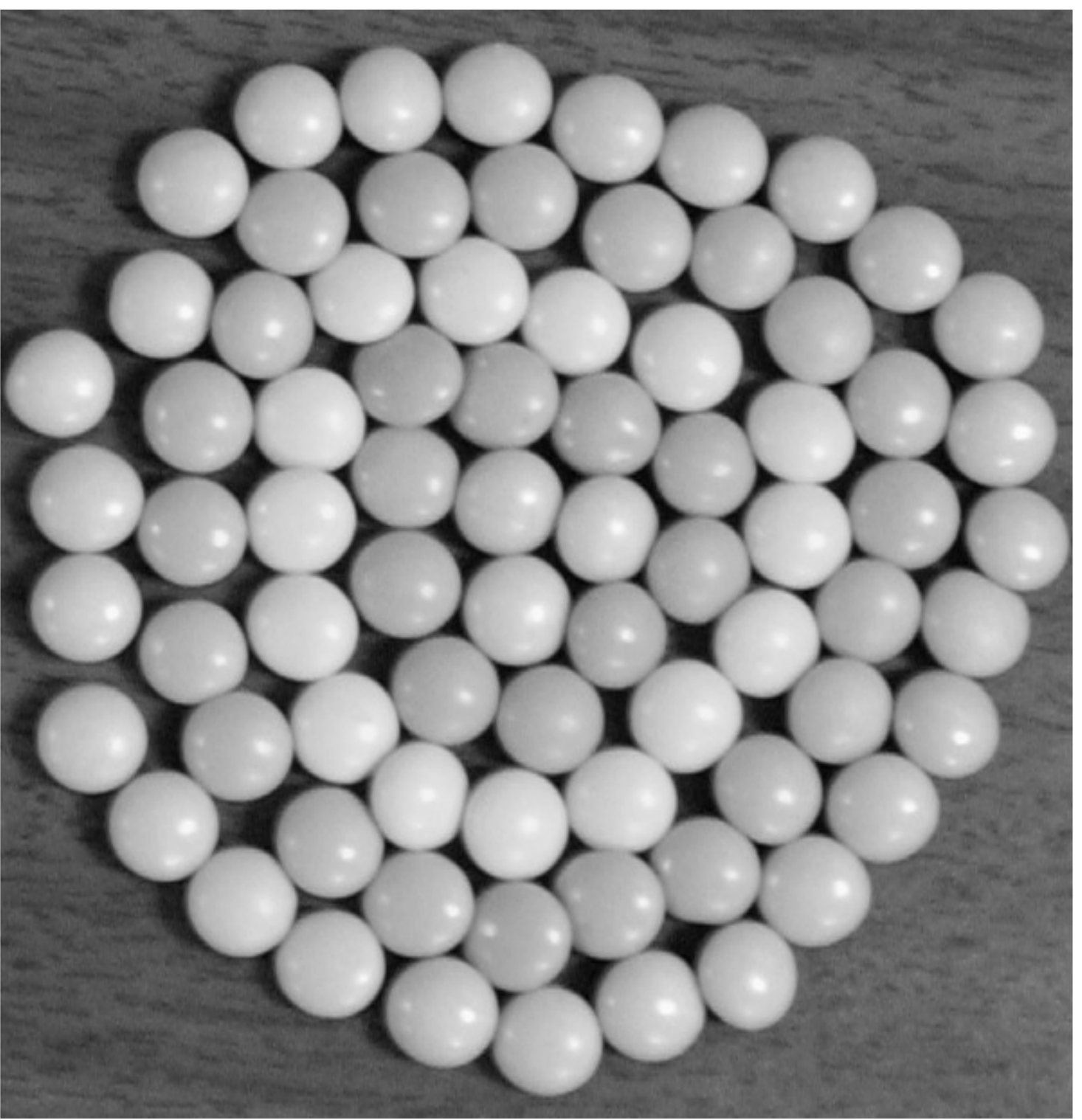}\\
\end{minipage}
\begin{minipage}[b]{0.32\linewidth}
\centering \small \textbf{ Laplacian (local) }
\includegraphics[width=\linewidth,trim=0 0 0 0,clip=true]{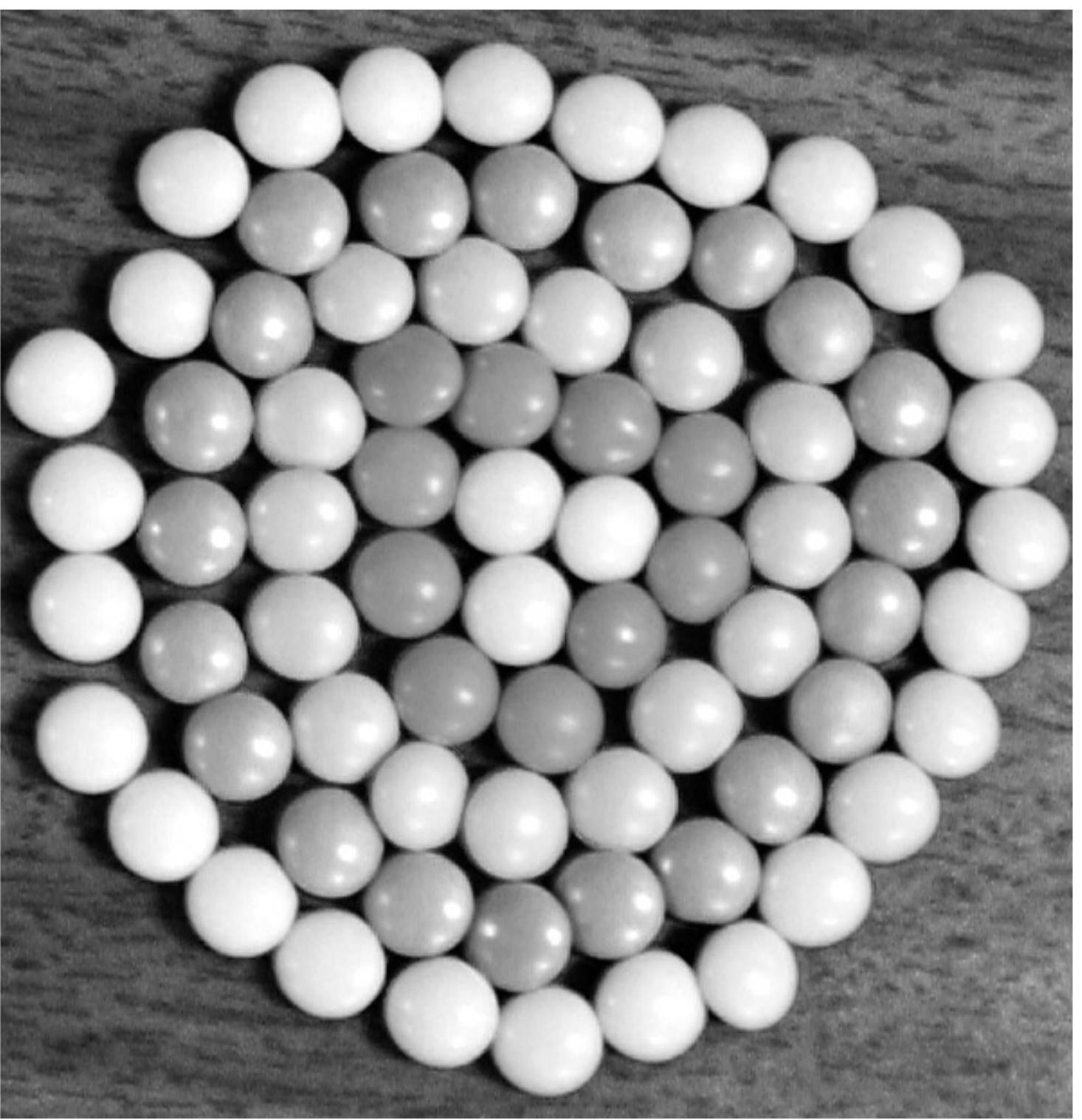}\\
\end{minipage}
\begin{minipage}[b]{0.32\linewidth}
\centering \small Clusters
\includegraphics[width=\linewidth,trim=0 0 0 0,clip=true]{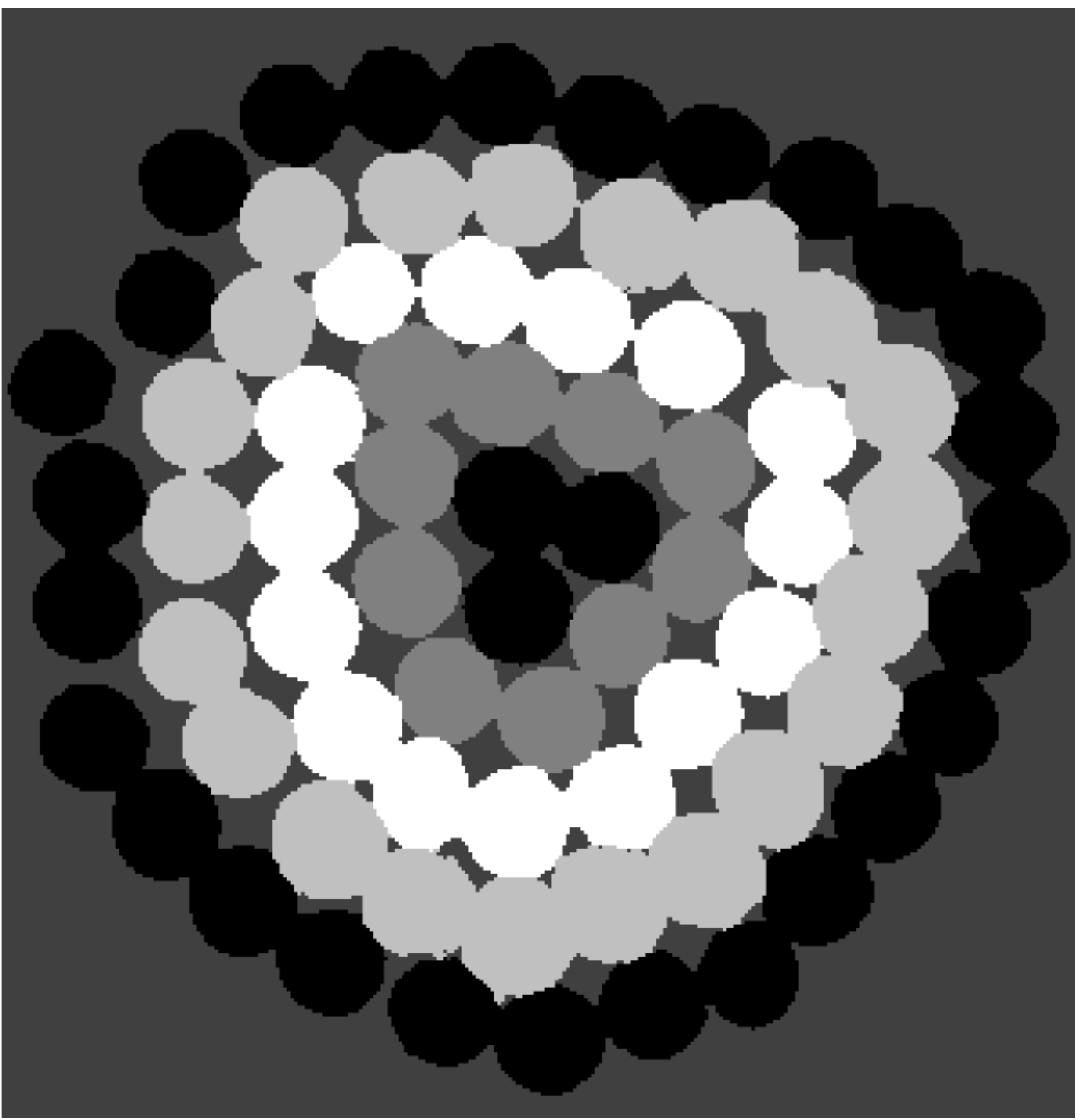}\\
\end{minipage}
\caption{Global vs Local maps results. Top row, left-to-right: original image, Luma, result by Lau et al. \cite{Lau11}.
Bottom row: Laplacian colormaps using a global (left) and local (middle) map; the spatial weights used in the latter.}
\label{fig:candies}
\vspace{-5mm}
\end{figure}

\textbf{Color-blind viewers. }
We model the color distortion of an RGB image $\Xx$ as perceived by a color-blind person by means of a map 
$\Psi: \RR^{NM\times d} \rightarrow \RR^{NM\times d}$. 
Since $\Psi$ is given and beyond our control, we try to `pre-transform' the original image by means of $\Phi_{\bb{\theta}}: \RR^{NM\times d} \rightarrow \RR^{NM\times d}$ in such a way that the image $(\Phi_{\bb{\theta}} \circ \Psi )(\Xx)$ that appears to the color-blind person has the structure of the original image $\Xx$. 
We extend our problem formulation so that the transformed image maintains its structure both \emph{when seen by a color-blind observer} 
and \emph{when seen by a regular observer}. 
In our optimization problem, this translates into requiring the two pairs of Laplacians 
$\Ll_{\Xx}, \Ll_{(\Phi_{\bb{\theta}}\circ \Psi)(\Xx)} $ and 
$\Ll_{\Xx}, \Ll_{\Phi_{\bb{\theta}}(\Xx)} $ to commute. 
The cost function is similar to the multiple Laplacians setting~(\ref{eq:costfunction3}): 
%
\begin{eqnarray}
\label{eq:costfunction2}
\min_{ \theta \in \RR^n} && \hspace{0mm}
\mu_{01}\| [ \Ll_{\Xx},  \Ll_{(\Phi_{\bb{\theta}} \circ \Psi )(\Xx)} ] \|_\mathrm{F}^2   
+ \mu_{02}\| [ \Ll_{\Xx},  \Ll_{\Phi_{\bb{\theta}}(\Xx)} ]\|_\mathrm{F}^2 \nonumber \\
&& + \mu_{11}\|\Ll_{\Xx} - \Ll_{(\Phi_{\bb{\theta}} \circ \Psi )(\Xx)} \|_\mathrm{F}^2 + \mu_{12}\|\Ll_{\Xx} - \Ll_{\Phi_{\bb{\theta}}(\Xx)} \|^2_\mathrm{F} \nonumber \\
&& + \mu_2 \|\bb{\theta}-\bb{\theta}_0\|_2^2  
\end{eqnarray}

\begin{figure*}[ht]
\begin{minipage}[b]{0.13\linewidth}
\centering \small Original image
\end{minipage}
\begin{minipage}[b]{0.26\linewidth}
\centering \small Color-blind 
\end{minipage}
\begin{minipage}[b]{0.26\linewidth}
\centering \small \cite{Lau11}
\end{minipage}
\begin{minipage}[b]{0.26\linewidth}
\centering \small \textbf{Laplacian}
\end{minipage}
\begin{minipage}[b]{0.05\linewidth}
\end{minipage}
\\
\begin{minipage}[b]{0.13\linewidth}
\includegraphics[width=\linewidth,trim=0 0 0 0,clip=true]{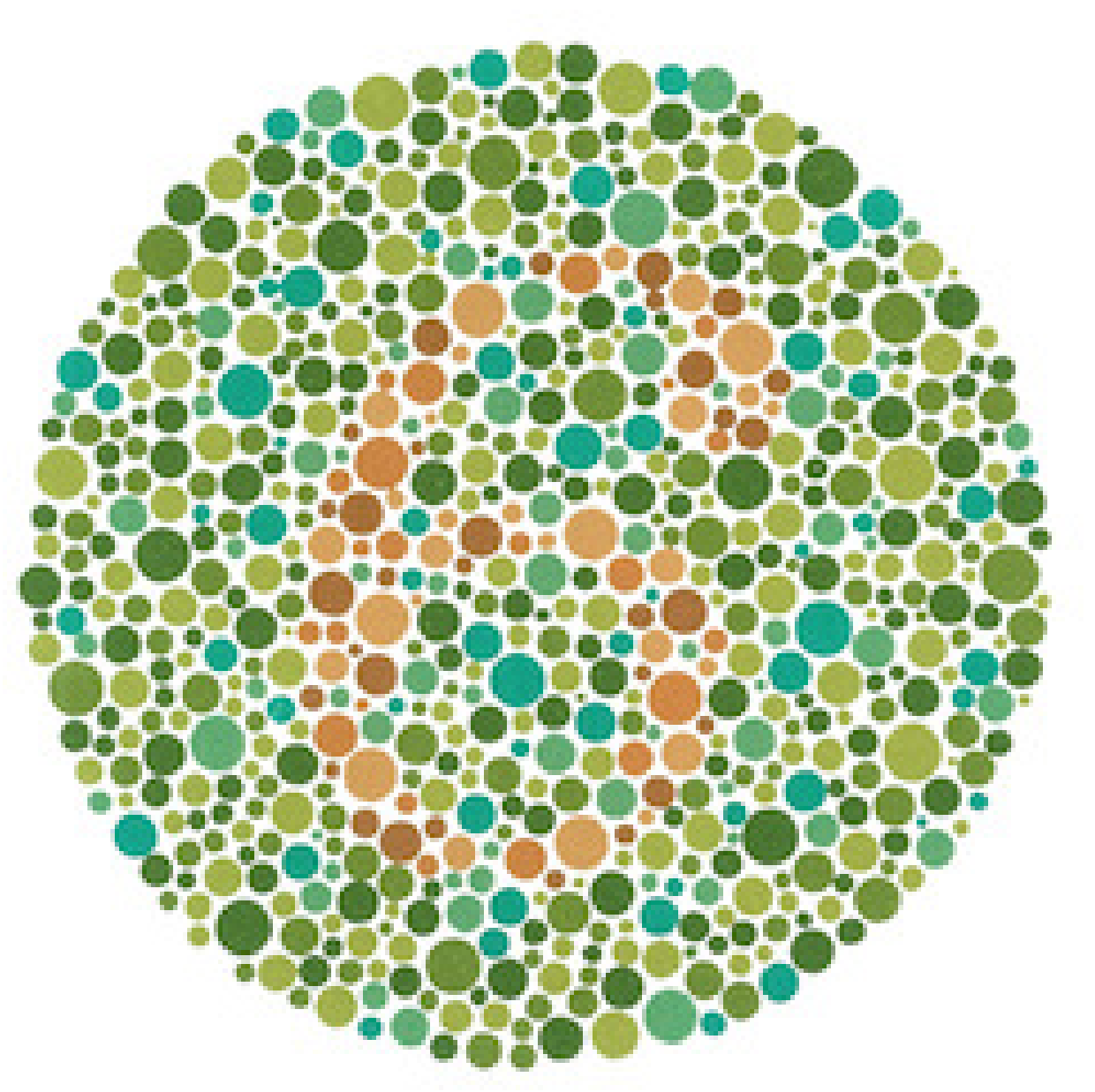}\\
\end{minipage}
\begin{minipage}[b]{0.13\linewidth}
\includegraphics[width=\linewidth,trim=0 0 0 0,clip=true]{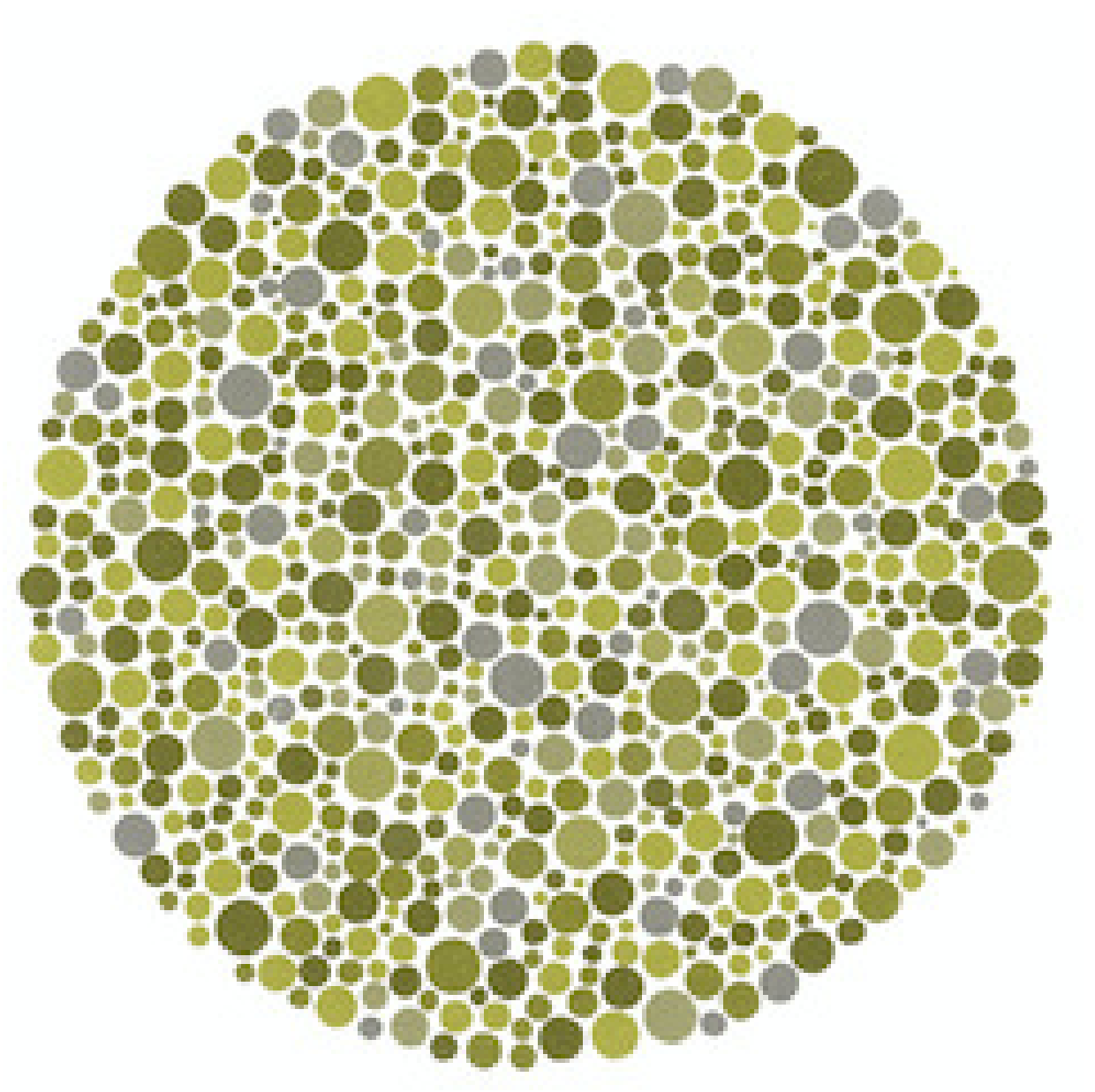}\\
\end{minipage}
\begin{minipage}[b]{0.13\linewidth}
\includegraphics[width=\linewidth,trim=0 10 65 20,clip=true]{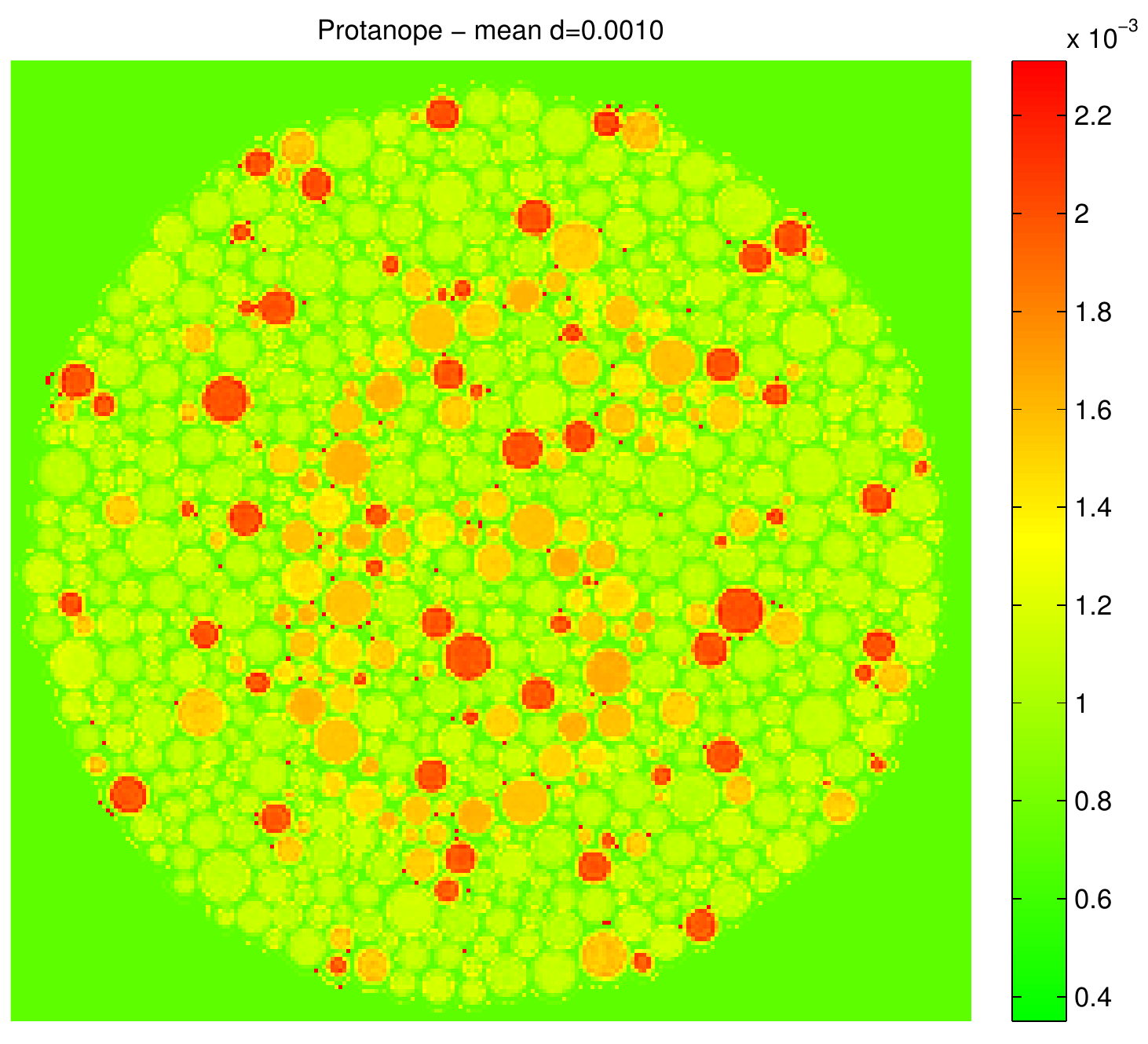}\\
\centering \small 0.98
\end{minipage}
\begin{minipage}[b]{0.13\linewidth}
\includegraphics[width=\linewidth,trim=0 0 0 0,clip=true]{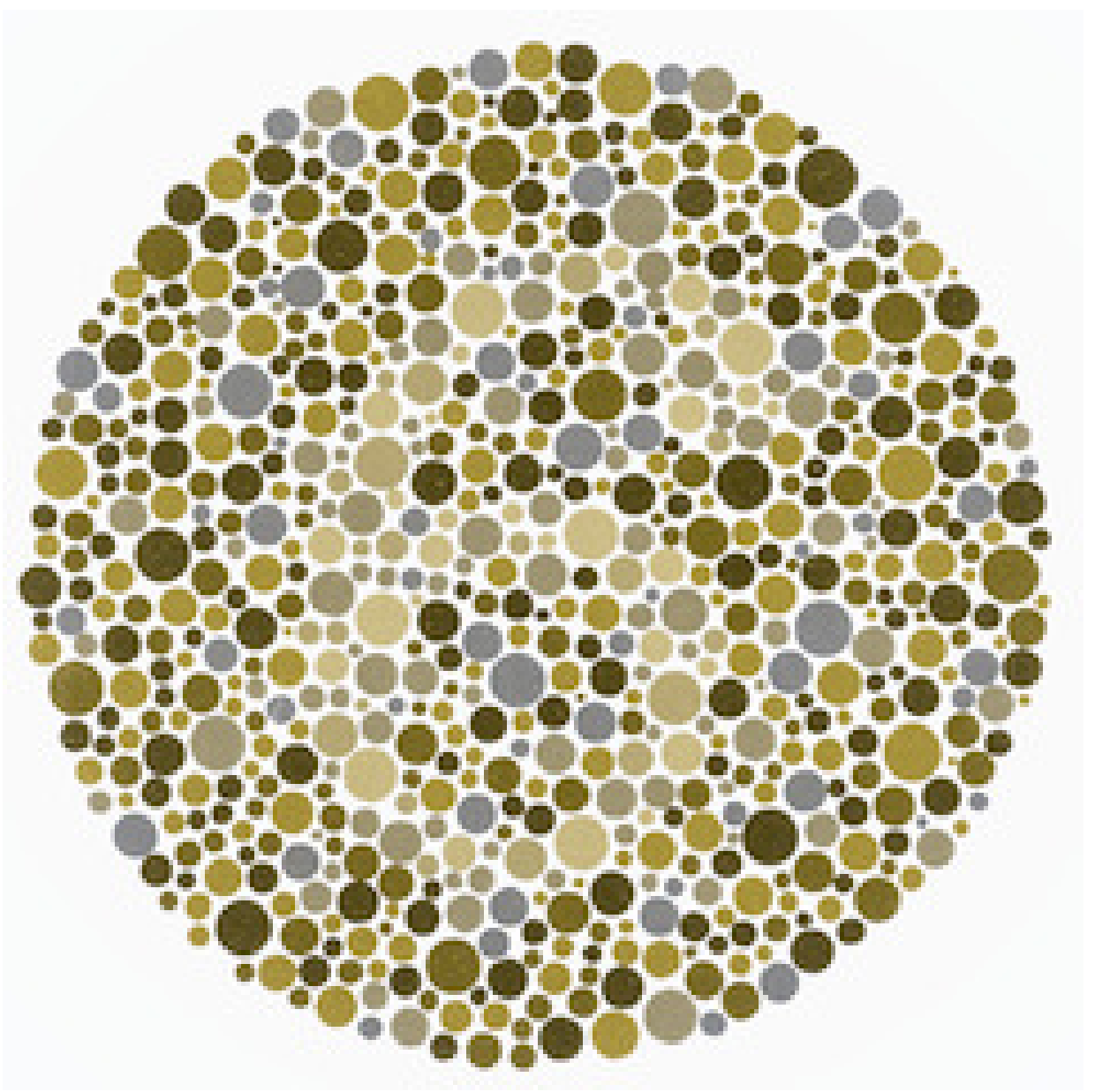}\\
\end{minipage}
\begin{minipage}[b]{0.13\linewidth}
\includegraphics[width=\linewidth,trim=0 10 65 20,clip=true]{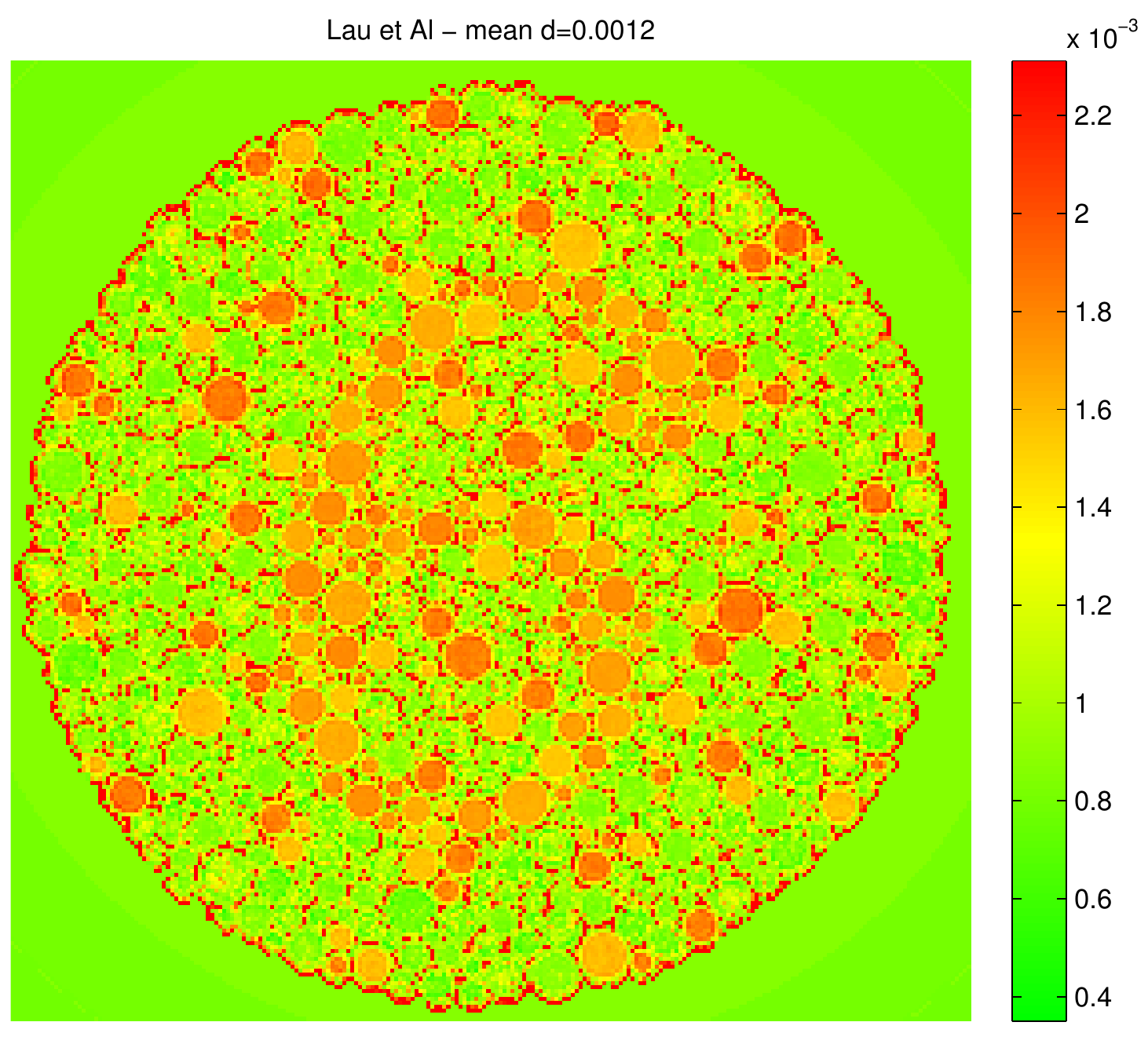}\\
\centering \small 1.23
\end{minipage}
\begin{minipage}[b]{0.13\linewidth}
\includegraphics[width=\linewidth,trim=0 0 0 0,clip=true]{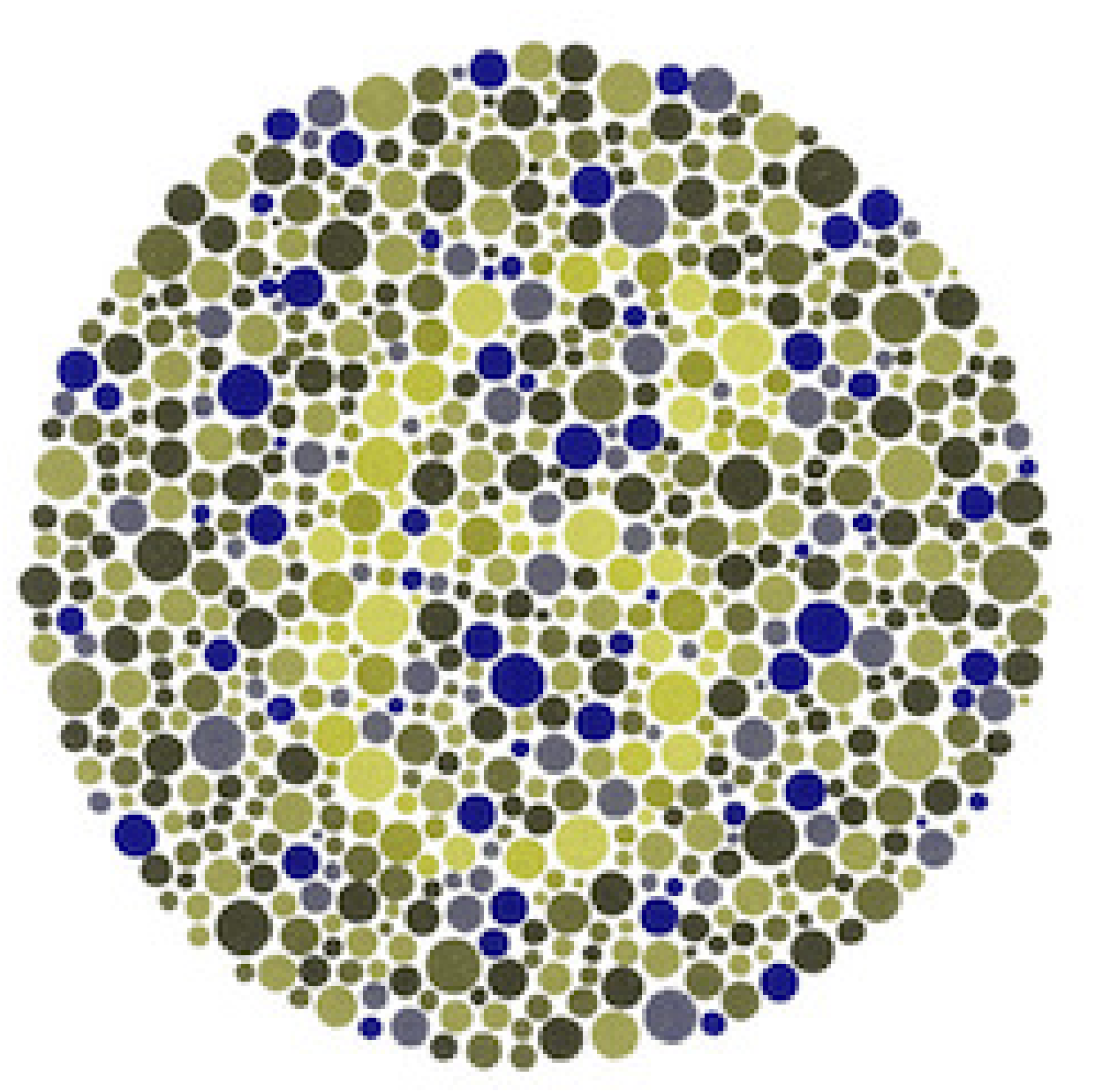}\\
\end{minipage}
\begin{minipage}[b]{0.13\linewidth}
\includegraphics[width=\linewidth,trim=0 10 65 20,clip=true]{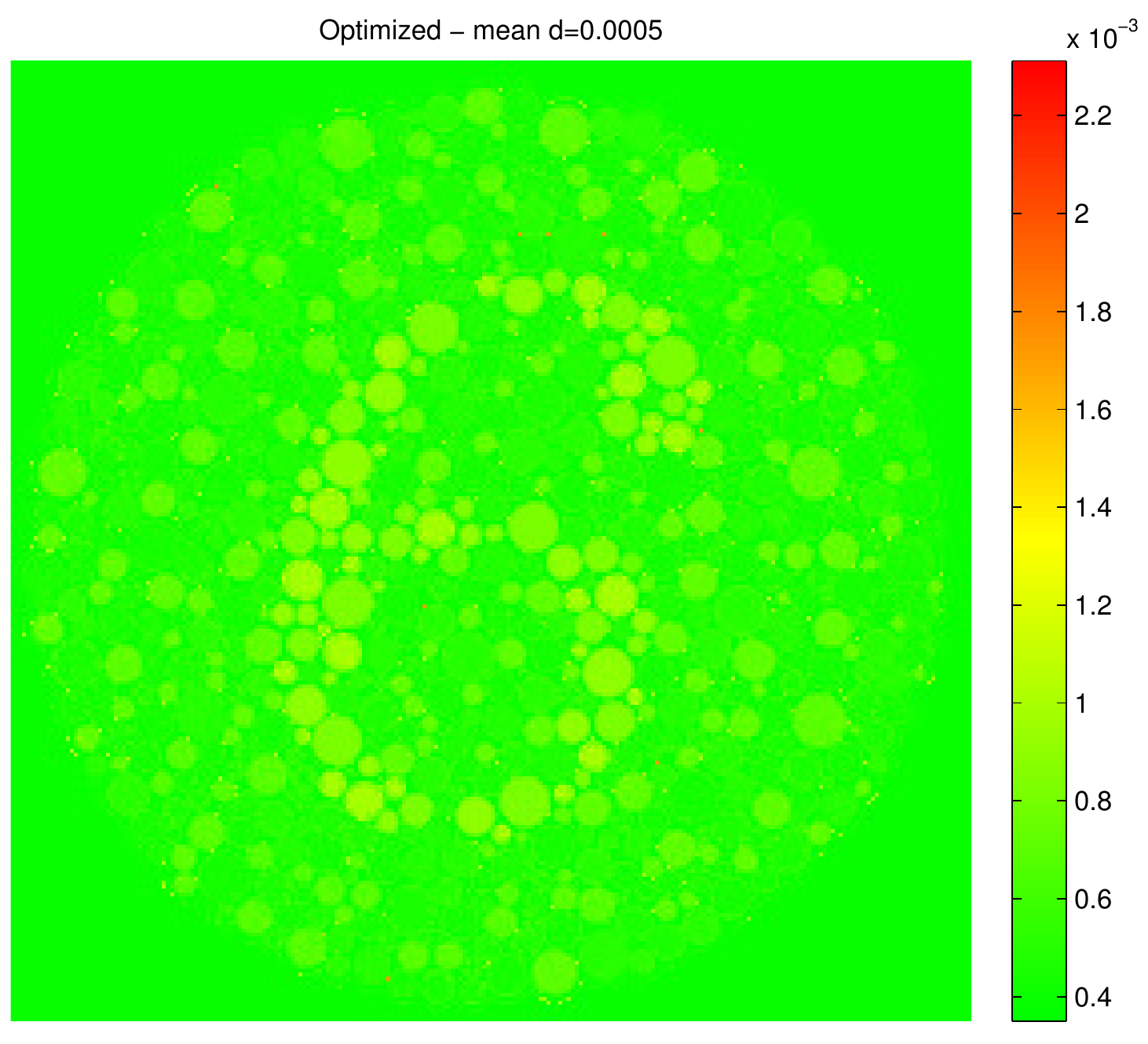}\\
\centering \small 0.50
\end{minipage}
\begin{minipage}[b]{0.05\linewidth}
\includegraphics[width=4.8mm,height=4.5mm,trim=0 0 0 0,clip=true]{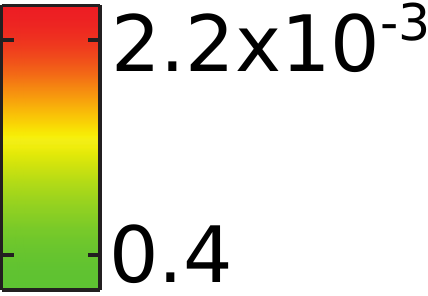}\\
\end{minipage}
\vspace{3mm}\\
\begin{minipage}[b]{0.13\linewidth}
\includegraphics[width=\linewidth,trim=0 0 0 0,clip=true]{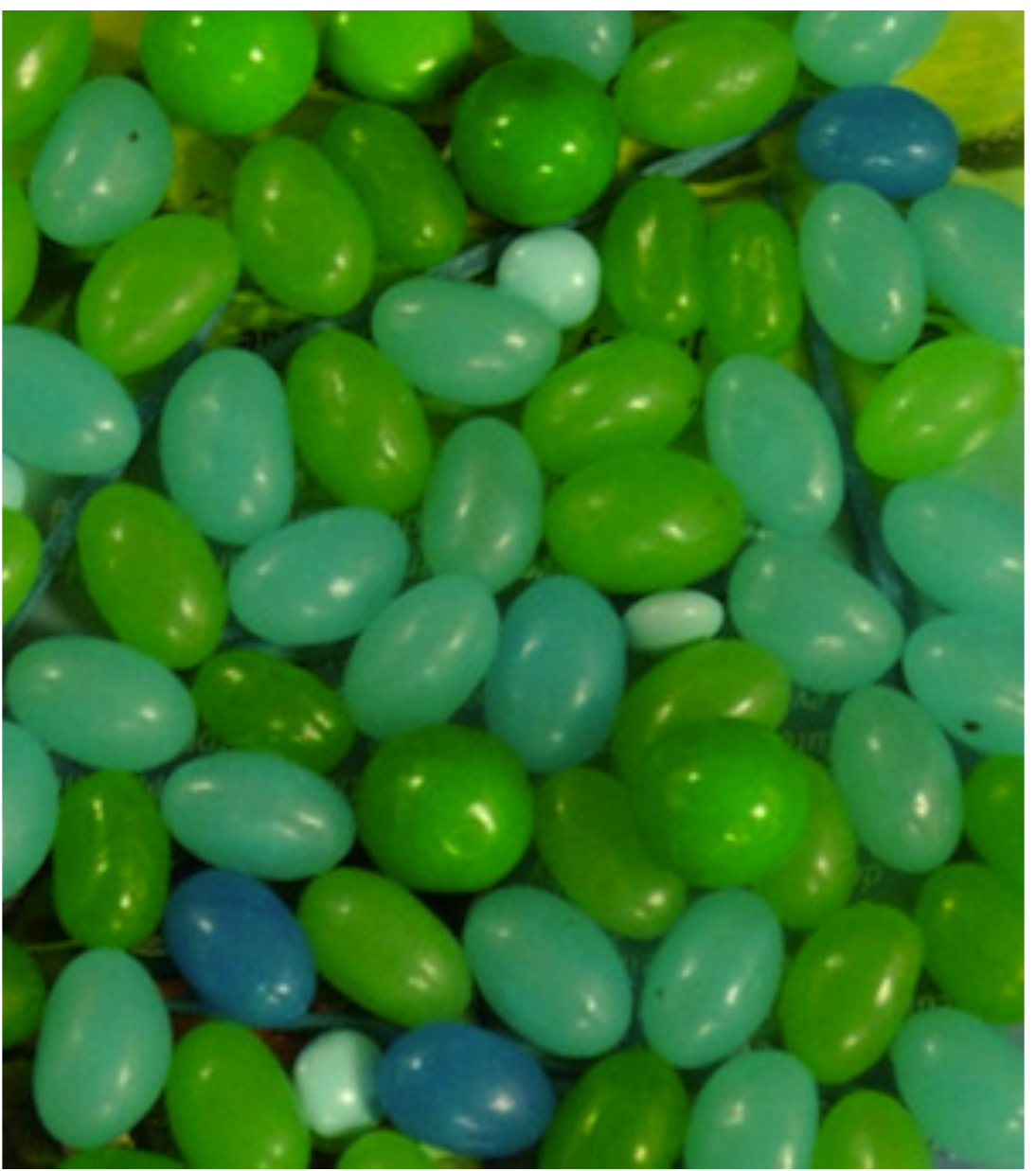}\\
\end{minipage}
\begin{minipage}[b]{0.13\linewidth}
\includegraphics[width=\linewidth,trim=0 0 0 0,clip=true]{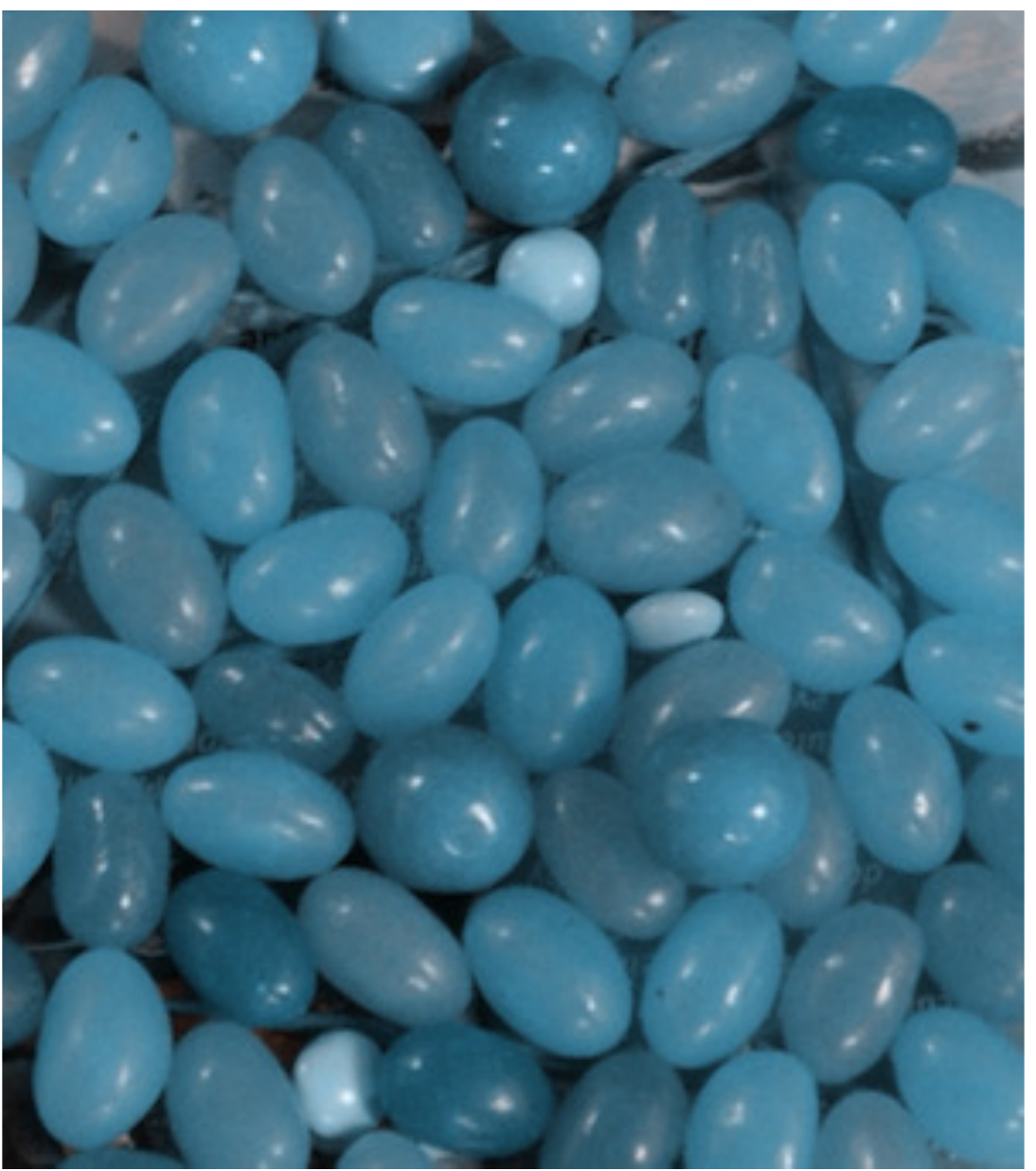}\\
\end{minipage}
\begin{minipage}[b]{0.13\linewidth}
\includegraphics[width=\linewidth,trim=0 10 65 20,clip=true]{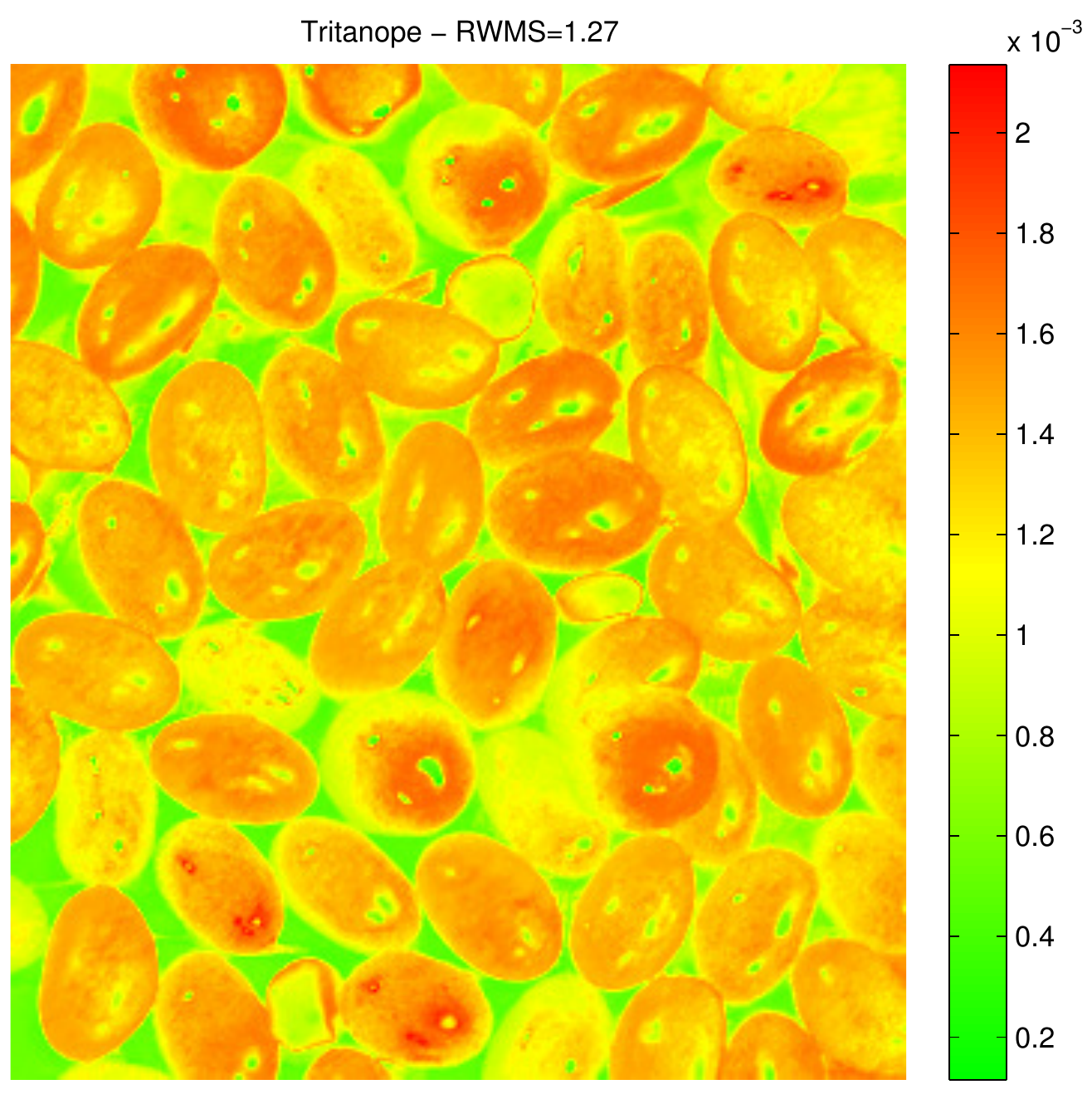}\\
\centering \small 1.27
\end{minipage}
\begin{minipage}[b]{0.13\linewidth}
\includegraphics[width=\linewidth,trim=0 0 0 0,clip=true]{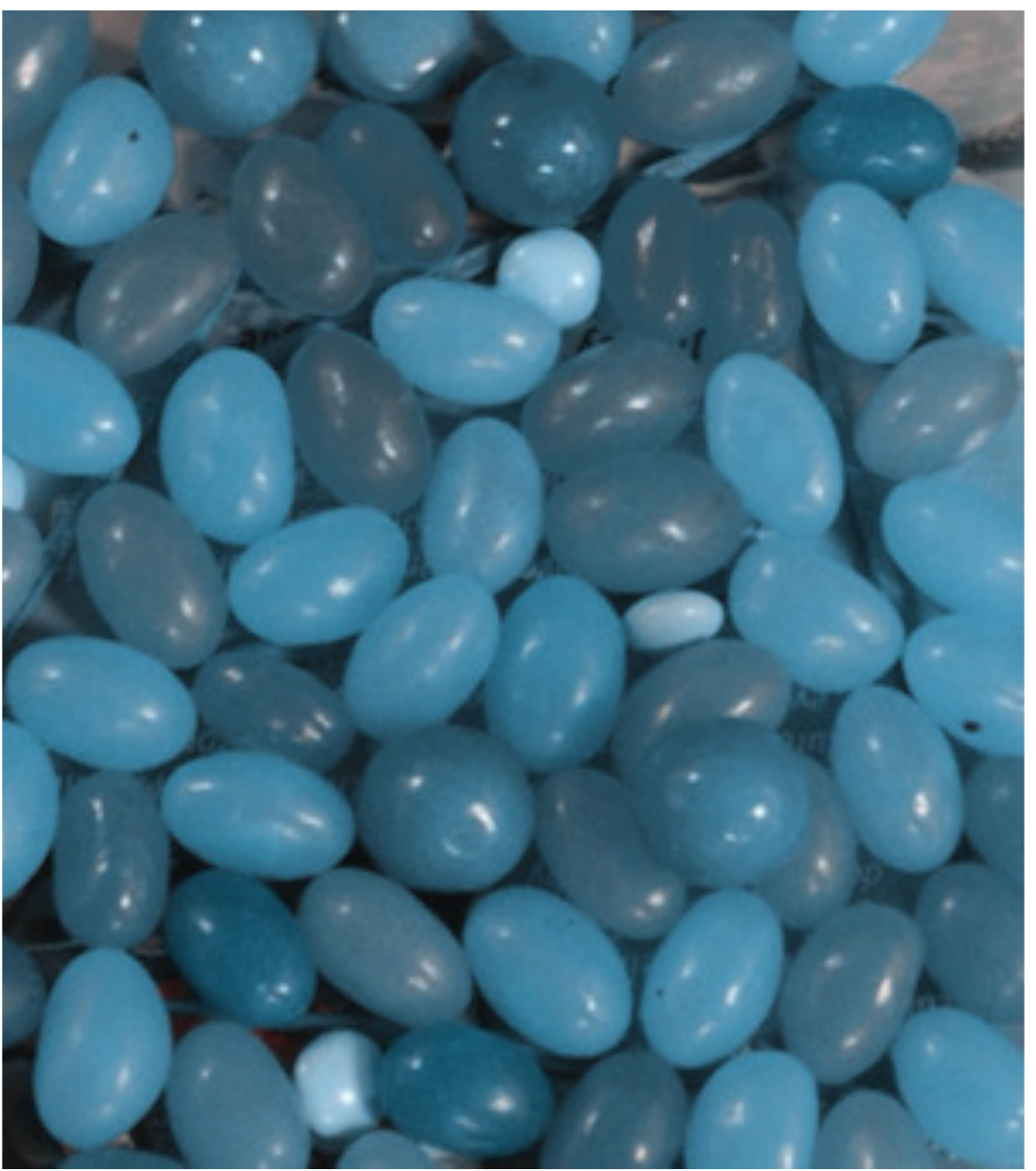}\\
\end{minipage}
\begin{minipage}[b]{0.13\linewidth}
\includegraphics[width=\linewidth,trim=0 10 65 20,clip=true]{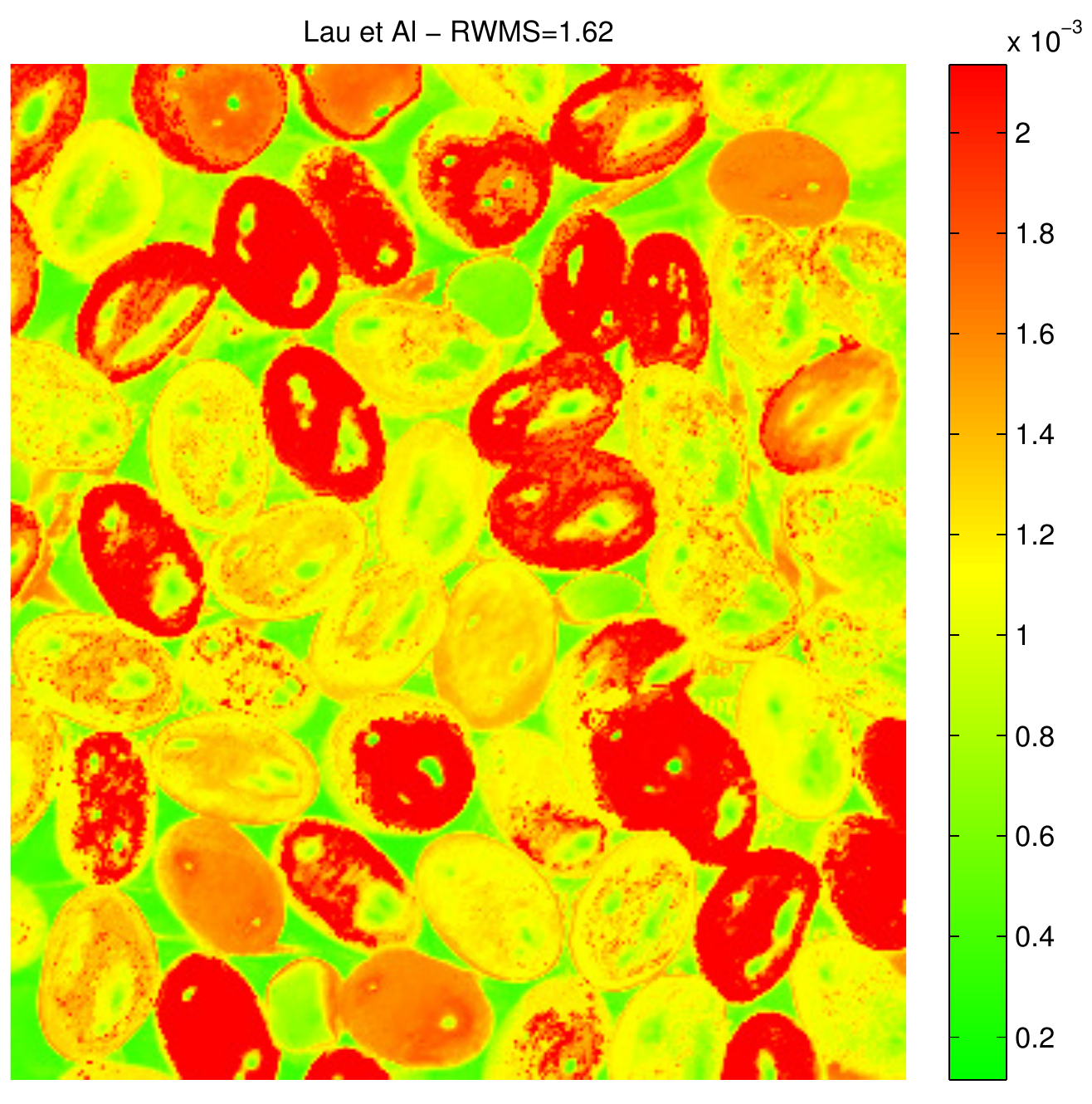}\\
\centering \small 1.69
\end{minipage}
\begin{minipage}[b]{0.13\linewidth}
\includegraphics[width=\linewidth,trim=0 0 0 0,clip=true]{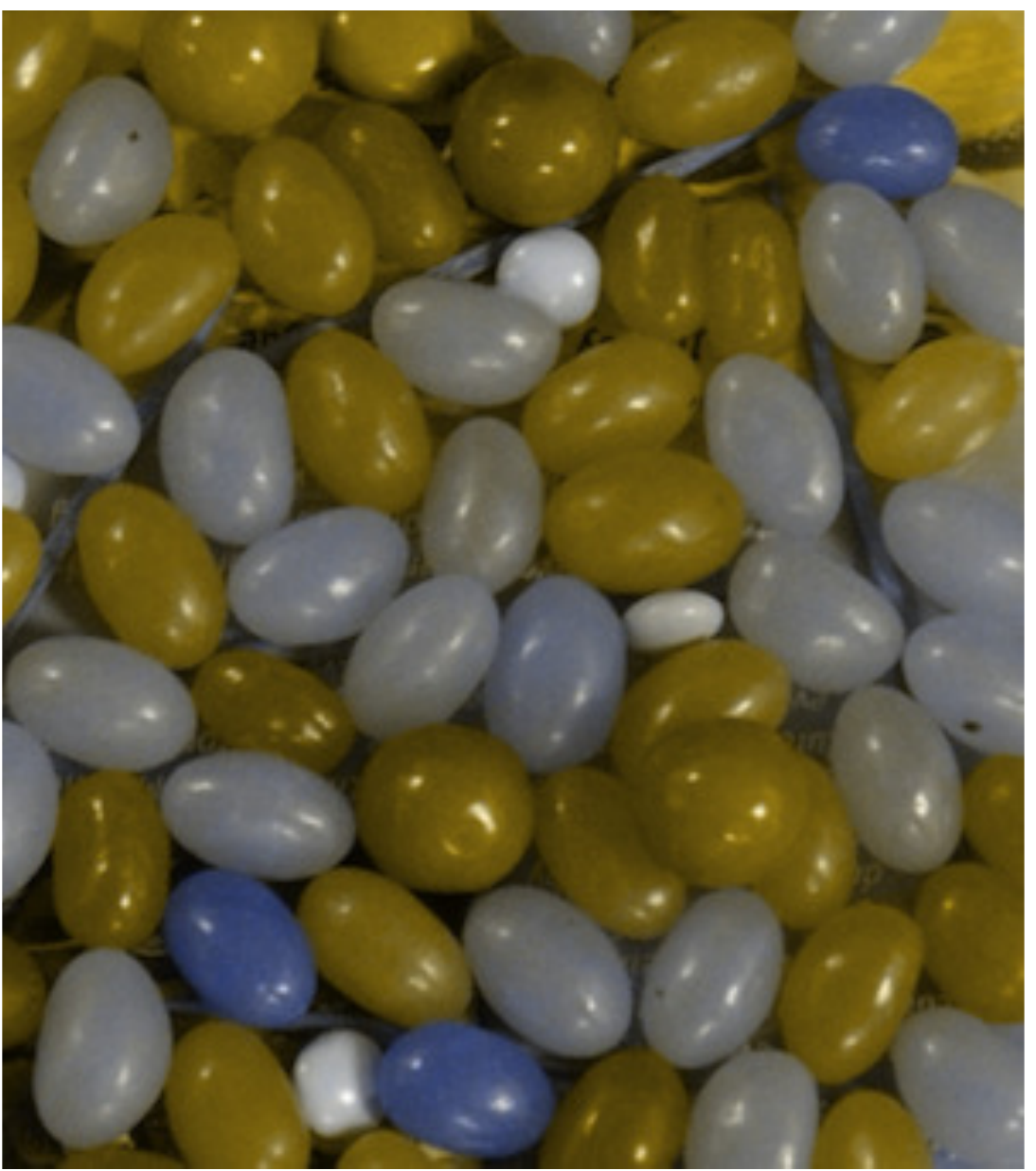}\\
\end{minipage}
\begin{minipage}[b]{0.13\linewidth}
\includegraphics[width=\linewidth,trim=0 10 65 20,clip=true]{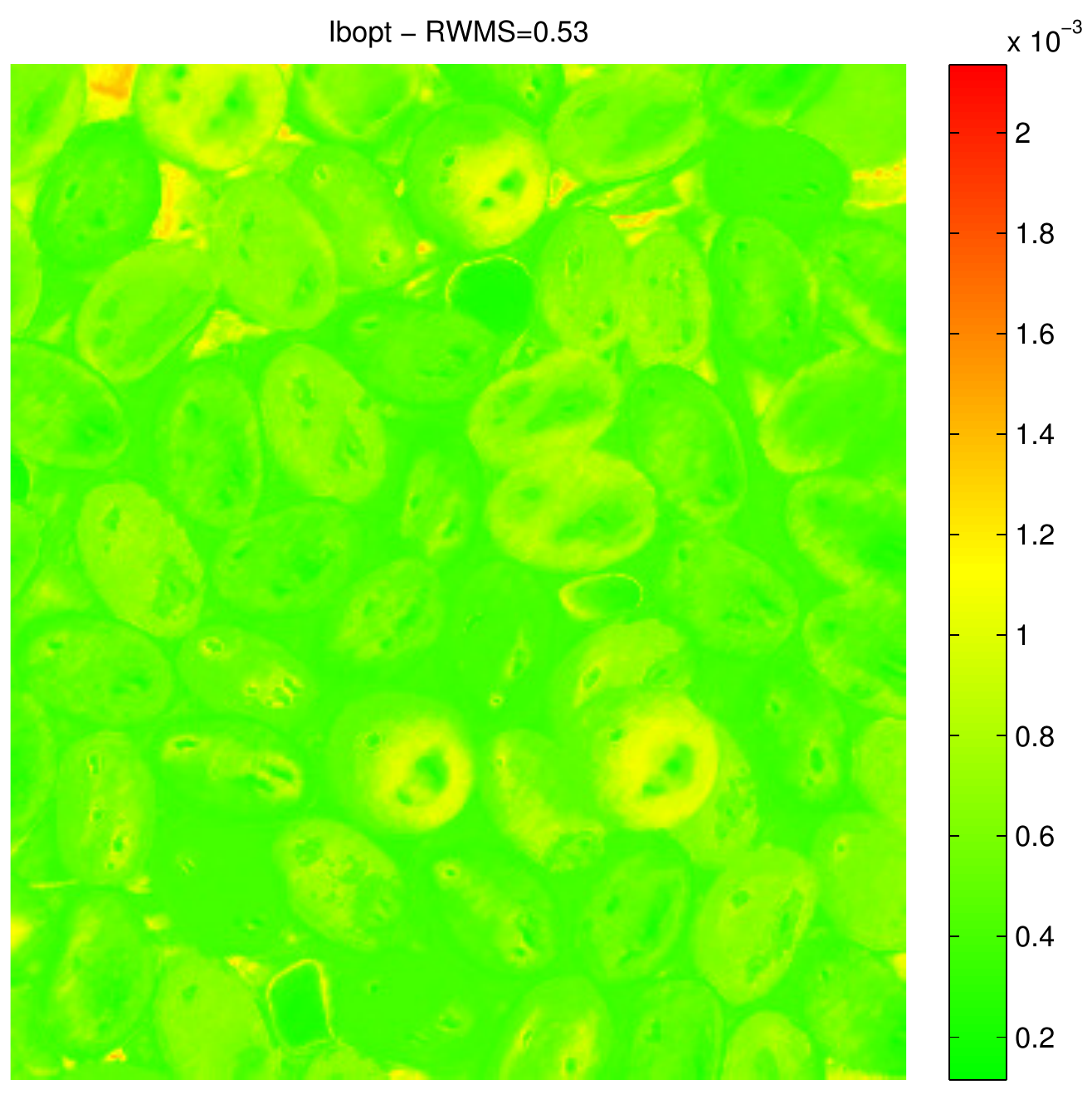}\\
\centering \small 0.53
\end{minipage}
\begin{minipage}[b]{0.05\linewidth}
\includegraphics[width=4.8mm,height=4.5mm,trim=0 0 0 0,clip=true]{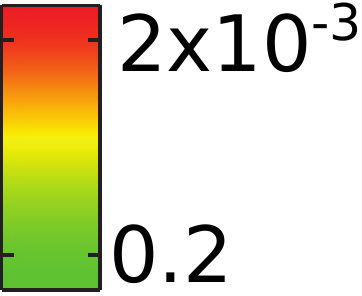}\\
\end{minipage}
\caption{Color mapping for color-blind observers (top: protanope, bottom: tritanope). From left: original image, simulated color-blind, result from \cite{Lau11}, and our result, with their respective RWMS error images and mean RWMS values.} 
\label{fig:colorblind} 
\vspace{-5mm}
\end{figure*}

Figure~\ref{fig:colorblind} shows Laplacian colormaps results for two different 
types of color blindness (protanopia and tritanopia). 
Qualitatively, our result appears to be much closer to the original image compared to~\cite{Lau11} (this is especially apparent in the tritanopia case) such that a `normal' viewer sees less distorted colors, while a color-deficient viewer can clearly see the structure structured in the image (digit 6 and different candies) which otherwise would disappear.    
Quantitatively, we obtain smaller RWMS error, suggesting  
that our mapping better preserves the original structure of the image, even in those 
areas that are critical for other approaches.

\begin{figure*}[ht]
\centering
\begin{minipage}[b]{0.24\linewidth}
\centering \small Original image
\includegraphics[width=\linewidth,trim=0 0 0 0,clip=true]{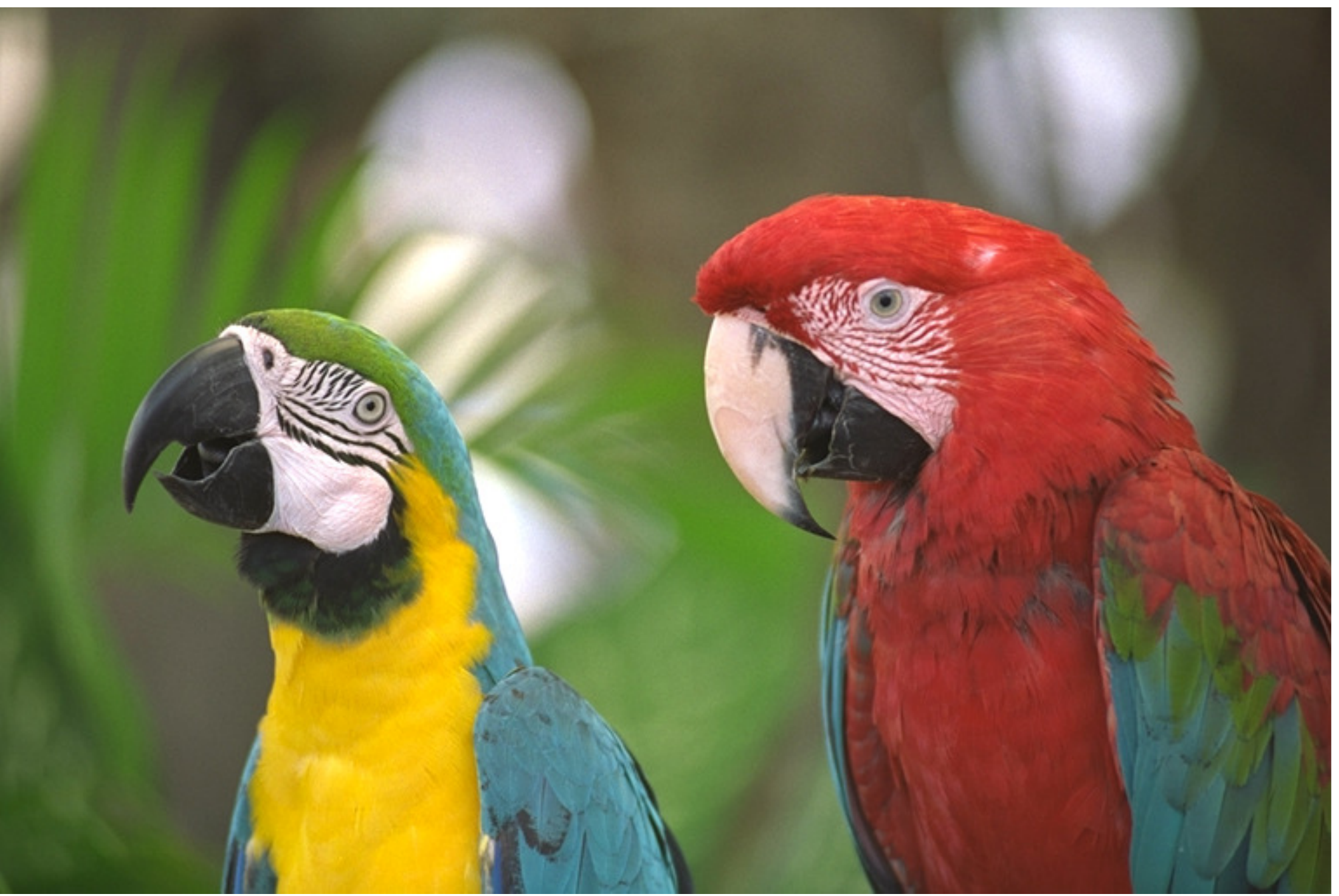}\\
\end{minipage}
\begin{minipage}[b]{0.24\linewidth}
\centering \small \cite{Lau11}
\includegraphics[width=\linewidth,trim=0 0 0 0,clip=true]{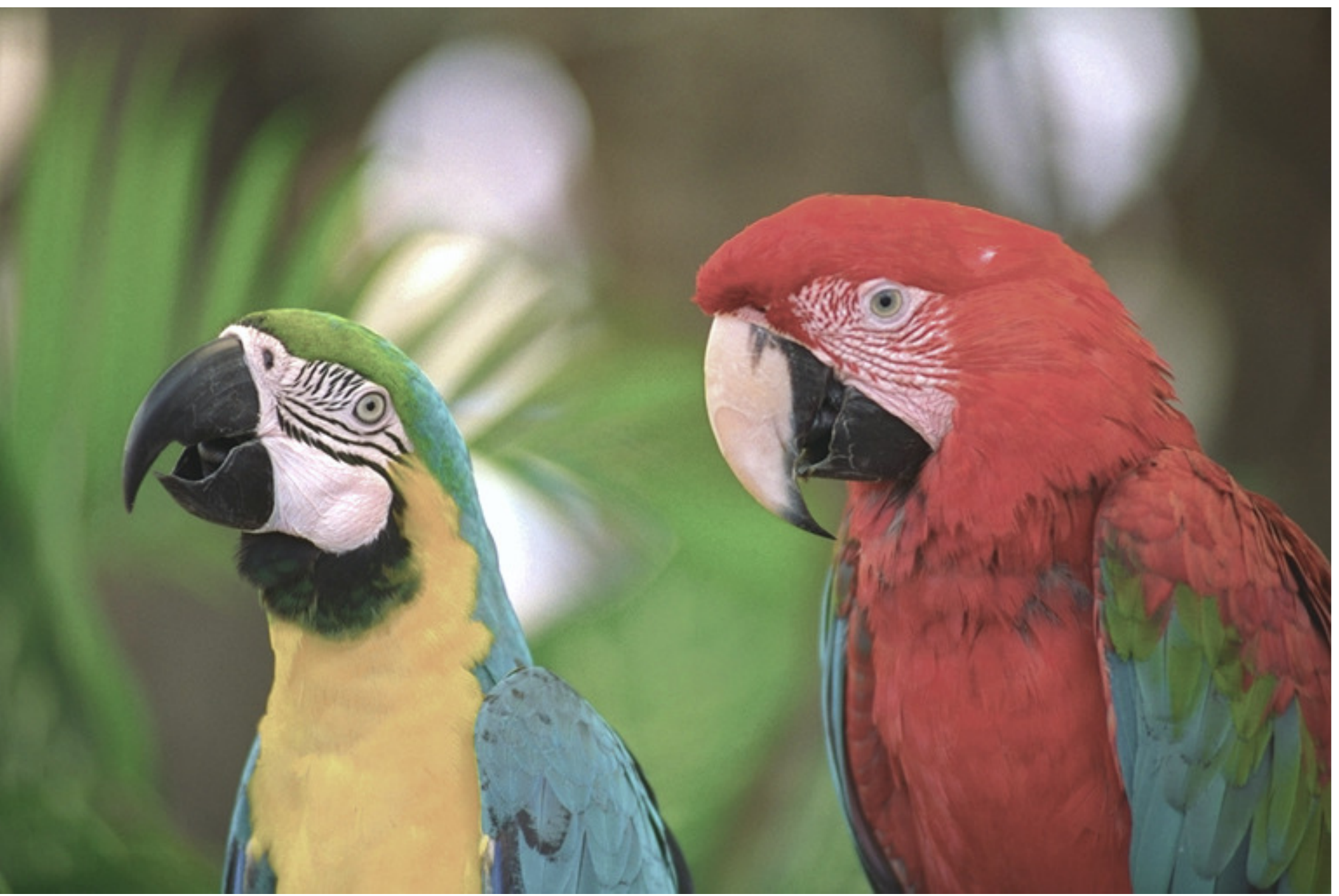}\\
\end{minipage}
\begin{minipage}[b]{0.24\linewidth}
\centering \small \textbf{Laplacian}
\includegraphics[width=\linewidth,trim=0 0 0 0,clip=true]{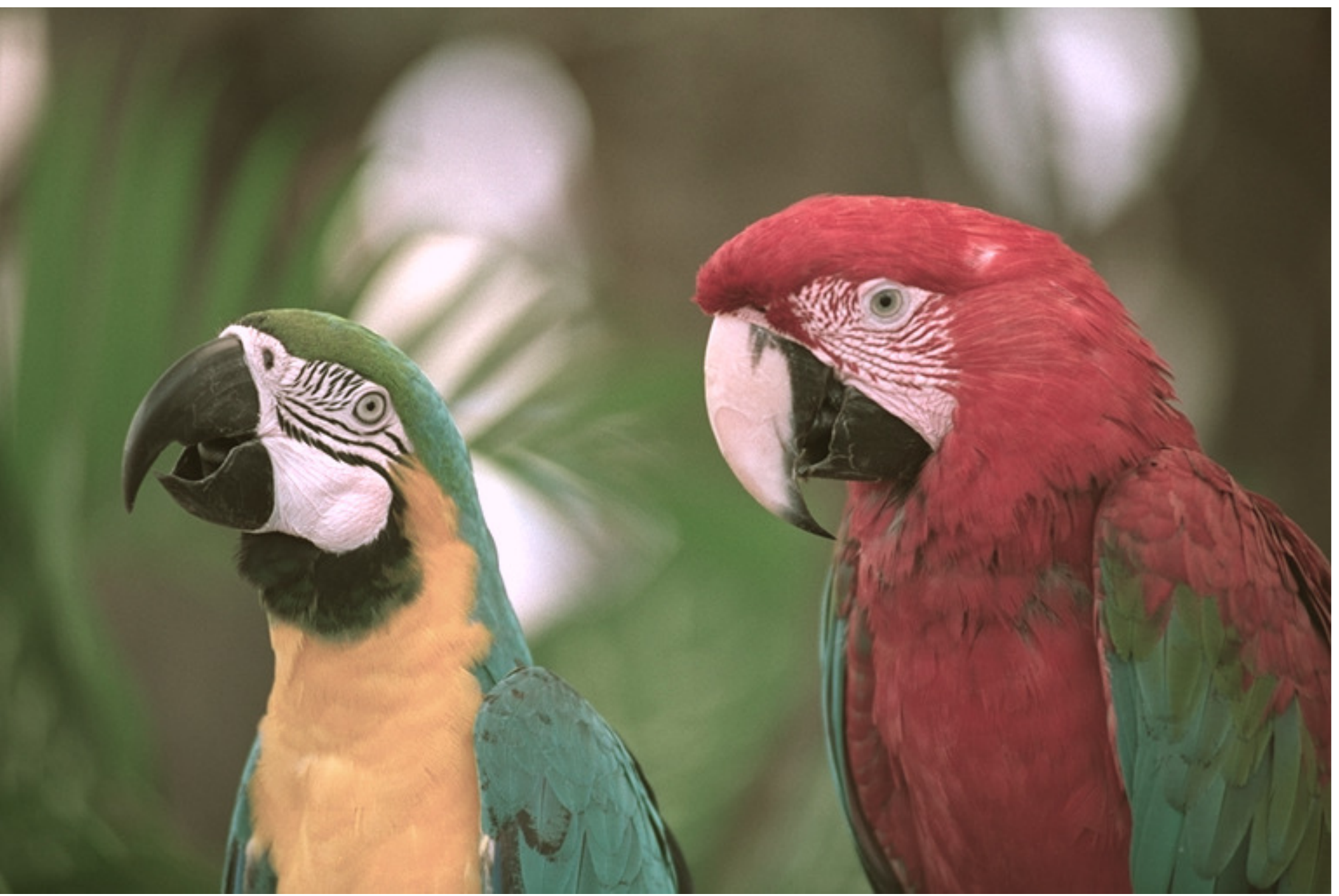}\\
\end{minipage}
\begin{minipage}[b]{0.24\linewidth}
\centering \small HPMINDE \cite{CIE04}
\includegraphics[width=\linewidth,trim=0 0 0 0,clip=true]{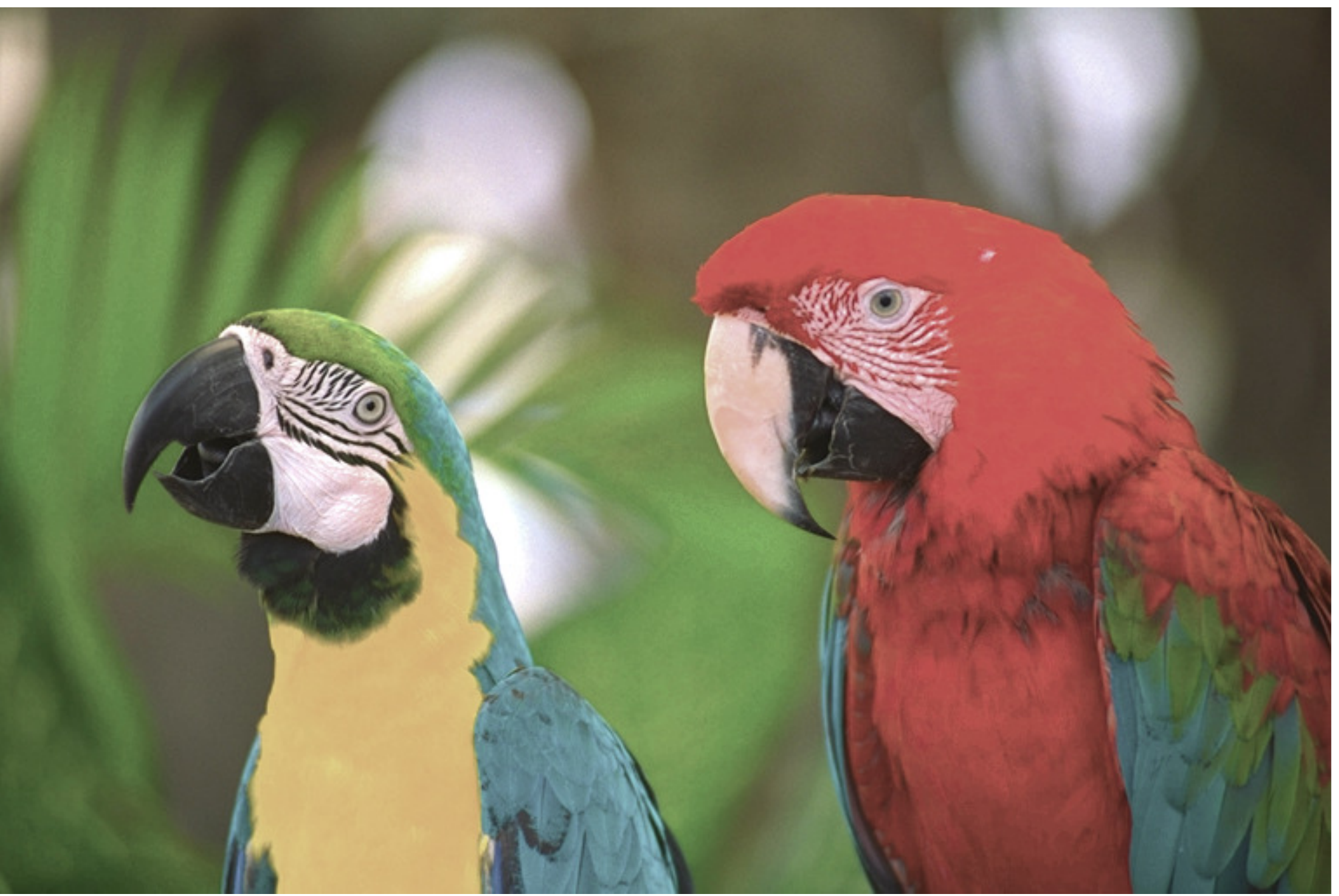}\\
\end{minipage}
\\
\begin{minipage}[b]{0.24\linewidth}
\includegraphics[width=\linewidth,trim=0 0 0 0,clip=true]{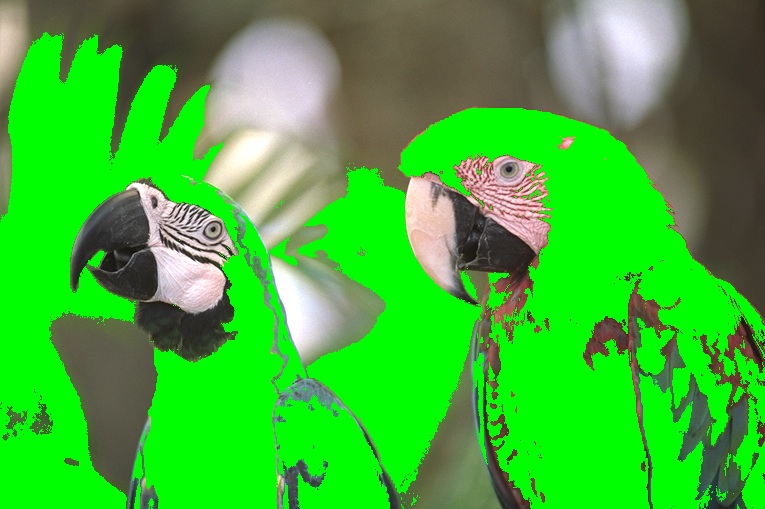}
\end{minipage}
\begin{minipage}[b]{0.24\linewidth}
\includegraphics[width=\linewidth,trim=0 0 0 0,clip=true]{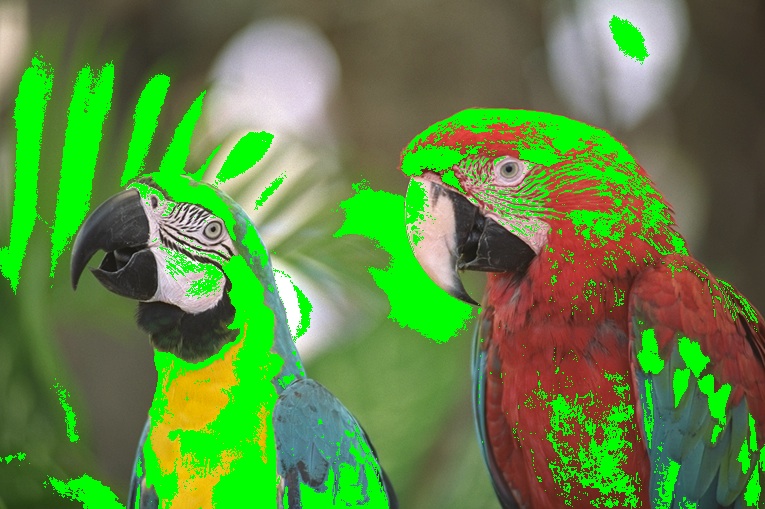}
\end{minipage}
\begin{minipage}[b]{0.24\linewidth}
\includegraphics[width=\linewidth,trim=0 0 0 0,clip=true]{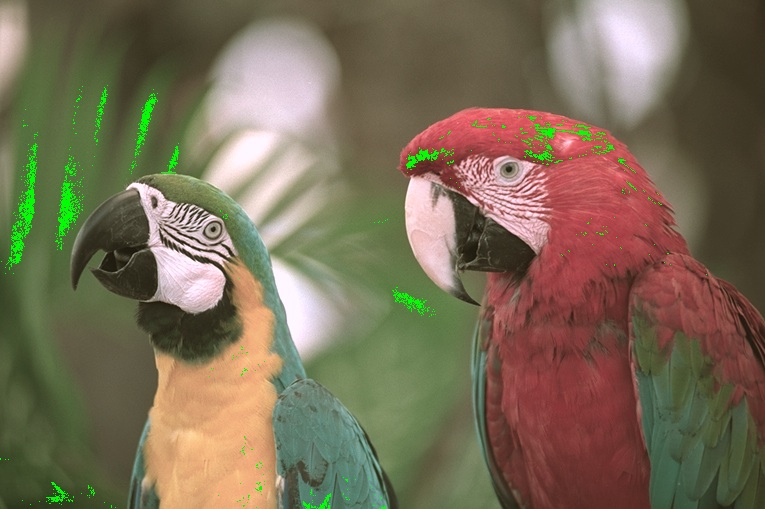}
\end{minipage}
\begin{minipage}[b]{0.24\linewidth}
\hspace{1mm}
\end{minipage}
\\
\vspace{2mm}
\begin{minipage}[b]{0.24\linewidth}
\centering \small Original image
\includegraphics[width=\linewidth,trim=0 0 0 0,clip=true]{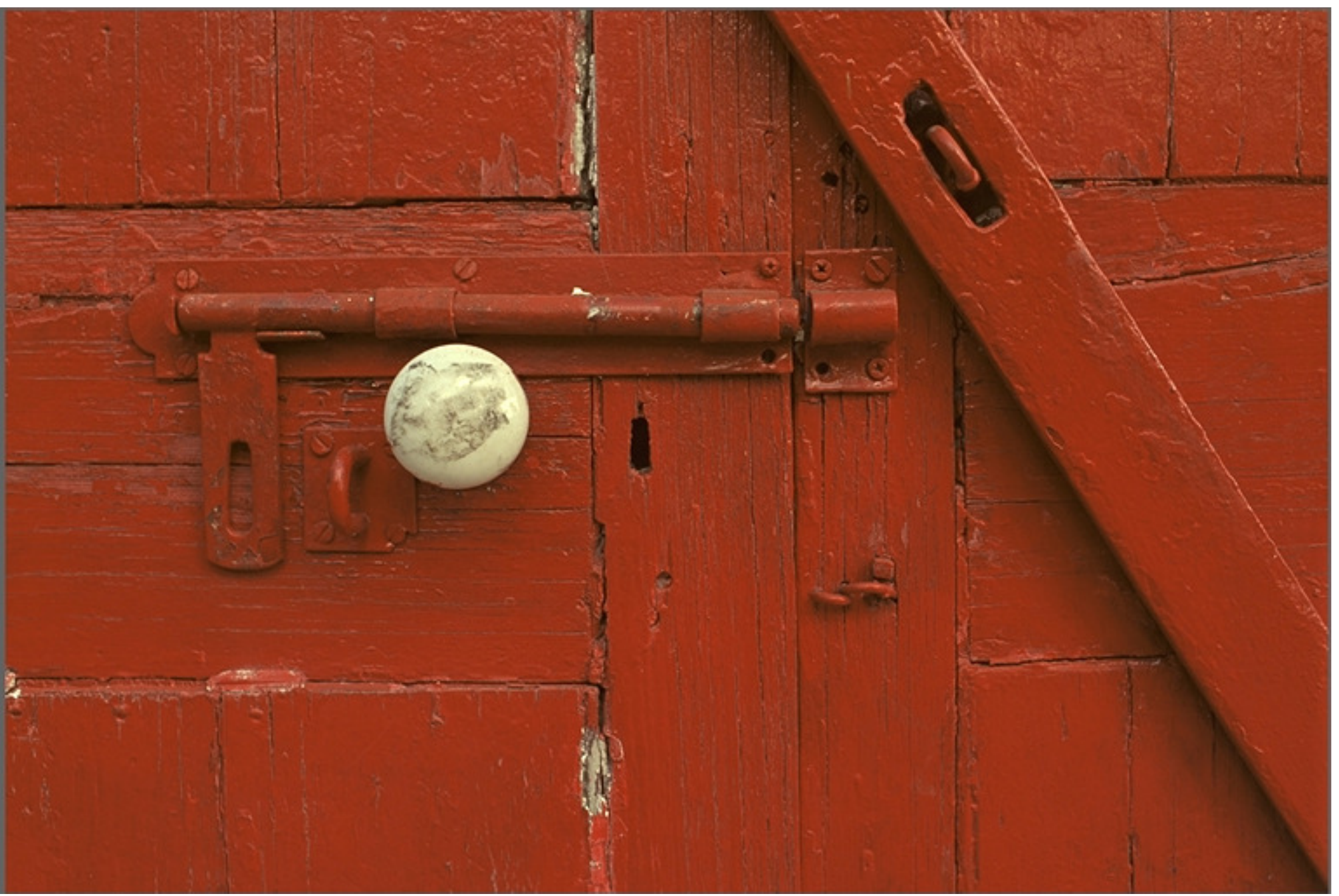}\\
\end{minipage}
\begin{minipage}[b]{0.24\linewidth}
\centering \small \cite{Lau11}
\includegraphics[width=\linewidth,trim=0 0 0 0,clip=true]{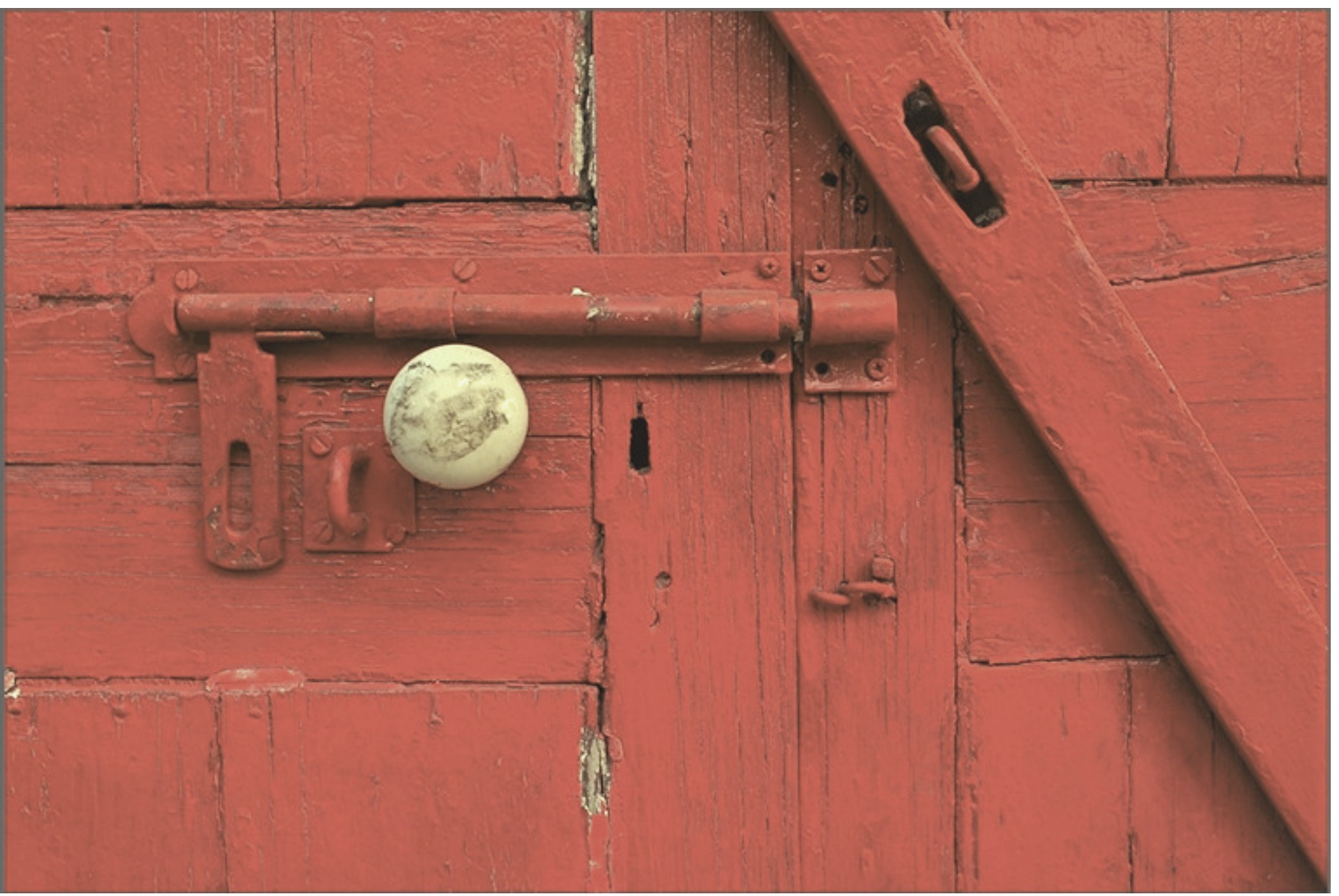}\\
\end{minipage}
\begin{minipage}[b]{0.24\linewidth}
\centering \small \textbf{Laplacian}
\includegraphics[width=\linewidth,trim=0 0 0 0,clip=true]{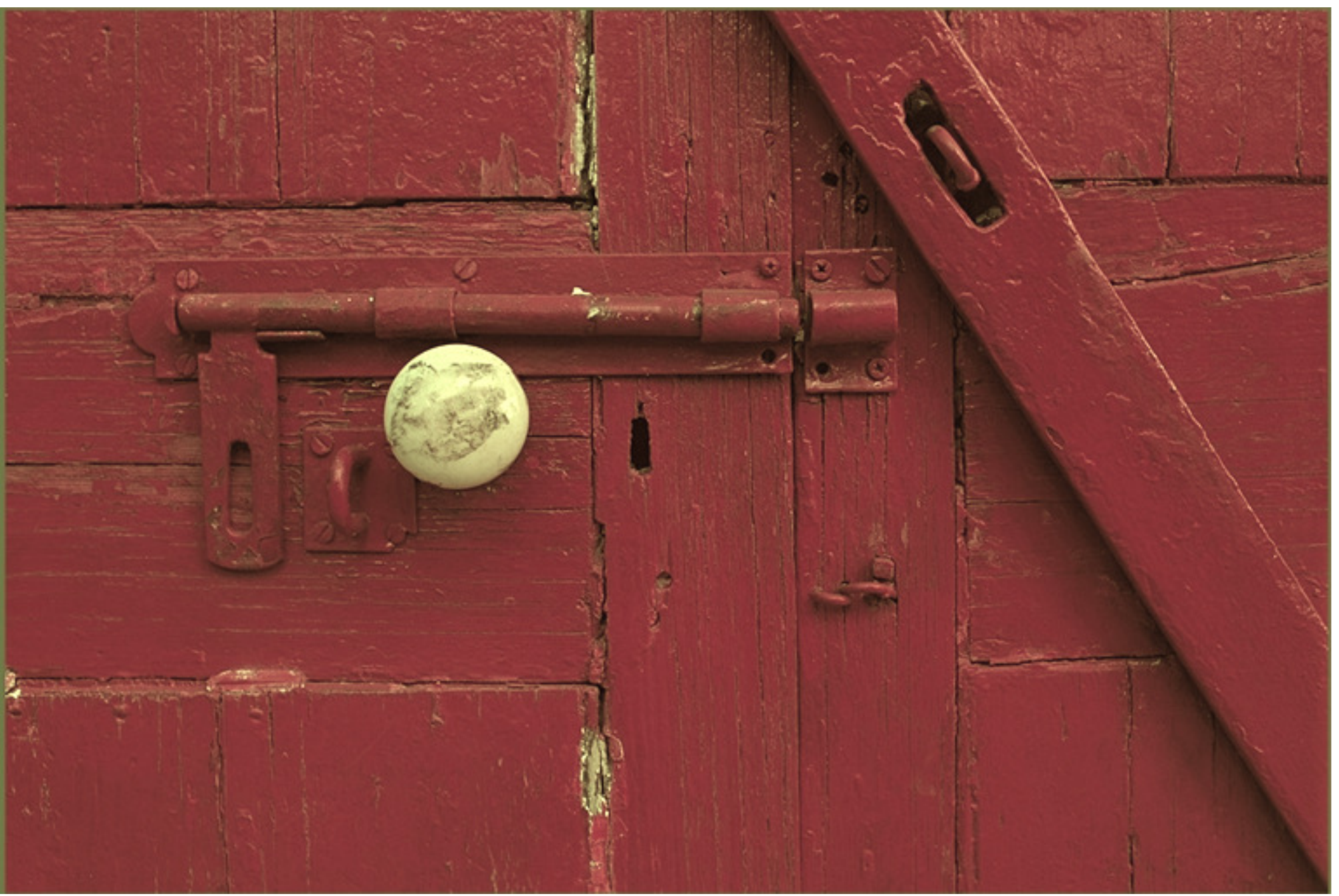}\\
\end{minipage}
\begin{minipage}[b]{0.24\linewidth}
\centering \small HPMINDE \cite{CIE04}
\includegraphics[width=\linewidth,trim=0 0 0 0,clip=true]{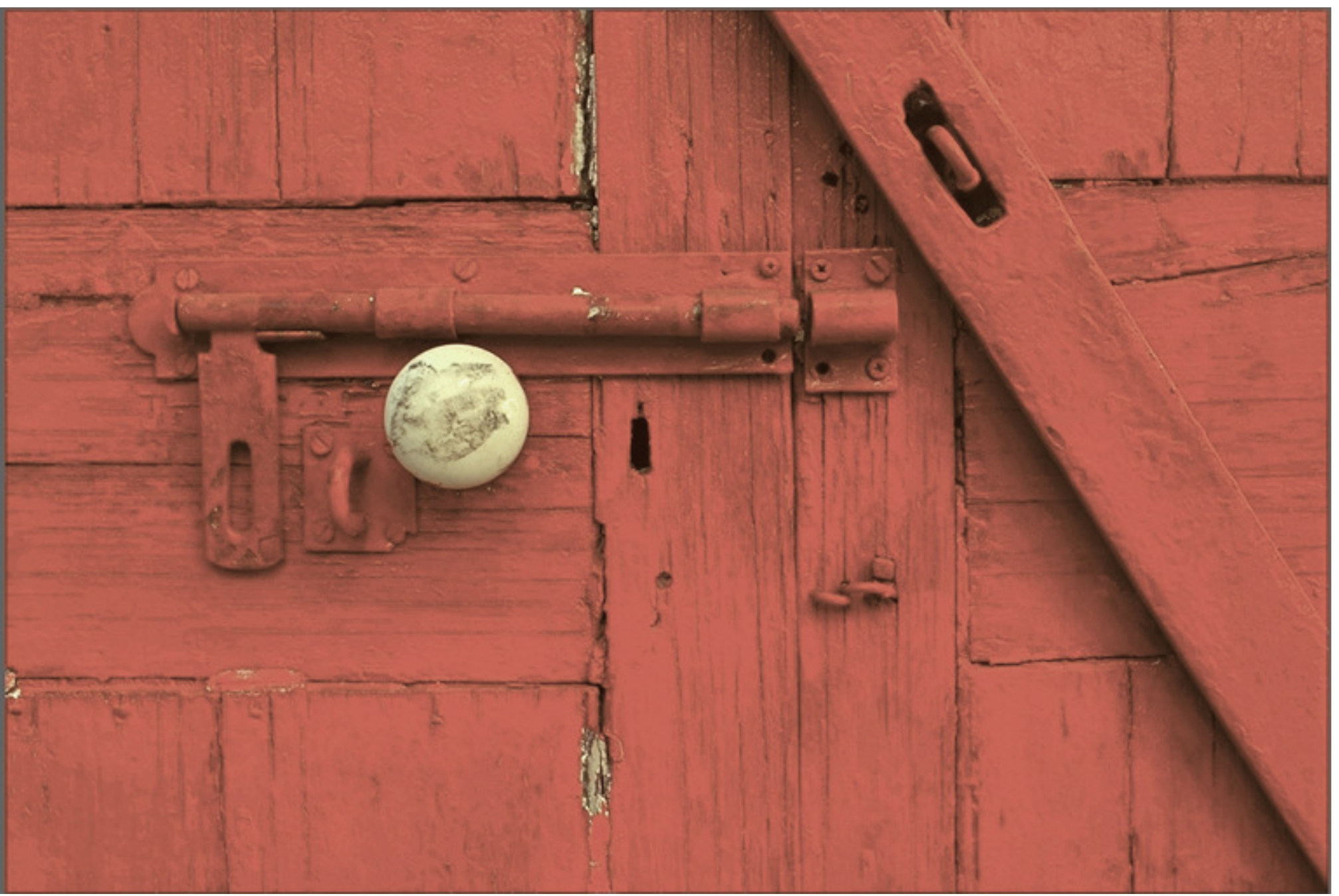}\\
\end{minipage}
\\
\begin{minipage}[b]{0.24\linewidth}
\includegraphics[width=\linewidth,trim=0 0 0 0,clip=true]{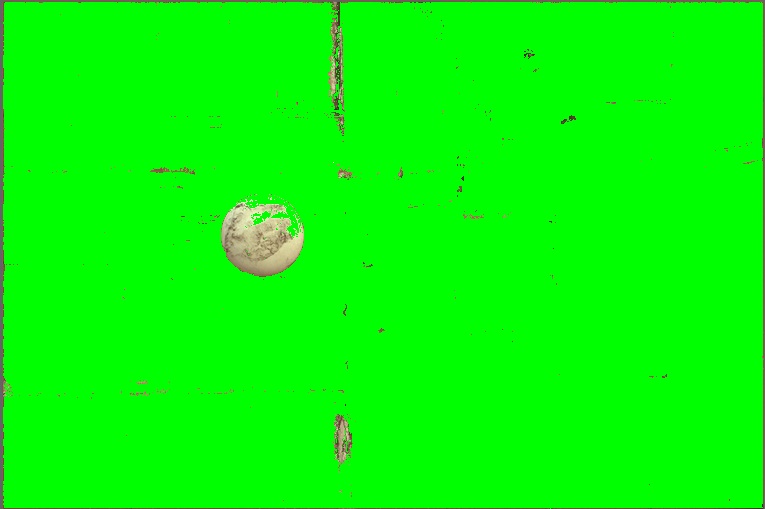}\\
\end{minipage}
\begin{minipage}[b]{0.24\linewidth}
\includegraphics[width=\linewidth,trim=0 0 0 0,clip=true]{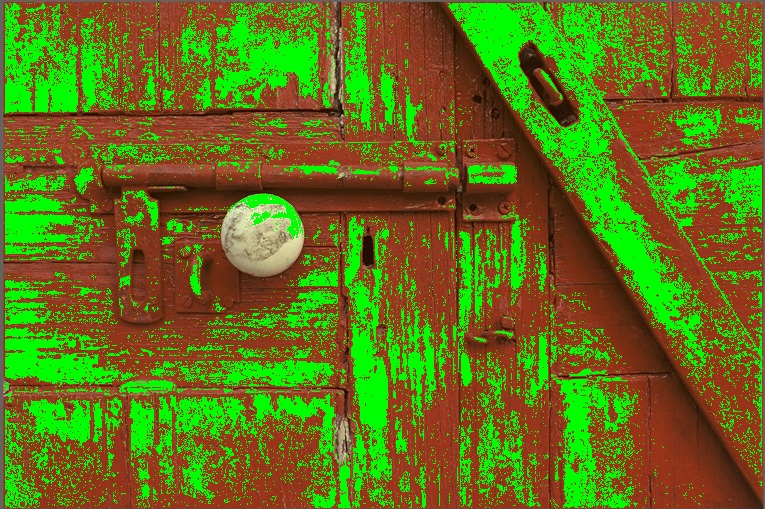}\\
\end{minipage}
\begin{minipage}[b]{0.24\linewidth}
\includegraphics[width=\linewidth,trim=0 0 0 0,clip=true]{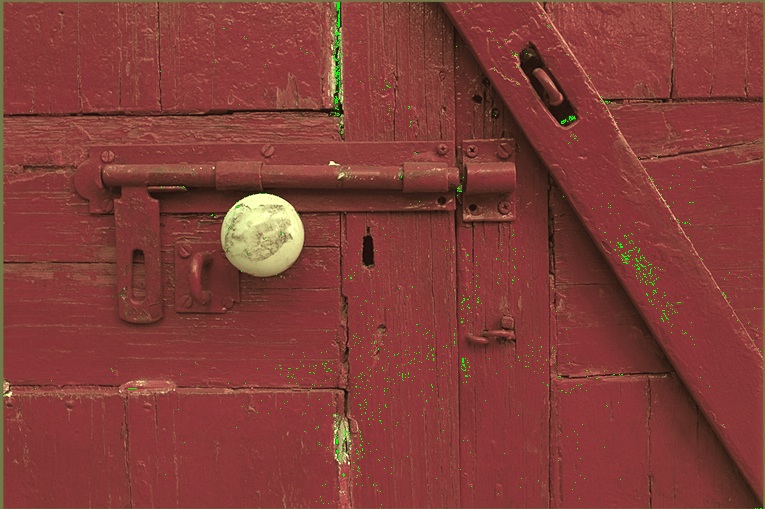}\\
\end{minipage}
\begin{minipage}[b]{0.24\linewidth}
\hspace{1mm}
\end{minipage}
  \caption{Gamut mapping results. Odd rows, left-to-right: original image, gamut mapping with method of Lau et al. \cite{Lau11}, our approach and  HPMINDE \cite{CIE04}. 
Even rows: gamut alerts for the images above (green shows the out-of-gamut pixels).}
\label{fig:gamut}
\vspace{-2mm}
\end{figure*}

\textbf{Gamut mapping} is a problem similar to the previous one, and has a setting similar to the one in the previous experiment.
A transformation $\Psi$ which maps colors from RGB to the XY chromaticity space and a
color gamut $G$ (a convex polytope, and in this particular experiment a triangle) are given. 
Our goal is to find $\bb{\theta}$ minimizing the cost~(\ref{eq:costfunction2}) subject to 
$(\Phi_{\bb{\theta}} \circ \Psi )(\Xx) \subseteq G$, which is imposed as a set of linear constraints. 
We used the parameters
$\mu_{01}=1, \mu_{11}=0.25, \mu_2=0.1$, and $\mu_{02}=\mu_{12}=0$.
Figure \ref{fig:gamut} compares our results with the outputs of HPMINDE \cite{CIE04} 
and by the method of Lau et al. \cite{Lau11}. 
Qualitatively, the output of Laplacian colormaps 
preserves more details of the original picture (see e.g. the plumage on the red parrot's head). 
Quantitatively, our algorithm outperforms the other methods in term of percentage of out-of-gamut. 

\textbf{Multispectral image fusion}
can be seen as an extension of the color-to-gray problem, where the number of input channels $d>3$ and the output image has $d'=3$ channels. 
We use the cost function~(\ref{eq:costfunction3}), with $\mu_{01}=\mu_{02}=\mu_{11}=\mu_{12}=\mu_2 = 1$ and $\mu_3 > 0$;  
the latter does not only act as regularization, but also provides us a way to automatically order the three output channels. 

\begin{figure*}[ht]
\centering
\begin{minipage}[b]{0.24\linewidth}
\centering \small NIR
\includegraphics[width=\linewidth,trim=0 0 0 0,clip=true]{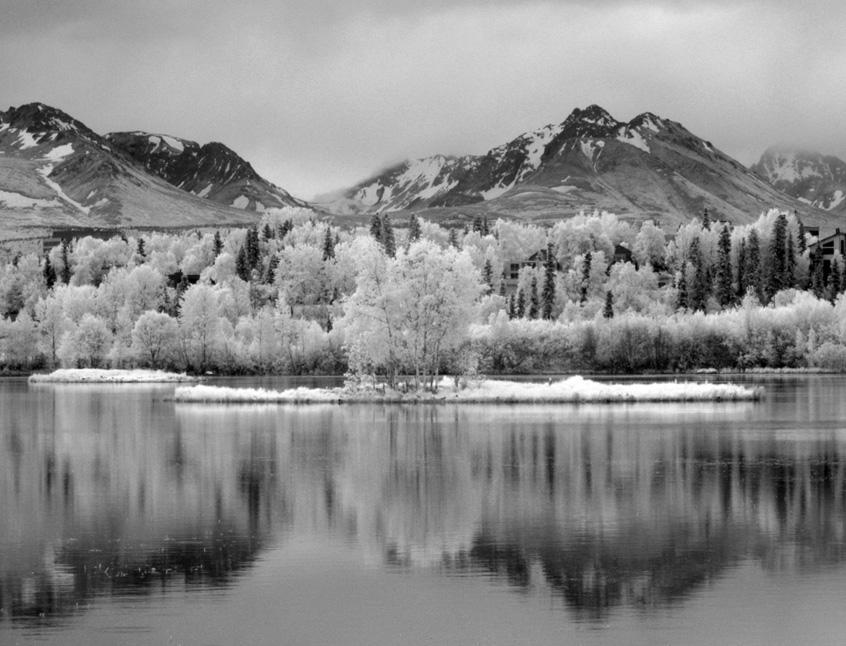}\\
\end{minipage}
\begin{minipage}[b]{0.24\linewidth}
\centering \small RGB
\includegraphics[width=\linewidth,trim=0 0 0 0,clip=true]{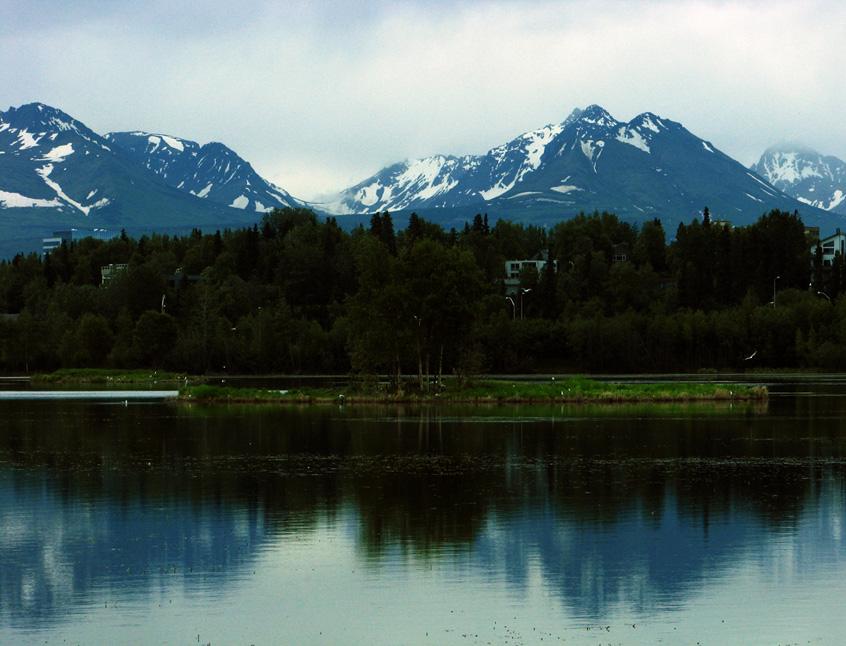}\\
\end{minipage}
\begin{minipage}[b]{0.24\linewidth}
\centering \small \cite{Lau11}
\includegraphics[width=\linewidth,trim=0 0 0 0,clip=true]{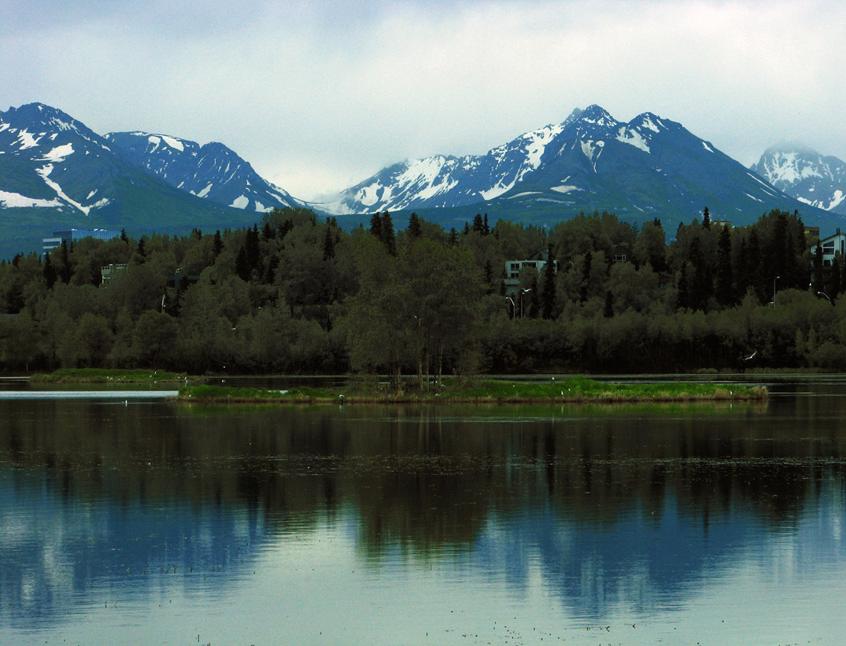}\\
\end{minipage}
\begin{minipage}[b]{0.24\linewidth}
\centering \small \textbf{Laplacian}
\includegraphics[width=\linewidth,trim=0 0 0 0,clip=true]{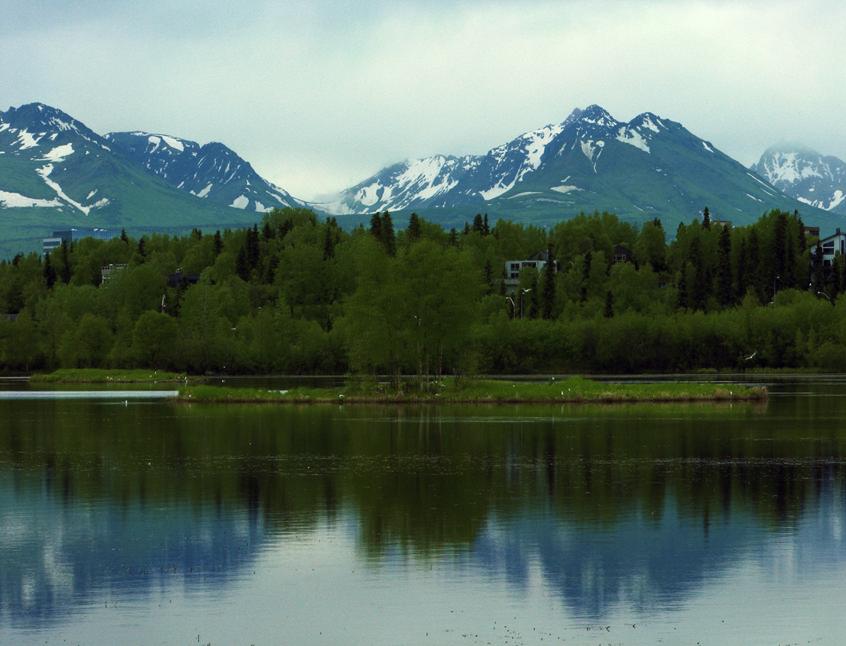}\\
\end{minipage}
\caption{Multispectral (RGB+NIR) fusion results.}
\label{fig:nir2}
\vspace{-5mm}
\end{figure*}

\begin{figure*}[ht]
\centering
\begin{minipage}[b]{0.19\linewidth}
\centering \small Morning
\includegraphics[width=\linewidth,trim=0 0 0 0,clip=true]{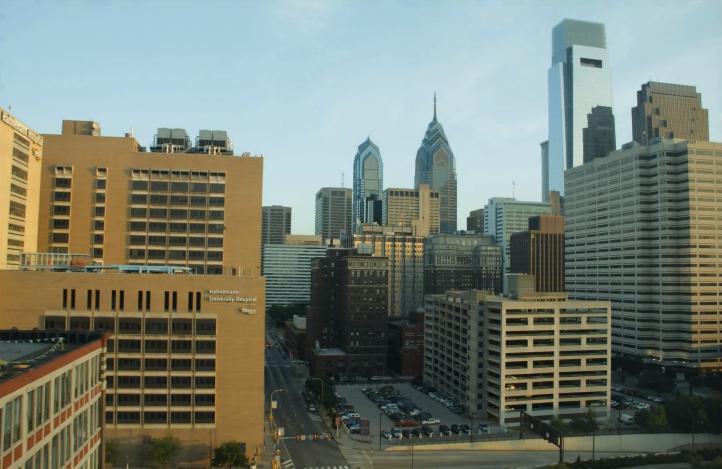}\\
\end{minipage}
\begin{minipage}[b]{0.19\linewidth}
\centering \small Day
\includegraphics[width=\linewidth,trim=0 0 0 0,clip=true]{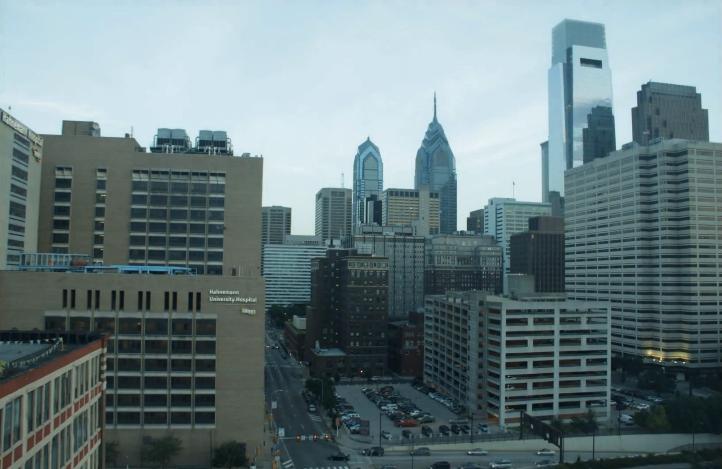}\\
\end{minipage}
\begin{minipage}[b]{0.19\linewidth}
\centering \small Evening
\includegraphics[width=\linewidth,trim=0 0 0 0,clip=true]{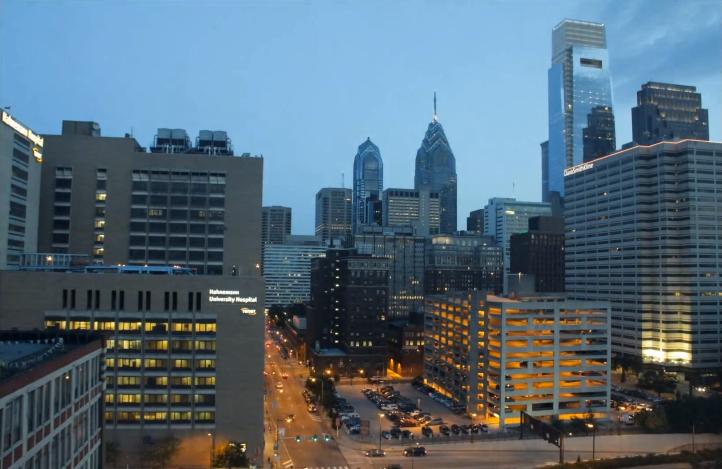}\\
\end{minipage}
\begin{minipage}[b]{0.19\linewidth}
\centering \small Night
\includegraphics[width=\linewidth,trim=0 0 0 0,clip=true]{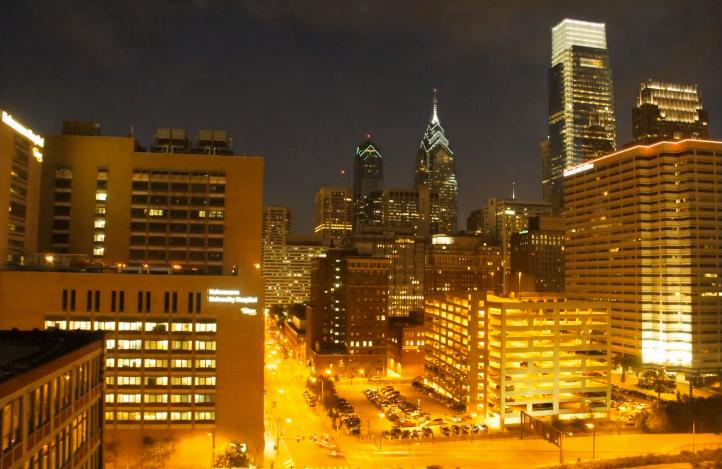}\\
\end{minipage}
\begin{minipage}[b]{0.19\linewidth}
\centering \small \textbf{Fusion}
\includegraphics[width=\linewidth,trim=0 0 0 0,clip=true]{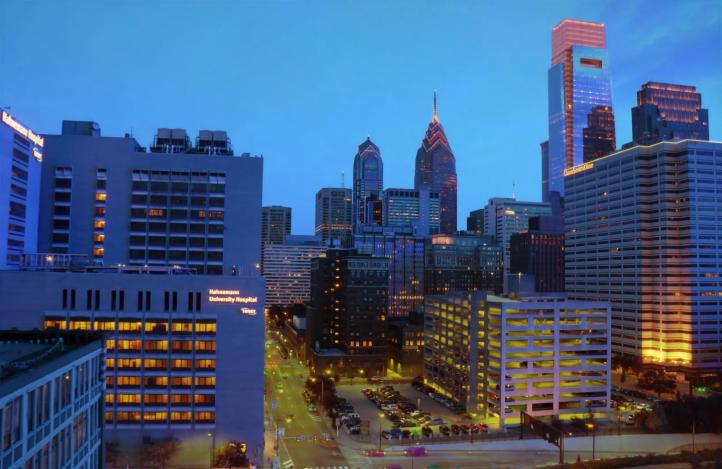}\\
\end{minipage}
\caption{Fusion of images of four different illuminations into a single RGB image (rightmost).}
\label{fig:timelapse}
\vspace{-5mm}
\end{figure*}

Figure \ref{fig:nir2} shows multispectral to RGB transformations where the input space is the concatenation of RGB and NIR ($d=4$). 
In this specific example, the NIR channel is used to enhance the RGB image with an additional source of information. 
Comparing our result with the method of Lau et al.~\cite{Lau11}, we can see that Laplacian colormap provides an enhanced version of RGB while preserving the correct colors (e.g. trees on the mountains have more detail than in RGB, but at the same time they do not present the blue-green halo that appears in~\cite{Lau11}).
Finally, in Figure \ref{fig:timelapse} we show a fusion of four photos of a city in different lighting conditions into a single image, which looks visually plausible.

\begin{figure*}[t!]
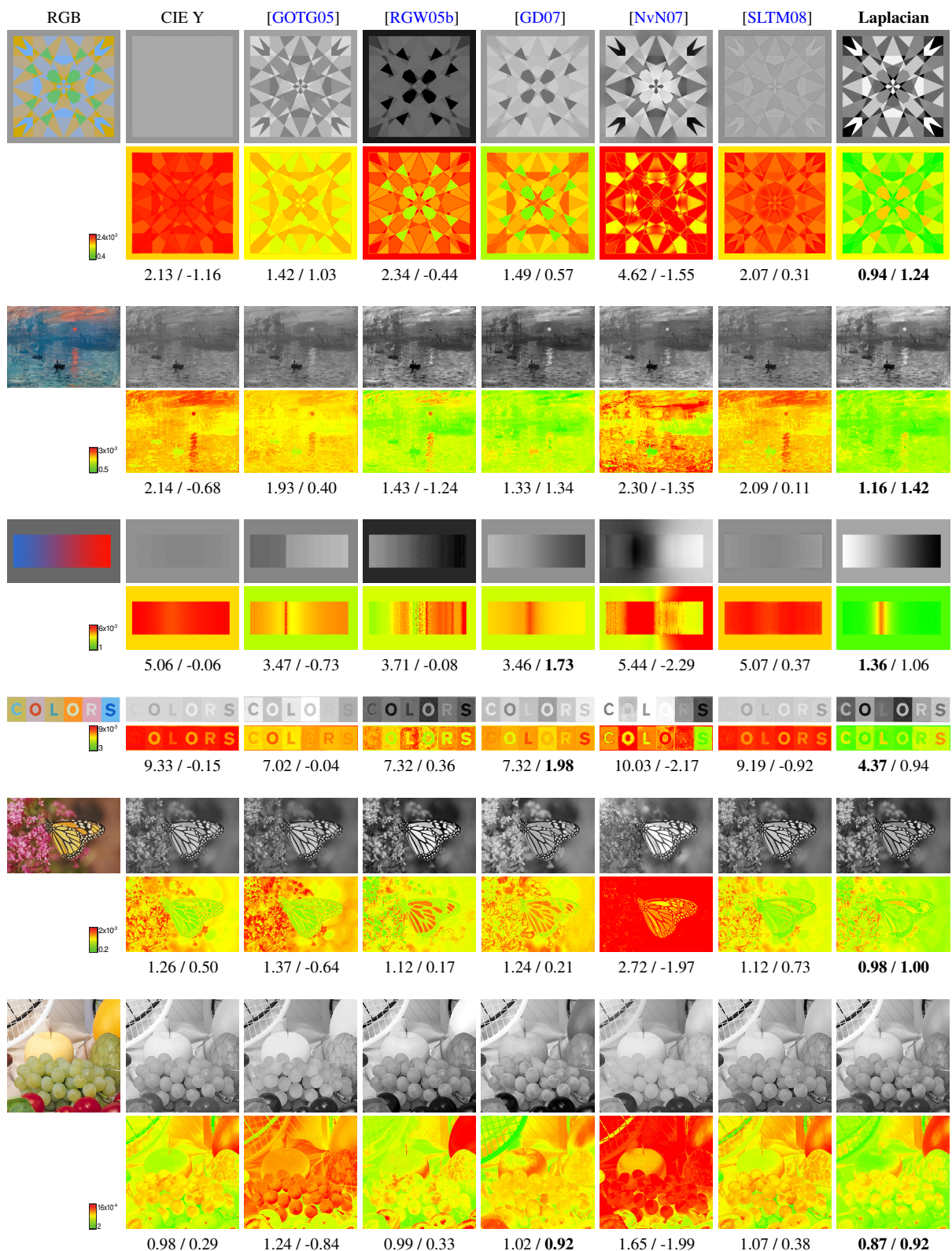

\centering
\begin{minipage}[b]{0.12\linewidth}
\centering \small RGB
\end{minipage}
\begin{minipage}[b]{0.12\linewidth}
\centering \small CIE Y
\end{minipage}
\begin{minipage}[b]{0.12\linewidth}
\centering \small \cite{Gooch05} 
\end{minipage}
\begin{minipage}[b]{0.12\linewidth}
\centering \small \cite{Rasche05} 
\end{minipage}
\begin{minipage}[b]{0.12\linewidth}
\centering \small \cite{Grundland07}
\end{minipage}
\begin{minipage}[b]{0.12\linewidth}
\centering \small \cite{Neumann07}
\end{minipage}
\begin{minipage}[b]{0.12\linewidth}
\centering \small \cite{Smith08}
\end{minipage}
\begin{minipage}[b]{0.12\linewidth}
\centering \small \textbf{Laplacian}
\end{minipage}
\rgbtogray{IM2-color}{2.13 / -1.16}{1.42 / 1.03}{2.34 / -0.44}{1.49 / 0.57}{4.62 / -1.55}{2.07 / 0.31}{\textbf{0.94} / \textbf{1.24}}
\vspace{0mm}\\ 
\rgbtogray{Sunrise312}{2.14 / -0.68}{1.93 / 0.40}{1.43 / -1.24}{1.33 / 1.34}{2.30 / -1.35}{2.09 / 0.11}{\textbf{1.16} / \textbf{1.42}}
\vspace{0mm}\\ 
\rgbtogray{ramp}{5.06 / -0.06}{3.47 / -0.73}{3.71 / -0.08}{3.46 / \textbf{1.73}}{5.44 / -2.29}{5.07 / 0.37}{\textbf{1.36} / 1.06}
\vspace{0mm}\\
\rgbtogray{ColorsPastel}{9.33 / -0.15}{7.02 / -0.04}{7.32 / 0.36}{7.32 / \textbf{1.98}}{10.03 / -2.17}{9.19 / -0.92}{\textbf{4.37} / 0.94}
\vspace{0mm}\\
\rgbtogray{monarch}{1.26 / 0.50}{1.37 / -0.64}{1.12 / 0.17}{1.24 / 0.21}{2.72 / -1.97}{1.12 / 0.73}{\textbf{0.98} / \textbf{1.00}}
\vspace{0mm}\\ 
\rgbtogray{fruits}{0.98 / 0.29}{1.24 / -0.84}{0.99 / 0.33}{1.02 / \textbf{0.92}}{1.65 / -1.99}{1.07 / 0.38}{\textbf{0.87} / \textbf{0.92}}
\\
\caption{Decolorization experiment results. Left: original RGB image, right: grayscale conversion results. 
Rows 2, 5, : RWMS error images and mean RWMS (the smaller the better) / z-score (the larger the better) values. Our Laplacian colormap method performs the best in most cases.
(Continues in next page)
}
\label{fig:rgb2gray}
\end{figure*}


\begin{figure*}[t!]
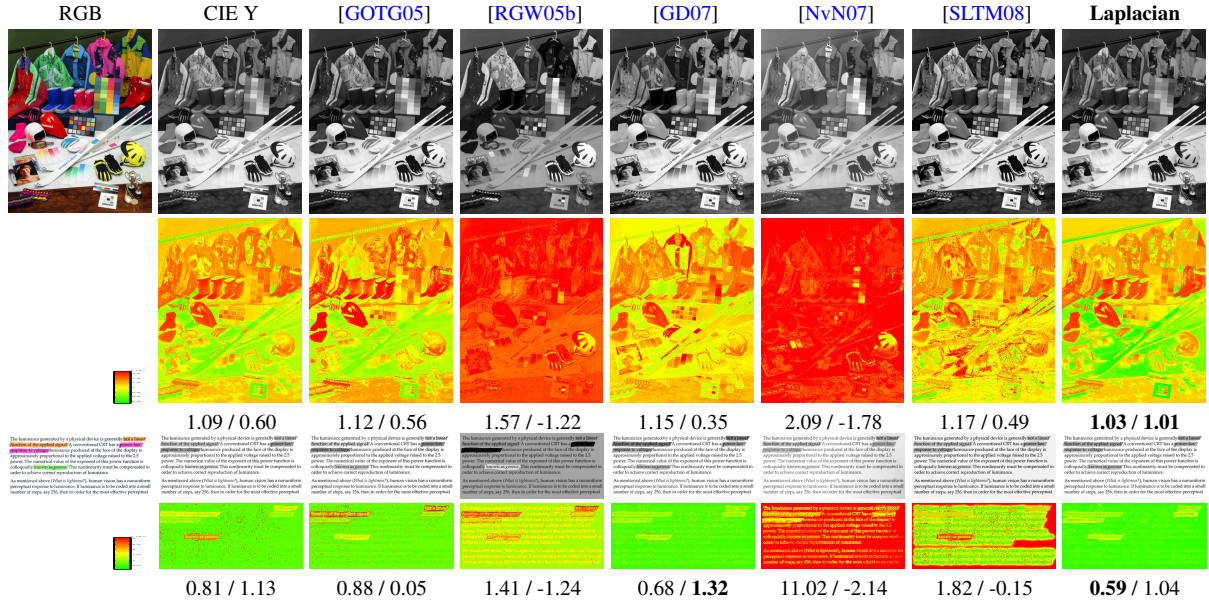

\ContinuedFloat 
\centering
\begin{minipage}[b]{0.12\linewidth}
\centering \small RGB
\end{minipage}
\begin{minipage}[b]{0.12\linewidth}
\centering \small CIE Y
\end{minipage}
\begin{minipage}[b]{0.12\linewidth}
\centering \small \cite{Gooch05} 
\end{minipage}
\begin{minipage}[b]{0.12\linewidth}
\centering \small \cite{Rasche05} 
\end{minipage}
\begin{minipage}[b]{0.12\linewidth}
\centering \small \cite{Grundland07}
\end{minipage}
\begin{minipage}[b]{0.12\linewidth}
\centering \small \cite{Neumann07}
\end{minipage}
\begin{minipage}[b]{0.12\linewidth}
\centering \small \cite{Smith08}
\end{minipage}
\begin{minipage}[b]{0.12\linewidth}
\centering \small \textbf{Laplacian}
\end{minipage}

\rgbtogray{Ski_TC8-03_sRGB}{1.09 / 0.60}{1.12 / 0.56}{1.57 / -1.22}{1.15 / 0.35}{2.09 / -1.78}{1.17 / 0.49}{\textbf{1.03} / \textbf{1.01}}
\rgbtogray{text}{0.81 / 1.13}{0.88 / 0.05}{1.41 / -1.24}{0.68 / \textbf{1.32}}{11.02 / -2.14}{1.82 / -0.15}{\textbf{0.59} / 1.04}
\vspace{0mm}\\ 

\caption{(Continues from previous page) Decolorization experiment results. Left: original RGB image, right: grayscale conversion results. RWMS error images and mean RWMS (the smaller the better) / z-score (the larger the better) values. Our Laplacian colormap method performs the best in most cases.
}
\end{figure*}

\section{Conclusions}

Laplacian colormaps address the problem of structure-preserving color transformations by relying on Laplacians
as image structure descriptors and using Laplacian commutativity as a criterion for structure preservation.
Given a parametric colormap, we optimize for the parameters that produce an image
whose Laplacian commutes as much as possible with the one of the original image, thus preserving its original structure. 
Since Laplacians can be defined in any colorspace, our approach can be applied to different kinds of colormaps
(global or local, with any number of input and output channels, and where part of the mapping is provided a priori). Moreover, Laplacians can be computed using similarity of local feature descriptors rather than individual pixels colors.  
%
Computationally, pixel-wise relationships is the main bottleneck of our approach: a benchmark we ran on all the 25 pictures of \u{C}ad\'{i}k's dataset using a MacBook Pro with 8GB RAM showed that the average computation time for color-to-grayscale conversion was 117 seconds. 
This cost can be alleviated by computing Laplacians on resized images, at a price of potentially lower accuracy. 


Overall, we believe that our results show the promise in the use of Laplacians commutators to measure structure similarity, and seem to be the first application of rather theoretical results on joint diagonalization of matrices to very practical problems in image processing. 
In future works, we intend to explore additional applications such as image correspondence and alignment. We believe that our approach will be especially useful when handling visually different but structurally similar scenes.

\vspace{-2mm}
\section*{Acknowledgements} 
\vspace{-2mm}
This research was supported by the ERC Starting Grant No. 307047 (COMET).
\vspace{-2mm}


\bibliographystyle{eg-alpha-doi}

\bibliography{paper}

\section*{Appendix A: Gradients of the cost function}
\begin{small}
%
Let $\bb{L}_{\Phi_{\bt}(\Xx)} = \bb{D}_{\Phi_{\bt(\Xx)}} - \bb{W}_{\Phi_{\bt(\Xx)}}$ be the image Laplacian as defined in~(\ref{eq:laplacian1}). 
We denote by $|W|$ the number of non-zero elements in the adjacency matrix $\bb{W}_{\Phi_{\bt(\Xx)}}$, and by $n$ the number of parameters $\bb{\theta}$ of the colormap, respectively. 
The non-zero elements $w_{ij} > 0$ are indexed as $\bb{w}_{\bb{\theta}} = (w_1, \hdots, w_{|W|}) = (w_{i_1,j_1}, \hdots, w_{i_{|W|},j_{|W|}})$. 
%
%
%
%
%
$\phi_{\bb{\theta}}^i : \RR ^{d} \rightarrow \RR$ denotes the $i$th channel of the colormap, such that 
$\phi_{\bb{\theta}}(\xx) = (\phi_{\bb{\theta}}^1(\xx), \cdots,\phi_{\bb{\theta}}^{d'}(\xx) )^ {\Tr}$, and $\nabla_{\bb{\theta}}\phi_{\bb{\theta}}^i$ is its gradient w.r.t. $\bb{\theta}$.  

We now derive the gradients of the cost function~(\ref{eq:costfunction1}). 
The gradient of the $\mu _2$-term is trivial, 
$$
\nabla_{\bb{\theta}}\| \bb{\theta} - \bb{\theta}_0 \|_2^2 = 2 (\bt - \bt _0). 
$$

Denote by $\bb{G}_ {{\Phi_{\bt}}(\Xx)}^i$ the matrix of size $NM \times n$, whose $j$th row is the gradient of the $i$th channel at the $j$th pixel, 
$\nabla_{\bb{\theta}} \phi_{\bb{\theta}}^i (\xx _j)^\Tr$, 
and define $NM d' \times n$ matrix 
$\bb{G}_ {{\Phi_{\bt}}(\Xx)}  = ( (\bb{G}_ {{\Phi_{\bt}}(\Xx)} ^1)^ {\Tr}, \cdots,  (\bb{G}_ {{\Phi_{\bt}}(\Xx)} ^{d'})^ {\Tr})^{\Tr}$. Differentiating the $\mu _3$-term w.r.t $\bt$ gives 
\begin{align*}
\label{eq:gradmu3}
\nabla _{\bt}  \|\Phi_{\bb \theta}(\Xx_\mathrm{c}) -\Yy_\mathrm{c}\|_\mathrm{F}^2 = 2\bb{G}_ {{\Phi_{\bt}}(\Xx)}^ \Tr \left(  \mathrm{vec}( \Phi_{\bb \theta}(\Xx_\mathrm{c})) - \mathrm{vec}(\Yy_\mathrm{c})   \right). 
\end{align*}

The gradients of the first two terms of~(\ref{eq:costfunction1}) are obtained by applying the chain rule. 
%
%
First, we differentiate the terms w.r.t $\bb{w}_{\bb{\theta}}$, obtaining a gradient of size $|W| \times 1$. Next, we differentiate w.r.t. $\bb{\theta}$. 
The gradient of the adjacency matrix elements $w_{ij}$ w.r.t. $\bb{\theta}$ is  
\begin{equation*}
\label{eq:gradwij}
\nabla _{\bt} w_{ij} = -\frac{w_{ij}}{ \sigma_r^{2} } \sum \limits _{k=1}^{d'}  (\phi_{\bb{\theta}}^k (\xx _i) - \phi_{\bb{\theta}}^k (\xx _j) )(\nabla _{\bt} \phi_{\bb{\theta}}^k (\xx _i) - \nabla _{\bt} \phi_{\bb{\theta}}^k (\xx _j) ).
\end{equation*}
The gradient of the commutator ($\mu_0$-term) is: 
\begin{align*}
\label{eq:gradcomut}
\notag
& \frac{\dd}{\dd w_{ij} }  \|[ \Ll_{\Xx}, \Ll_{\Phi_{\bb{\theta}}(\Xx)}]  \|_\mathrm{F}^2  = \\ &  -2 \left(  \bb{O}_{1} - \Ll_{\Phi_{\bb{\theta}}(\Xx)}^{\Tr}  [ \Ll_{\Phi_{\bb{\theta}}(\Xx)},\Ll_{\Xx} ]  - \bb{O}_{2} +    [\Ll_{\Phi_{\bb{\theta}}(\Xx)},\Ll_{\Xx}] \Ll_{\Phi_{\bb{\theta}}(\Xx)}^{\Tr}  \right)_{ij}.
\end{align*}
The gradient of the $\mu_1$-term is: 
\begin{equation*}
\label{eq:gradLxmL}
\frac{\dd}{\dd w_{ij}} \|\Ll_{\Xx} - \Ll_{\Phi_{\bb{\theta}}(\Xx)} \|_\mathrm{F}^2 = 2\left( \bb{O} + \Ll_{\Xx} - \Ll_{\Phi_{\bb{\theta}}(\Xx)}  \ \right) _{ij}.
\end{equation*}
Here, $\bb{O}, \bb{O}_k$ are matrices with equal columns given by 
$$
\bb{O} = (\mathrm{diag}(\Ll_{\Xx} ), \cdots, \mathrm{diag}(\Ll_{\Xx} )), 
$$  
$$
\bb{O}_1 = (\mathrm{diag}(\Ll_{\Phi_{\bb{\theta}}(\Xx)}^{\Tr}   [\Ll_{\Phi_{\bb{\theta}}(\Xx)},\Ll_{\Xx}]   ), \cdots, \mathrm{diag}(\Ll_{\Phi_{\bb{\theta}}(\Xx)}^{\Tr}  [\Ll_{\Phi_{\bb{\theta}}(\Xx)},\Ll_{\Xx}] ) )
$$
$$
\bb{O}_2 = (\mathrm{diag}(   [\Ll_{\Phi_{\bb{\theta}}(\Xx)},\Ll_{\Xx}]    \Ll_{\Phi_{\bb{\theta}}(\Xx)}^{\Tr}  ), \cdots, \mathrm{diag}(  [\Ll_{\Phi_{\bb{\theta}}(\Xx)},\Ll_{\Xx}]   \Ll_{\Phi_{\bb{\theta}}(\Xx)}^{\Tr})    ).
$$

Finally, the gradient of the colormap appearing in the expressions above depends on the choice of the colormap. 
For all the experiments using the colormap $\phi_{\bb{\theta}}^i(\xx)$ defined in Section~\ref{sec:res}, the derivation of the gradient is straightforward. 
In the experiments simulating color blindness, the colormap is $(\phi_{\bb{\theta}}^i \circ \Psi)(\xx)$, whose gradient is given as $\nabla_{\bb{\theta}} (\phi_{\bb{\theta}}^i \circ \Psi)(\xx) = \bb{J} _{\Psi}^{\Tr} \nabla_{\bt} \phi_{\bb{\theta}}^i(\xx)$, where  $\bb{J}_{\Psi}$ is the  Jacobian of $\Psi$. In our experiments, transformation $\Psi$ simulating the deficient observer is linear  $\Psi(\xx) = \bb{A}\xx$, and thus $\bb{J} _{\Psi}=\bb{A}$. 

\end{small}


\end{document}